\documentclass{article}

\usepackage[main, final]{neurips_2026}

\usepackage[utf8]{inputenc} % allow utf-8 input
\usepackage[T1]{fontenc}    % use 8-bit T1 fonts
\usepackage{hyperref}       % hyperlinks
\usepackage{url}            % simple URL typesetting
\usepackage{booktabs}       % professional-quality tables
\usepackage{amsfonts}       % blackboard math symbols
\usepackage{nicefrac}       % compact symbols for 1/2, etc.
\usepackage{microtype}      % microtypography

\usepackage[most]{tcolorbox} % colored boxes
\hypersetup{
    colorlinks=true,
    linkcolor=blue!75!black,
    citecolor=blue!75!black,
    urlcolor=black
}

\usepackage[inline]{enumitem}
\usepackage{tasks}
\title{\ftmt: Morpho Adaptation via Asymmetric Expansion of Kronecker Product}

\author{%
  An Nguyen \\
  Department of Artificial Intelligence\\
  Korea University \\
  \texttt{nguyenan@korea.ac.kr} \\
  \And
  Anh Tong\\
  Department of Artificial Intelligence\\
  Korea University \\
  \texttt{anhtong12@korea.ac.kr} \\
}

\usepackage{amsmath,amsfonts,bm}

\def\1{\bm{1}}

\def\rvg{{\mathbf{g}}}

\def\rvm{{\mathbf{m}}}

\def\rvx{{\mathbf{x}}}
\def\rvy{{\mathbf{y}}}

\def\zeromatrix{{\mathbf{0}}}

\def\rmA{{\mathbf{A}}}
\def\rmB{{\mathbf{B}}}
\def\rmC{{\mathbf{C}}}
\def\rmD{{\mathbf{D}}}
\def\rmE{{\mathbf{E}}}
\def\rmF{{\mathbf{F}}}
\def\rmG{{\mathbf{G}}}

\def\rmI{{\mathbf{I}}}

\def\rmK{{\mathbf{K}}}
\def\rmL{{\mathbf{L}}}
\def\rmM{{\mathbf{M}}}

\def\rmO{{\mathbf{O}}}
\def\rmP{{\mathbf{P}}}
\def\rmQ{{\mathbf{Q}}}
\def\rmR{{\mathbf{R}}}
\def\rmS{{\mathbf{S}}}
\def\rmT{{\mathbf{T}}}
\def\rmU{{\mathbf{U}}}
\def\rmV{{\mathbf{V}}}
\def\rmW{{\mathbf{W}}}
\def\rmX{{\mathbf{X}}}
\def\rmY{{\mathbf{Y}}}
\def\rmZ{{\mathbf{Z}}}

\def\ermA{{\textnormal{A}}}
\def\ermB{{\textnormal{B}}}

\def\ermI{{\textnormal{I}}}

\def\vtheta{{\bm{\theta}}}

\def\ve{{\bm{e}}}

\def\vm{{\bm{m}}}

\def\vu{{\bm{u}}}
\def\vv{{\bm{v}}}

\def\vx{{\bm{x}}}
\def\vy{{\bm{y}}}
\def\vz{{\bm{z}}}

\def\evs{{s}}

\def\mC{{\bm{C}}}
\def\mD{{\bm{D}}}

\def\mI{{\bm{I}}}

\def\mP{{\bm{P}}}
\def\mQ{{\bm{Q}}}

\def\mU{{\bm{U}}}

\def\mX{{\bm{X}}}

\DeclareMathAlphabet{\mathsfit}{\encodingdefault}{\sfdefault}{m}{sl}
\SetMathAlphabet{\mathsfit}{bold}{\encodingdefault}{\sfdefault}{bx}{n}
\newcommand{\tens}[1]{\bm{\mathsfit{#1}}}

\def\tL{{\tens{L}}}

\def\tR{{\tens{R}}}

\def\gB{{\mathcal{B}}}

\def\gD{{\mathcal{D}}}

\def\gF{{\mathcal{F}}}
\def\gG{{\mathcal{G}}}

\def\gL{{\mathcal{L}}}

\def\gR{{\mathcal{R}}}
\def\gS{{\mathcal{S}}}
\def\gT{{\mathcal{T}}}

\def\sR{{\mathbb{R}}}

\def\sZ{{\mathbb{Z}}}

\newcommand{\E}{\mathbb{E}}

\newcommand{\Var}{\mathrm{Var}}

\DeclareMathOperator*{\argmin}{arg\,min}

\DeclareMathOperator{\vecop}{vec}
\DeclareMathOperator{\unvecop}{unvec}
\DeclareMathOperator{\rankop}{rank}
\DeclareMathOperator{\erankop}{erank}
\def\sRs{{\tilde{\mathbb{R}}}}

\DeclareMathOperator*{\row}{Row}
\DeclareMathOperator{\nul}{Null}

\usepackage{tikz}
\usepackage{tikz-3dplot}
\usetikzlibrary{arrows.meta,calc}
\usetikzlibrary{math}
\usetikzlibrary{angles, quotes}
\usetikzlibrary{3d,calc,perspective}

\usepackage{booktabs, multirow}

\usepackage{relsize} % \inv
\usepackage{amsmath}
\usepackage{amsfonts}
\usepackage{amssymb}      
\usepackage{amsbsy}       

\usepackage[x11names,table]{xcolor} 
\definecolor{tableblue}{RGB}{215, 245, 250}
\usepackage{arydshln} % after colortbl
\usepackage{mathtools}
\usepackage{subcaption} % 2 figures
\usepackage{siunitx} % scientific numbers

\usepackage{amsthm}
\usepackage{cleveref}
\newtheorem{definition}{Definition}[section]
\newtheorem{theorem}{Theorem}[section]
\newtheorem{lemma}{Lemma}[section]
\newtheorem*{theorem*}{Theorem}
\newtheorem*{corollary*}{Corollary} % indexed by theorem
\newtheorem{corollary}{Corollary}[section]
\usepackage{caption} % caption without {table}

\crefname{equation}{Eq.}{Eqs.}
\Crefname{equation}{Equation}{Equations}
\crefname{section}{Sec.}{Secs.}
\Crefname{section}{Section}{Sections}
\crefname{figure}{Fig.}{Figs.}
\Crefname{figure}{Figure}{Figures}
\crefname{table}{Table}{Tables}
\crefname{theorem}{Theorem}{Theorems}
\crefname{appendix}{Appendix}{Appendices}
\Crefname{appendix}{Appendix}{Appendices}
\crefname{lemma}{Lemma}{Lemmas}

\crefname{corollary}{Corollary}{Corollaries}

\usepackage{listings}
\usepackage{xcolor}

\definecolor{codebg}{gray}{0.95}              % Màu nền xám nhạt
\definecolor{codestring}{rgb}{0.6, 0.1, 0.1}  % Màu đỏ mận cho chuỗi (string)
\definecolor{codekeyword}{rgb}{0.0, 0.3, 0.7} % Màu xanh lam cho từ khóa (nếu cần)
\definecolor{codecomment}{rgb}{0.2, 0.5, 0.2} % Màu xanh lá cho comment

\lstdefinestyle{pytorchstyle}{
    language=Python,                  % Khai báo ngôn ngữ là Python
    backgroundcolor=\color{codebg},   % Màu nền
    basicstyle=\ttfamily\small,       % Font chữ code
    keywordstyle=\color{codekeyword}\bfseries, % Đổi màu và in đậm từ khóa (def, return, import...)
    stringstyle=\color{codestring},   % Đổi màu chuỗi (phần chữ nằm trong dấu nháy "")
    commentstyle=\color{codecomment}\itshape,  % Đổi màu và in nghiêng chú thích (#)
    frame=tb,                         % Kẻ đường ở trên và dưới
    framerule=0.8pt,                  
    rulecolor=\color{darkgray},       
    showstringspaces=false,           
    aboveskip=12pt,                   
    belowskip=12pt,                   
    xleftmargin=4pt,                  
    xrightmargin=4pt,
    morekeywords={torch, einsum}
}
\usepackage{etoc}

\usepackage{algorithm}
\usepackage{algpseudocode}
\usepackage{float}

\usepackage{wrapfig}
\usepackage{afterpage} % footnote for tables
\newcommand{\vX}{\mathbf{X}}

\DeclareMathOperator{\diag}{diag}
\usepackage[normalem]{ulem}
\usepackage{multirow}
\usepackage{float}
\usepackage{siunitx}
\usepackage{verbatim} % comment env
\usepackage{caption} % table and fig in a single column
\usepackage{titletoc}

\usepackage{xspace} % space after \ftmt
\newtcolorbox{promptbox}{
    colback=white,
    colframe=black,
    sharp corners,
    boxrule=0.5pt,
    fontupper=\small\ttfamily,
    left=10pt, right=10pt, top=10pt, bottom=10pt
}

\newcommand{\ftmt}{\text{SaMA}\xspace}
\newcommand{\ftmrm}{\mathrm{SaMA}}
\newcommand{\oftmto}{\text{SaMA}}
\newcommand{\repname}{\text{Morpho}\xspace}
\newcommand{\mset}{\gF}

\begin{document}

\maketitle

% Result commands

\newcommand{\tableDropMainResult}{
    \begin{table}[htbp]
        \centering
        \small
        \caption{Benchmark of various fine-tuning methods on the DROP dataset. Baselines  with $\dagger$ are from QuanTA~\citep{chen2024quanta}. Overall, \ftmt consistently exhibits better performance compared to other baselines. }
        \setlength{\tabcolsep}{5pt}% col gap
        \resizebox{0.55\textwidth}{!}{%
        \begin{tabular}{llcc}
            \toprule
            \textbf{Model} & \textbf{FT Method} & \textbf{\%Params} & \textbf{$F_1$ Score ($\uparrow$)} \\
            \midrule
            \multirow{8}{*}{Llama2$_{\text{7B}}$} 
            & FFT (\texttt{q\_proj} and \texttt{v\_proj}) & 15.93 & 68.98 \\
            & Series$^\dagger$ & 0.747 & 58.8 \\
            & Parallel$^\dagger$ & 0.747 & 59.0 \\
            & LoRA$_{r=8}$ & 0.062 & 62.39 \\
            & LoRA$_{r=32}$ & 0.249 & 62.40 \\
            & LoRA$_{r=128}$ & 0.996 & 58.82 \\
            & QuanTA$_{\text{16-8-8-4}}$ & 0.041 & 62.09 \\
            & QuanTA$_{\text{16-16-16}}$ & 0.261 & 62.72 \\
            \rowcolor{tableblue} & \textbf{SaMA}$_\text{2,2}$ & 0.016 & {66.01} \\
            \rowcolor{tableblue} & \textbf{SaMA}$_\text{4,4}$ & 0.031 & {65.15} \\
            \rowcolor{tableblue} & \textbf{SaMA}$_\text{8,8}$ & 0.062 & {66.61} \\
            \rowcolor{tableblue} & \textbf{SaMA}$_\text{16,16}$ & 0.124 & {67.17} \\
            \rowcolor{tableblue} & \textbf{SaMA}$_\text{32,32}$ & 0.248 & \textit{67.95} \\
            \rowcolor{tableblue} & \textbf{SaMA}$_\text{64,64}$ & 0.496 & \textbf{68.07} \\
            \midrule
            \multirow{2}{*}{Llama2$_{\text{13B}}$} 
            & LoRA$^\dagger_{r=8}$ & 0.050 & 61.0 \\
            & QuanTA$^\dagger_{\text{16-8-8-5}}$ & 0.029 & 69.0 \\
            \rowcolor{tableblue} & \textbf{SaMA}$_\text{4,4}$ & 0.025 & \textbf{73.12} \\
            \midrule
            \bottomrule
        \end{tabular}%
        }
        \label{tab:llama_drop_bench}
    \end{table}
}

\newcommand{\tableArithMainResult}{
        \setlength{\tabcolsep}{4pt}
        \caption{Comparison of PEFT methods across math benchmarks. Baselines ($\dagger$) are from~\citep{singhal2025abba}. See \cref{tab:arith_comparison_full_results} for the detailed report.}
            \begin{tabular}{l cc cc} % Đã bỏ các dấu |
                \toprule
                \multirow{2}{*}{\textbf{Method}} & \multicolumn{2}{c}{\textbf{Mistral-7B}} & \multicolumn{2}{c}{\textbf{Gemma-2 9B}} \\
                \cmidrule(lr){2-3} \cmidrule(lr){4-5}
                 & \textbf{GSM8K} & \textbf{MATH} & \textbf{GSM8K } & \textbf{MATH} \\
                \midrule
                Full FT$^\dagger$ & 63.87 & 17.65 & 79.23 & 38.02 \\
                LoRA$_{r=32}$ & 62.93 & 14.14 & 77.45 & 35.05 \\
                PiSSA$^\dagger$ & 62.43 & 16.52 & 77.12 & 37.04 \\
                LoRA-Pro$^\dagger$ & 63.07 & 17.32 & 78.26 & 37.53 \\
                ABBA$_{r=32}$ & 64.48 & 17.95 & 78.85 & \textit{39.59} \\
                \midrule
                \rowcolor{tableblue} \textbf{SaMA}$_\text{8,8}$ & \textit{67.70} & \textit{18.44} & \textit{80.36} & {39.36} \\
                \bottomrule
            \end{tabular}
        \label{tab:arith_comparison_results}
}

\newcommand{\tableArithMainResultFull}{
    \begin{table}[htbp]
        \centering
        \small
        \setlength{\tabcolsep}{8pt}
        \captionof{table}{Detailed arithmetic reasoning results across PEFT methods on Mistral-7B and Gemma-2 9B. This table extends the main-paper summary with full comparison along with parameter counts. Baselines ($\dagger$) are from~\citep{singhal2025abba}. Best results are in \textbf{bold}, the runner-ups are in \textit{italic}. In general, \ftmt consistently surpasses other baseline regardless of lower budgets.}
            \begin{tabular}{l ccc ccc}
                \toprule
                \multirow{2}{*}{\textbf{Method}} & \multicolumn{3}{c}{\textbf{Mistral-7B}} & \multicolumn{3}{c}{\textbf{Gemma-2 9B}} \\
                \cmidrule(lr){2-4} \cmidrule(lr){5-7}
                 & \textbf{\# Params} & \textbf{GSM8K} & \textbf{MATH} & \textbf{\# Params} & \textbf{GSM8K} & \textbf{MATH} \\
                \midrule
                Full FT$^\dagger$ & 7.24 B & 63.87 & 17.65 & 9.24 B & 79.23 & 38.02 \\
                LoRA$_{r=32}$ & 83.88 M & 62.93 & 14.14 & 108.04 M & 77.45 & 35.05 \\
                rsLoRA$^\dagger$ & 83.88 M & 62.15 & 16.24 & 108.04 M & 76.84 & 36.88 \\
                PiSSA$^\dagger$ & 83.88 M & 62.43 & 16.52 & 108.04 M & 77.12 & 37.04 \\
                DoRA$^\dagger$ & 85.26 M & 62.65 & 16.64 & 109.88 M & 77.58 & 37.04 \\
                LoRA-Pro$^\dagger$ & 83.88 M & 63.07 & 17.32 & 108.04 M & 78.26 & 37.53 \\
                HiRA$^\dagger$ & 83.88 M & 63.15 & 17.44 & 108.04 M & 78.47 & 38.22 \\
                ABBA$_{r=32}$ & 83.88 M & 64.48 & 17.95 & 108.04 M & 78.85 & \textit{39.59} \\
                \midrule
                \rowcolor{tableblue} \textbf{SaMA}$_\text{8,8}$ & 10.49 M & {67.70} & \textit{18.44} & 13.51 M & \textit{80.36} & {39.36} \\
                \rowcolor{tableblue} \textbf{SaMA}$_\text{16,16}$ & 41.94 M & \textbf{68.23} & \textbf{19.18} & 54.02 M & \textbf{81.58} & \textbf{39.76} \\
                \bottomrule
            \end{tabular}
        \label{tab:arith_comparison_full_results}
    \end{table}
}

\newcommand{\tableCmsAttnResult}{
    % \vspace{-10pt}
    \begin{table*}[htbp]
    \centering
    \caption{Comparison of \ftmt and QuanTA in Q-V attention fine-tuning settings. Results with a symbol $\star$ are from QuanTA paper~\citep{chen2024quanta}. Across all model scales, although the configurations are set with comparable or 20-40\% lower parameter budgets, \ftmt stably results in superior performance. This demonstrate the substantial generalization capability of our method.}
    \label{tab:main_results_final}
    \resizebox{0.95\textwidth}{!}{%
        \begin{tabular}{llcccccccccc}
        \toprule
        \textbf{Model} & \textbf{Method} & \textbf{\#Params\%} & \textbf{BoolQ} & \textbf{PIQA} & \textbf{SIQA} & \textbf{HellaS.} & \textbf{WinoG.} & \textbf{ARC-e} & \textbf{ARC-c} & \textbf{OBQA} & \textbf{AVG} \\
        \midrule
        ChatGPT & -- & -- & 73.1 & 85.4 & 68.5 & 78.5 & 66.1 & 89.8 & 79.9 & 74.8 & 77.0 \\
        \midrule
        \multirow{1}{*}{LLaMA-7B} 
            & QuanTA$^\star$ & 0.041 & 71.6 & 83.0 & 79.7 & 91.8 & 81.8 & 84.0 & 68.3 & 82.1 & 80.3 \\
            \rowcolor{tableblue} & $\textbf{SaMA}_{4 \times 4}$ & 0.031 & 70.8 & 84.0 & 81.4 & 93.8 & 83.4 & 86.7 & 71.3 & 84.0 & \textbf{81.9} \\
        \midrule
        \multirow{1}{*}{LLaMA-13B} 
            & QuanTA$^\star$ & 0.029 & 73.2 & 85.4 & 82.1 & 93.4 & 85.1 & 87.8 & 73.3 & 84.4 & 83.1 \\
            \rowcolor{tableblue} & $\textbf{SaMA}_{4 \times 4}$ & 0.025 & 72.6 & 87.4 & 83.3 & 95.6 & 85.6 & 88.4 & 76.2 & 88.4 & \textbf{84.7} \\
        \midrule
        \multirow{1}{*}{LLaMA2-7B} 
            & QuanTA$^\star$ & 0.041 & 72.4 & 83.8 & 79.7 & 92.5 & 83.9 & 85.3 & 72.5 & 82.6 & 81.6 \\
            \rowcolor{tableblue} & $\textbf{SaMA}_{4 \times 4}$ & 0.031 & 72.5 & 85.1 & 81.2 & 94.4 & 84.8 & 87.9 & 75.0 & 84.5 & \textbf{83.2} \\
        \midrule
        \multirow{1}{*}{LLaMA2-13B} 
            & QuanTA$^\star$ & 0.041 & 75.8 & 86.9 & 81.2 & 94.4 & 87.0 & 89.6 & 77.9 & 85.2 & 84.8 \\
            \rowcolor{tableblue} & $\textbf{SaMA}_{4 \times 4}$ & 0.025 & 75.4 & 87.3 & 82.9 & 96.0 & 88.0 & 91.5 & 80.8 & 86.2 & \textbf{86.0} \\
        \midrule
        \multirow{1}{*}{LLaMA3-8B} 
            & QuanTA$^\star$ & 0.035 & 74.3 & 88.1 & 81.8 & 95.1 & 87.3 & 91.1 & 81.7 & 87.2 & 85.8 \\
            \rowcolor{tableblue} & $\textbf{SaMA}_{4 \times 4}$ & 0.021 & 74.0 & 90.1 & 82.1 & 96.2 & 87.2 & 93.2 & 81.7 & 87.0 & \textbf{86.4} \\
        \bottomrule
        \end{tabular}%
    }
    \end{table*}
}

\newcommand{\tableAppVtabResults}{
\begin{table*}[htbp]
    \centering
    \caption{Performance comparison of \ftmt with the baseline and efficient methods on the VTAB-1k classification benchmark with ViT-B/16 86M pre-trained on ImageNet-21k. The best results among these baselines are in \textbf{bold}, excluding SSF$^\star$ which utilizes augmented ViT backbone by AugReg. All baseline results are from \citep{dong2024rlrr}.}
    \label{tab:app_vtab_results}
    
    \resizebox{\textwidth}{!}{
        \begin{tabular}{l cccccccc ccccc ccccccccc cc}
        \toprule
        \multirow{2}{*}{\textbf{Methods}} & \multicolumn{8}{c}{\textbf{Natural}} & \multicolumn{5}{c}{\textbf{Specialized}} & \multicolumn{9}{c}{\textbf{Structured}} & \multicolumn{1}{c}{\multirow{2}{*}[-1.4em]{\rotatebox{90}{\textbf{Mean Total}}}} & \multicolumn{1}{c}{\multirow{2}{*}[-1.4em]{\rotatebox{90}{\textbf{Params.(M)}}}} \\
        \cmidrule(lr){2-9} \cmidrule(lr){10-14} \cmidrule(lr){15-23}
        & \rotatebox{90}{CIFAR-100} & \rotatebox{90}{Caltech101} & \rotatebox{90}{DTD} & \rotatebox{90}{Flowers102} & \rotatebox{90}{Pets} & \rotatebox{90}{SVHN} & \rotatebox{90}{Sun397} & \rotatebox{90}{\textbf{Mean}} & \rotatebox{90}{Camelyon} & \rotatebox{90}{EuroSAT} & \rotatebox{90}{Resisc45} & \rotatebox{90}{Retinopathy} & \rotatebox{90}{\textbf{Mean}} & \rotatebox{90}{Clevr-Count} & \rotatebox{90}{Clevr-Dist} & \rotatebox{90}{DMLab} & \rotatebox{90}{KITTI-Dist} & \rotatebox{90}{dSpr-Loc} & \rotatebox{90}{dSpr-Ori} & \rotatebox{90}{sNORB-Azim} & \rotatebox{90}{sNORB-Ele} & \rotatebox{90}{\textbf{Mean}} & & \\ 
        \midrule
            Full fine-tuning & 68.9 & 87.7 & 64.3 & 97.2 & 86.9 & 87.4 & 38.8 & 75.9 & 79.7 & 95.7 & 84.2 & 73.9 & 83.4 & 56.3 & 58.6 & 41.7 & 65.5 & 57.5 & 46.7 & 25.7 & 29.1 & 47.6 & 65.6 & 85.80 \\
            Linear probing & 63.4 & 85.0 & 63.2 & 97.0 & 86.3 & 36.6 & 51.0 & 68.9 & 78.5 & 87.5 & 68.6 & 74.0 & 77.2 & 34.3 & 30.6 & 33.2 & 55.4 & 12.5 & 20.0 & 9.6 & 19.2 & 26.9 & 52.9 & 0.04 \\
        \midrule
            Adapter & 74.1 & 86.1 & 63.2 & 97.7 & 87.0 & 34.6 & 50.8 & 70.5 & 76.3 & 88.0 & 73.1 & 70.5 & 77.0 & 45.7 & 37.4 & 31.2 & 53.2 & 30.3 & 25.4 & 13.8 & 22.1 & 32.4 & 55.8 & 0.27 \\
            Bias & 72.8 & 87.0 & 59.2 & 97.5 & 85.3 & 59.9 & 51.4 & 73.3 & 78.7 & 91.6 & 72.9 & 69.8 & 78.3 & 61.5 & 55.6 & 32.4 & 55.9 & 66.6 & 40.0 & 15.7 & 25.1 & 44.1 & 62.1 & 0.14 \\
            VPT-Shallow & 77.7 & 86.9 & 62.6 & 97.5 & 87.3 & 74.5 & 51.2 & 76.8 & 78.2 & 92.0 & 75.6 & 72.9 & 79.7 & 50.5 & 58.6 & 40.5 & 67.1 & 68.7 & 36.1 & 20.2 & 34.1 & 47.0 & 64.9 & 0.11 \\
            VPT-Deep & \textbf{78.8} & 90.8 & 65.8 & 98.0 & 88.3 & 78.1 & 49.6 & 78.5 & 81.8 & 96.1 & 83.4 & 68.4 & 82.4 & 68.5 & 60.0 & 46.5 & 72.8 & 73.6 & 47.9 & 32.9 & 37.8 & 55.0 & 69.4 & 0.60 \\
            LoRA & 67.1 & 91.4 & 69.4 & 98.8 & 90.4 & 85.3 & 54.0 & 79.5 & 84.9 & 95.3 & 84.4 & 73.6 & 84.6 & 82.9 & \textbf{69.2} & 49.8 & 78.5 & 75.7 & 47.1 & 31.0 & \textbf{44.0} & 59.8 & 72.3 & 0.29 \\
            AdaptFormer & 70.8 & 91.2 & 70.5 & 99.1 & 90.9 & 86.6 & 54.8 & 80.6 & 83.0 & 95.8 & 84.4 & 76.3 & 84.9 & 81.9 & 64.3 & 49.3 & 80.3 & 76.3 & 45.7 & 31.7 & 41.1 & 58.8 & 72.3 & 0.16 \\
            FacT-TK$_{<32}$ & 70.6 & 90.6 & 70.8 & 99.1 & 90.7 & 88.6 & 54.1 & 80.6 & 84.8 & \textbf{96.2} & 84.5 & 75.7 & 85.3 & 82.6 & 68.2 & 49.8 & 80.7 & 80.8 & 47.4 & 33.2 & 43.0 & 60.7 & 73.2 & 0.07 \\
            ARC & 72.2 & 90.1 & 72.7 & 99.0 & 91.0 & \textbf{91.9} & 54.4 & 81.6 & 84.9 & 95.7 & \textbf{86.7} & 75.8 & 85.8 & 80.7 & 67.1 & 48.7 & 81.6 & 79.2 & 51.0 & 31.4 & 39.9 & 60.0 & 73.4 & 0.13 \\
            RLRR & 75.6 & \textbf{92.4} & \textbf{72.9} & \textbf{99.3} & 91.5 & 89.8 & 57.0 & \textbf{\textit{82.7}} & 86.8 & 95.2 & 85.3 & 75.9 & 85.8 & 79.7 & 64.2 & \textbf{53.9} & \textbf{82.1} & 83.9 & 53.7 & \textbf{33.4} & 43.6 & \textbf{61.8} & 74.5 & 0.33 \\
        \midrule
            SSF$^\star$ & 69.0 & 92.6 & 75.1 & 99.4 & 91.8 & 90.2 & 52.9 & 81.6 & 87.4 & 95.9 & 87.4 & 75.5 & 86.6 & 75.9 & 62.3 & 53.3 & 80.6 & 77.3 & 54.9 & 29.5 & 37.9 & 59.0 & 73.1 & 0.24 \\
        \midrule
            \rowcolor{tableblue} $\textbf{SaMA}_{2-2}$ & 75.3 & \textbf{92.4} & 72.53 & \textbf{99.2} & \textbf{92.0} & 89.4 & \textbf{57.8} & \textbf{\textit{82.7}} & \textbf{88.2} & 95.8 & 86.5 & \textbf{76.4} & \textbf{\textit{86.7}} & \textbf{83.0} & 63.4 & 53.0 & 81.4 & \textbf{84.7} & \textbf{53.8} & 33.0 & 41.4 & 61.7 & \textbf{\underline{74.7}} & 0.33 \\
        \bottomrule
        \end{tabular}
    }
\end{table*}
}

\begin{abstract}

Low-rank adaptation (LoRA) has attracted significant attention in the parameter-efficient fine-tuning (PEFT) landscape. However, its design suffers from two fundamental limitations. First, its invariance manifold induces flat directions in the loss landscape, creating an optimization bottleneck for adaptive optimizers. Second, its effective rank scales only \emph{linearly} with the parameter budget. In this paper, we propose \emph{Scalable Morpho Adaptation (SaMA)}, a PEFT method grounded in a principled generalization of the Kronecker product. By relaxing the shared-block constraint in the perfect-shuffle factorization, SaMA spans from low to full rank and achieves a practical rank that scales \emph{quadratically} with the parameter budget, up to full-rank. Furthermore, we show that its invariance manifold collapses to a diagonal group, making it structurally more optimizer-friendly. Empirically, SaMA consistently outperforms strong PEFT baselines on commonsense and arithmetic reasoning benchmarks while being more parameter-efficient.
\end{abstract}
\section{Introduction}

Parameter-efficient fine-tuning (PEFT) has become the standard approach for adapting large pre-trained foundation models~\cite{vaswani2017attention,dosovitskiy2020image,team2025geminifamilyhighly,singh2025openaigpt5system,ai2025deepseekv32pushingfrontier,qwen2023report} to downstream tasks. Among existing methods, low-rank adaptation (LoRA)~\citep{hu2022lora} is the most widely adopted, and a growing body of research work seeks to improve and extend it~\cite{meng2024pissa,hayou2024loraplusefficient,reploara25truong,chen2024quanta,albert2025highereffectiverank,zhang2024spectraladapter}. Yet, despite its practical success, LoRA suffers from two structural limitations that grow more severe as the rank increases.

\textbf{Optimization pathology from gauge freedom.} Any factorization $\Delta\rmW = \rmB\rmA$ is invariant under the reparameterization $(\rmB\rmC)(\rmC^{-1}\rmA)$ for any invertible $\rmC$, defining an $r^2$-dimensional manifold of equivalent solutions, which is not restricted to rescaling individual latent channels, but contains rotations and shears mixing them.  This invariance introduces flat directions in the loss landscape that are poorly handled by adaptive optimizers such as Adam~\citep{kingma2017adam,ilya2019adamw}, whose per-coordinate preconditioner implicitly assumes axis-aligned curvature. The resulting symptoms are rank collapse and growing hyperparameter sensitivity at higher ranks~\citep{biderman2024loralearnsforgets}. 

\textbf{Linear cost for limited expressivity.} A rank-$r$ LoRA adapter spans only $r$ independent directions of weight change. When a task requires richer updates, increasing $r$ scales the parameter count \emph{linearly} without a commensurate gain in structural expressivity. Recent high-rank variants~\cite{wang2024kasaknowledgeawaresingularvalue,chen2024quanta,gurses2025diablodiagonalblocks} attempt to bridge this gap, but typically incur their own costs including materializing large intermediate matrices~\cite{huang2025hira} or relying on deep tensor chains~\cite{chen2024quanta}.

In this paper, we revisit sparse structured matrices for PEFT through a generalization of the Kronecker product. The key observation is that $\rmA \otimes \rmB$ can be rewritten as the product of two block-diagonal matrices interleaved by a perfect-shuffle permutation. Relaxing the shared-block constraint of this factorization yields a strictly richer family of updates, which we call the \emph{\repname} family, and which forms the foundation of \emph{\underline{S}c\underline{a}lable \underline{M}orpho \underline{A}daptation (\ftmt)}. The adapter rank in \ftmt is governed by the block sizes, allowing it to reach full rank while remaining sparse throughout.
Crucially, we prove that the gauge group of \ftmt collapses from $GL(r)$-group invariance to a group of \emph{diagonal scalings} that acts on each channel separately. This thereby eliminates the pathological flat directions that hinder Adam-style optimizers. Empirically, we evaluate \ftmt on natural language reasoning benchmarks—including reading comprehension and arithmetic reasoning. Our method consistently outperforms strong baselines across model families and parameter budgets.

Our contributions are threefold:
\textbf{(1)} We propose \ftmt, a PEFT method derived from a structural generalization of the Kronecker product via \repname block-diagonal factorization, supporting a broad spectrum of ranks from low to full rank with high parameter efficiency.
\textbf{(2)} We analyze the rank, scaling, and gauge properties of \ftmt, showing \emph{quadratic} rank growth with respect to the parameter count, stable scaling behavior, and a diagonal invariant manifold that is inherently more optimizer-friendly than that of LoRA, contains no channel-mixing transformations.
\textbf{(3)} We demonstrate strong empirical results across diverse reasoning tasks, showing that the structural expressivity of \ftmt offers a robust and favorable alternative to existing PEFT methods.

% fram
\newtcolorbox{myquestion}{
    colback=gray!5!white, % background
    colframe=gray!75!black, % outliner
    fonttitle=\bfseries,
    title=Research Questions,
    arc=2pt, % corner
}
\section{Methodology}
\label{sec_method}

This section reviews the Kronecker product and Monarch parameterization,
establishes a structural connection between them, and shows how that
connection motivates the design of \repname adapters. Building on this
formulation, we provide theoretical analyses of rank flexibility, scaling
stability, and parameter efficiency, followed by a discussion of practical
implementation.

\begin{figure}[th]
    \centering
    \includegraphics[width=0.95\textwidth]{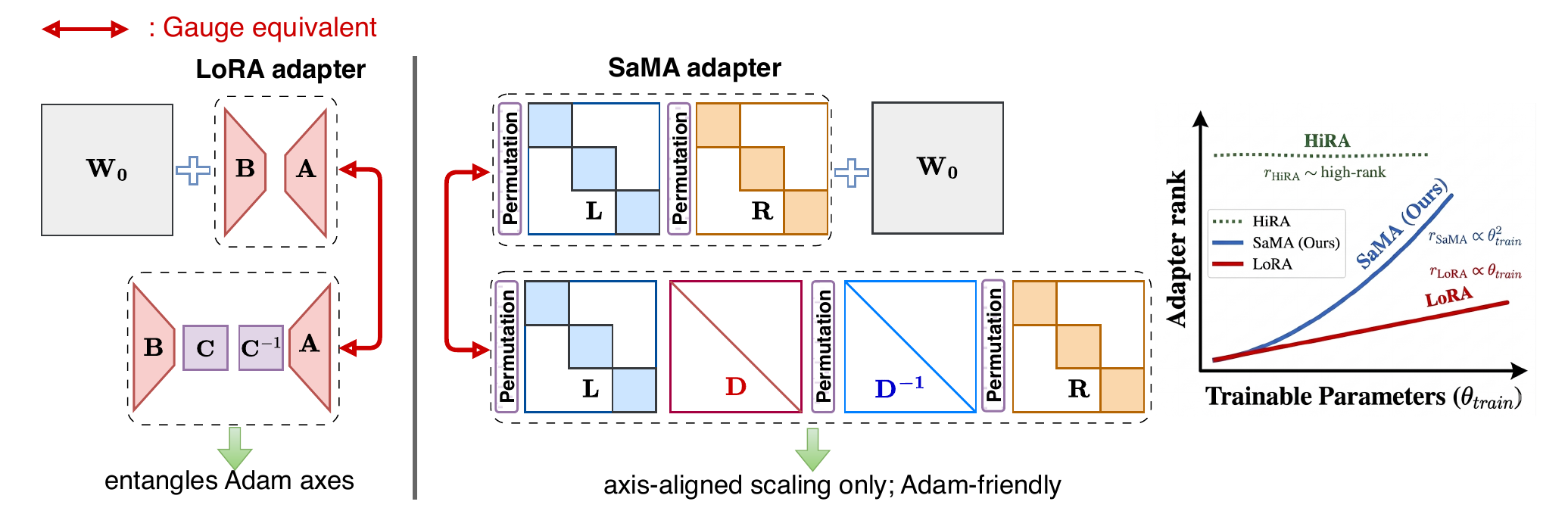}
    \vspace{5pt}
    \caption{Gauge symmetry in LoRA vs.\ \repname adapter
    parameterizations. \emph{(Left)} LoRA's update $\Delta\rmW = \rmB\rmA$
    admits any invertible $\rmC \in GL(r)$ as a gauge transform, including
    rotations and shearing. \emph{(Center)} The block-diagonal factorization
    $\mP_l^\top\rmL\mP_r\rmR$ reduces the gauge freedom to diagonal
    rescalings only: $\rmL \leftarrow \rmL\rmD$ and
    $\rmR \leftarrow \mP_r^\top\rmD^{-1}\mP_r\rmR$ for diagonal $\rmD$.
    Unlike LoRA, this coordinate-aligned scaling does not entangle
    latent features. \emph{(Right)} Expressivity: the rank of a \ftmt adapter scales as a quadratic function of the training budget.}
    \label{fig:main2}
\end{figure}

%────────────────────────────────────────────────
\subsection{Kronecker Product}
\label{sec_kronecker}
%────────────────────────────────────────────────

\begin{minipage}{0.53\textwidth}
The Kronecker product~\citep{kron_prod}, denoted $\otimes$, takes two
matrices $\rmA \in \sR^{m_1 \times n_1}$ and
$\rmB \in \sR^{m_2 \times n_2}$ and produces
$\rmC = \rmA \otimes \rmB \in \sR^{M \times N}$,
where $M = m_1 m_2$ and $N = n_1 n_2$:
\end{minipage}
\hfill
\begin{minipage}{0.44\textwidth}
    \begin{equation*}
    \rmC = \begin{bmatrix}
        a_{11}\rmB & \cdots & a_{1n_1}\rmB \\
        \vdots     & \ddots & \vdots       \\
        a_{m_1 1}\rmB & \cdots & a_{m_1 n_1}\rmB
    \end{bmatrix}
    \end{equation*}
\end{minipage}

\noindent
For an input vector $\rvx \in \sR^{N}$, define the matrix views
$\rmX = \unvecop(\rvx)$ and $\rmY = \unvecop(\rvy)$, where $\vecop(\cdot)$
stacks columns into a vector and $\unvecop(\cdot)$ reverses this operation.
Applying the Kronecker product as a linear map admits the equivalent
forms:
\begin{equation}
    \rvy = (\rmA \otimes \rmB)\,\rvx
    \;\Leftrightarrow\;
    \rmY = \rmB\,\rmX\,\rmA^\top
    \;\Leftrightarrow\;
    \rvy = \vecop\!\bigl(\rmB\,\unvecop(\rvx)\,\rmA^\top\bigr).
    \label{eq_kr_matrix_form}
\end{equation}
The matrix form in \cref{eq_kr_matrix_form} avoids materializing the full $M \times N$ product and is efficient in practice.

%────────────────────────────────────────────────
\subsection{Monarch Parameterization for Square Matrices}
\label{sec_square_monarch}
%────────────────────────────────────────────────

\begin{definition}{\citep{dao2022monarch}}
Let $N = n^2$. A \emph{square Monarch matrix} $\rmM \in \sR^{N \times N}$
has the form:
\[
    \rmM = \mP^\top \rmL\, \mP \rmR,
\]
where $\rmL = \diag(\rmL_1, \ldots, \rmL_n)$ and
$\rmR = \diag(\rmR_1, \ldots, \rmR_n)$ are block-diagonal matrices each
comprising $n$ dense blocks of size $n \times n$, and $\mP$ is the
permutation matrix that transposes the two indices of an $n \times n$
matrix obtained by reshaping a vector of length $N$.
\end{definition}

Monarch parameterization~\citep{dao2022monarch} is an extension of butterfly matrices~\citep{tridao2019fastltf,Bailey1990FFTs}, representing a matrix as the product of two block-diagonal matrices interleaved by permutations.

%────────────────────────────────────────────────
\subsection{\repname: Bridging the Kronecker Product and Monarch Parameterization}
\label{sec:method_sama_extension}
%────────────────────────────────────────────────

We now introduce the \emph{\repname matrix}, a generalization of both the
Kronecker product and Monarch matrices, designed for scalable, omni-rank
PEFT.

\subsubsection{Kronecker Product as a Two-Stage Sparse Factorization}
\label{sec_sama_kr}

We analyze the structure of $\rmA \otimes \rmB$ by decomposing it into two
sequential sparse transformations. Applying the mixed-product property of
the Kronecker product~\citep{vanloan2000ubiquitous,zhang2013kronecker},
$(\rmA \otimes \rmE)(\rmF \otimes \rmB) = (\rmA\rmF) \otimes (\rmE\rmB)$,
and setting $\rmE = \rmI_{m_2}$, $\rmF = \rmI_{n_1}$ yields:
\begin{equation}
    \rmA \otimes \rmB
    =\underbrace{(\rmA \otimes \rmI_{m_2})}_{\tL'}
     \underbrace{(\rmI_{n_1} \otimes \rmB)}_{\tR}.
    \label{eq_kr_hidden_dim}
\end{equation}
This reveals a two-stage structure: the right factor $\tR$ maps the input
from dimension $N = n_1 n_2$ to a hidden dimension $n_1 m_2$, and the left
factor $\tL'$ maps the result to the output dimension $M = m_1 m_2$.

\begin{wrapfigure}{r}{0.35\textwidth}
    \centering
    \vspace{-18pt}
    \begin{equation*}
        \tL' = \begin{bmatrix}
        \rmD_{11} & \ldots  & \rmD_{1m_2} \\
        \vdots    & \ddots  & \vdots      \\
        \rmD_{m_21} & \ldots & \rmD_{m_2m_2}
        \end{bmatrix}
    \end{equation*}
    \vspace{-15pt}
\end{wrapfigure}

Concretely, $\tR = \rmI_{n_1} \otimes \rmB = \diag(\rmB, \ldots, \rmB)$
is block-diagonal with $n_1$ identical copies of $\rmB$. The left factor
$\tL'$, however, is \emph{not} block-diagonal: its $m_1 \times n_1$ blocks
of size $m_2 \times m_2$ are scaled identities (with scalar entries drawn
from $\rmA$), as shown above.

To expose the hidden block-diagonal structure of $\tL'$, we apply the
\emph{perfect shuffle} permutation~\citep{vanloan2000ubiquitous}
(see \cref{app:math_notation}), which reshapes a vector of length $r$
into a $p \times q$ matrix in row-major order, transposes it, and
re-vectorizes. Using the identity
$\rmA \otimes \rmI_{m_2} = \mP_{m_1,m_2}^\top\,(\rmI_{m_2}\otimes\rmA)\,\mP_{n_1,m_2}$,
we rewrite:
\begin{equation}
    \tL' = \mP_l^\top\,\rmL\,\mP_r,
    \qquad
    \rmL = \diag(\rmA, \ldots, \rmA),
    \quad
    \mP_l^\top = \mP_{m_1,m_2}^\top,
    \quad
    \mP_r = \mP_{n_1,m_2}.
    \label{eq_left_rewrite}
\end{equation}
\Cref{eq_left_rewrite} reveals that both factors of the Kronecker product
share the same block-diagonal structure, up to a pair of perfect shuffle
permutations. This symmetric view directly motivates the generalization
introduced in the next section.
\subsubsection{Generalizing the Kronecker Decomposition}
\label{sec_sama_expand}

In the Kronecker product, all diagonal blocks within each factor are
\emph{tied} — every block equals either $\rmA$ or $\rmB$. Releasing this
constraint and allowing each block to be an \emph{independent} dense matrix
yields a strictly richer family of transformations while preserving the
computational advantages of block-diagonal structure.

\textbf{Fine-tuning via reparameterization.}\enspace
Given a frozen pre-trained weight $\rmW_0 \in \sR^{M \times N}$, we
introduce a trainable additive update $\Delta\rmW$ and define the
fine-tuned forward pass as:
\begin{equation*}
    \rmW_\mathrm{ft} = \rmW_0 + \Delta\rmW,
    \qquad
    \rvy = \rmW_0\,\rvx + \Delta\rmW\,\rvx.
\end{equation*}
The design of $\Delta\rmW$ is the central contribution of \ftmt.

\paragraph{Scalable Morpho Adaptation (\repname).}
Motivated by the Kronecker decomposition in \cref{sec_sama_kr} and the
Monarch structure in \cref{sec_square_monarch}, we parameterize the update
as a \emph{\repname matrix}:
\begin{equation}
    \Delta\rmW = \evs_{\ftmrm}\,\mP_l^\top\,\rmL\,\mP_r\,\rmR,
    \label{eq_sama_delta}
\end{equation}
where $\rmL$ and $\rmR$ are block-diagonal matrices with \emph{independent}
dense blocks,
\begin{equation}
    \rmL = \diag(\rmL_1, \ldots, \rmL_{m_2}),
    \qquad
    \rmR = \diag(\rmR_1, \ldots, \rmR_{n_1}),
    \label{eq:method_def_LR}
\end{equation}
with $\rmL_i \in \sR^{m_1 \times n_1}$ and $\rmR_j \in \sR^{m_2 \times n_2}$
each independently parameterized, and $n_1, m_2 \ll \min(M, N)$. The
permutations $\mP_l = \mP_{m_1,m_2}$ and $\mP_r = \mP_{n_1,m_2}$ are
fixed perfect shuffles (see \cref{eq_shuffle}) and contribute no trainable
parameters. The scalar $\evs_{\ftmrm}$ controls the update magnitude; its
role in training stability is analyzed in \cref{sec:theo:scaling_stability}.
We index a specific adapter configuration by the tuple $(n_1, m_2)$,
written $\ftmrm_{n_1 \times m_2}$.

We denote the set of all matrices expressible in the form of
\cref{eq_sama_delta} as $\mset^{(M,N)}$ (abbreviated $\mset^{(N)}$ for
square matrices), and the set of all compatible factor pairs $(\rmL, \rmR)$
satisfying \cref{eq:method_def_LR} as $\mset_g$.

\textbf{Remark.}\enspace
The permutation $\mP_{n_1,n_2}$ maps a vector $[x_1, \ldots, x_N]$ to
$[x_1, x_{1+n_2}, \ldots,\allowbreak x_{1+(n_1-1)n_2},\;\allowbreak x_2, x_{2+n_2}, \ldots,\;
x_{n_2}, x_{2n_2}, \ldots, x_{n_1 n_2}]$,
equivalent to reading an $n_1 \times n_2$ matrix in column-major order
after writing it in row-major order. When $M = N = n^2$ and all blocks are
square, \cref{eq_sama_delta} reduces exactly to the Monarch parameterization
of \cref{sec_square_monarch}.

\textbf{Discussion.}\enspace
\ftmt differs fundamentally from Dense-to-Sparse fine-tuning
(D2S-FT)~\citep{dao2022monarch}, which applies Monarch \emph{square
matrices} in pre-training and requires projecting weights into sparse form
before training. \ftmt instead operates purely in the additive adapter
regime, requiring no such pre-processing. Critically, it spans low-rank to
full-rank updates within a single parameterization, addressing both the
low intrinsic dimensionality of fine-tuning
landscapes~\citep{chunyuan2018intrisicdim,aghajanyan2021intrinsic} and the
broader need for omni-rank adaptation which establishes a new baseline
alongside the LoRA family~\citep{hu2022lora,liu2024dora,meng2024pissa} and
recent high-rank methods~\citep{chen2024quanta,he2025smt}.

\textbf{Comparison with block-diagonal methods.}\enspace
\ftmt is architecturally distinct from MoRE~\citep{tan2025more} and similar
block-diagonal adapters. MoRE partitions the weight matrix spatially into $b_M$
decoupled blocks, in both left and right matrices, like a collection of low rank adapter, results in a risk of sparse adapter. \ftmt is instead constructed via the extended Kronecker product: the factor
dimensions $(n_1, m_2)$ are not spatial slices but intrinsically coupled
tensor indices, influencing block counts. This mutually dependent structure guarantees dense output and only allow diagonal entanglement in parameter space degeneracy. (See \cref{sec:app_more_dist})

\ftmt follows the reparameterization convention of LoRA~\citep{hu2022lora} that
$\rmR$ is initialized from a small-variance Gaussian and $\rmL = 0$,
ensuring $\Delta\rmW = 0$ at the start of training with non-vanishing
gradients. At inference time, the adapter merges into $\rmW_0$ with zero
overhead, and forward passes through $\Delta\rmW$ during training cost
$O(\nu N)$ FLOPs via two batched block-matrix multiplies — with no
Dense-to-Sparse pre-processing required. Full implementation details are
given in \cref{sec:method_prac_regime}.
\section{Theoretical Analysis of \ftmt}

This section examines several aspects of \ftmt, including rank versatility in different regimes, stability, gauge symmetry in parameter space and expressivity, then establishes its theoretical properties.

% ==============================================================
\subsection{Rank Properties}
\label{sec:method:rank_property}
% ==============================================================

We establish three theoretical aspects of \ftmt including rank flexibility under random initialization, scaling stability as hidden dimension grows, and convergence to the full fine-tuning solution under mild conditions.

\begin{theorem}[Rank upper bound]
\label{theo_sama_rank}
Let $\rmL$ and $\rmR$ be block-diagonal matrices as defined in \cref{eq:method_def_LR},
with $n_1 m_2 \leq \min(M, N)$, and let $\evs \neq 0$.
Suppose every entry of each submatrix is drawn i.i.d.\ from a continuous distribution.
Then, with probability $1$:
\begin{equation}
    \rankop(\Delta\rmW)
    = \rankop\!\left(\evs\,\mP_l^\top\rmL\mP_r\rmR\right)
    = n_1 m_2.
    \label{main:eq_theo_rank}
\end{equation}
\end{theorem}

\Cref{theo_sama_rank} shows that the effective rank of the adapter is exactly $n_1 m_2$,
determined jointly by the block counts of $\rmL$ and $\rmR$.
This decoupling is structurally significant: by varying $n_1$ and $m_2$ independently,
\ftmt covers a broad range of rank regimes, from low-rank to full-rank adaptation.
Notably, for a square $N\times N$ weight with $N$ a perfect square, a full-rank update is achievable with only $2N\sqrt{N}$ parameters ($n_1=m_2=\sqrt{N}$).

To make the comparison concrete, consider a fixed parameter budget of $P = 2Nr$ shared by both methods. LoRA allocates this as two matrices of sizes $N \times r$ and $r \times M$, yielding adapter rank $\gR_{\text{LoRA}} = r$, which grows \emph{linearly} with $P$. \ftmt with configuration $(n_1, m_2) = (r, r)$ matches the same budget while achieving adapter rank $\gR_{\ftmt} = n_1 m_2 = r^2$, growing \emph{quadratically}.

% ==============================================================
\subsection{Expressivity of \repname}
\label{sec:sim_exp_exp}
% ==============================================================

The Eckart-Young theorem~\citep{Eckart1936TheAO} establishes rank-$r$ matrices as the \emph{optimal} Frobenius approximators of any target, given a budget of $r$ singular components.
This optimality is the cornerstone to understand the expressivity of LoRA and its variants. Here, we study the following question:
\emph{"Under what conditions can a higher-rank structure, specifically Morpho, surpasses the Eckart-Young bound at the same parameter budget?"}

\begin{theorem}[Expressivity crossover]
\label{theo:express_approx}
Let $\rmW \in \sR^{d \times d}$ be a rank-$R$ matrix with exactly $R$ unit singular values,
and fix a parameter budget corresponding to $n_1 = m_2 = r < R$ in \cref{eq_sama_delta}.
Denote $\rmW^*_{\textrm{LoRA}}$ as the optimal rank-$r$ approximation of $\rmW$ (the Eckart--Young truncation),
and $\rmW^*_{\textrm{Morpho}}$ as the \repname approximation of $\rmW$, under the same budget.
Then there exists a threshold $R_0 < d$ such that, with probability $1$:
\begin{equation}
    \|\rmW - \rmW^*_{\textrm{Morpho}}\|_F \;<\; \|\rmW - \rmW^*_{\textrm{LoRA}}\|_F, \qquad \forall\, R > R_0.
\end{equation}
\end{theorem}

\Cref{theo:express_approx} identifies \emph{effective rank} (eRank)~\citep{roy2007effective} as the key discriminating quantity.
At low eRank, $\rmW$'s spectrum is concentrated and Eckart--Young captures it nearly perfectly, whereas \repname offers no benefit.
At high eRank, the spectral mass of $\rmW$ is broadly distributed and the rank-$r$ ceiling becomes binding.  The \repname structure covers a higher-dimensional subspace than the LoRA bottleneck because of its global mixing between diagonal blocks. The proof of \cref{theo:express_approx} can be found in Appendix~\ref{sec:app_proof_express}.

\paragraph{Empirical validation.}

\begin{wrapfigure}{r}{0.43\textwidth}
  \centering
  \vspace{-13pt}
  \includegraphics[width=0.40\textwidth]{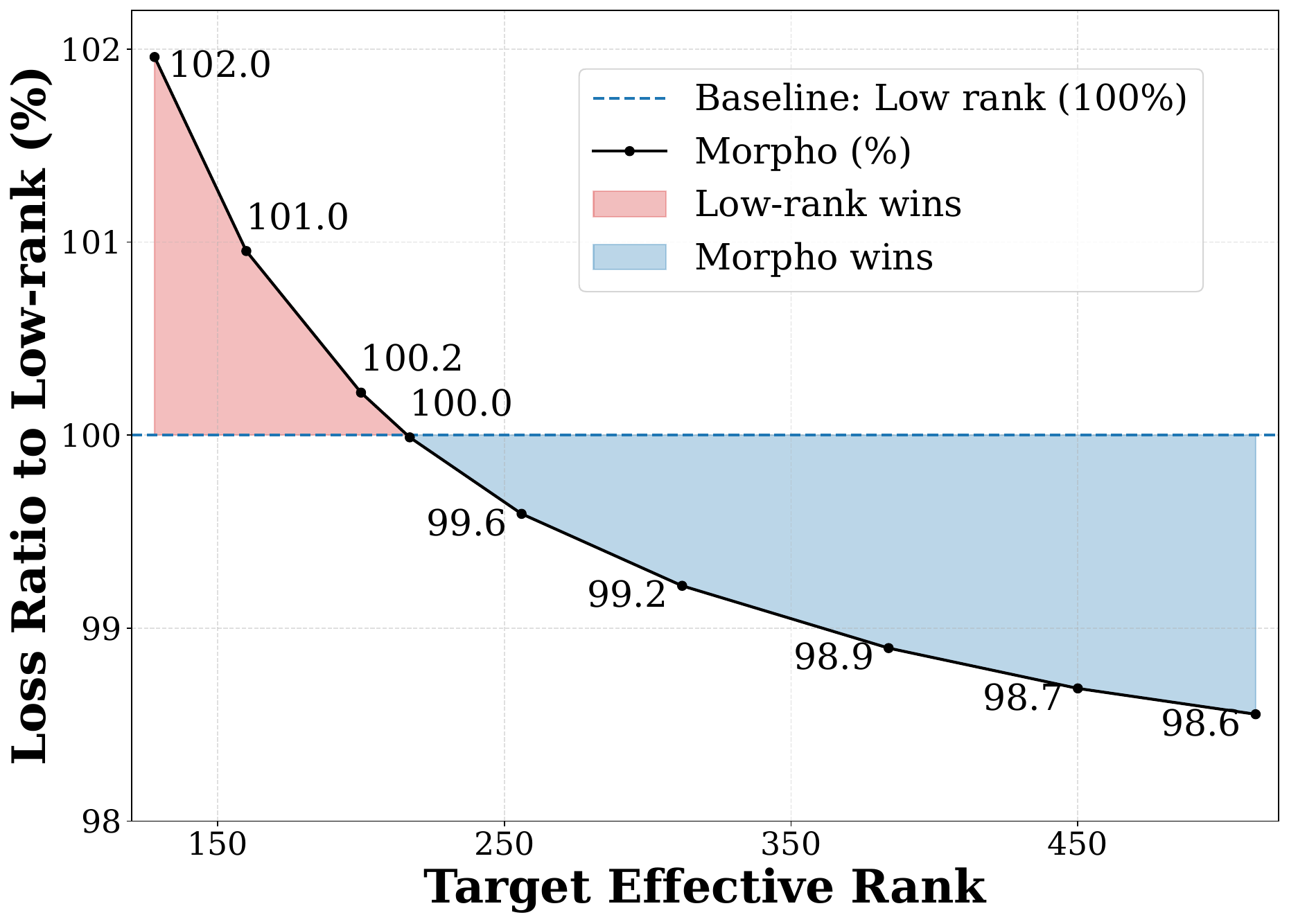}
  \caption{Relative approximation loss of \repname vs.\ low-rank under matched parameter budgets. \repname gains a decisive advantage beyond eRank $= 217$.}
  \label{fig:main_express_2bl}
  \vspace{0pt}
\end{wrapfigure}

To justify Theorem~\ref{theo:express_approx}, we design a controlled matrix-approximation stress test. For each parameterization class $\gS_i$, we solve
$\min_{\rmW' \in \gS_i} \|\rmW - \rmW'\|_F$
on synthetic targets with systematically varied eRank.
This setup deliberately isolates the approximation geometry from task signals, cleanly measuring how each parameterization covers the target manifold as the spectrum shifts. We deliberately design target matrices of size $1024 \times 1024$ with low entropy effective rank $r$, e.g. $\erankop(\rmW) = \rankop(\rmW)$, to avoid spectrum decay in the low-rank target, then evaluate the performance on matrix approximation of LoRA and \ftmt across a wide range of effective ranks. (See details in \cref{sec:app_detail_express}.)

The results in \cref{fig:main_express_2bl} confirm the theoretical prediction.
At low eRank, low-rank approximation is competitive or superior, consistent with Eckart--Young.
Beyond eRank $= 217$, \repname achieves strictly lower approximation loss under the same budget, with the gap widening monotonically. This result is precisely anticipated by \cref{theo:express_approx}.

% --------------------------------------------------------------
\subsection{Scaling stability}
\label{sec:theo:scaling_stability}
% --------------------------------------------------------------

\begin{theorem}[Numerical stability] % of the scaling factor
\label{main:theorem_rank_stable}
The scaling factor $\evs_{\ftmrm}$ that keeps the initialization variance of $\Delta\rmW$ stable after finite $n$ SGD updates,
as the hidden dimension $n_1$ grows satisfies:
\begin{equation}
    \evs_{\ftmrm} = \alpha\, n_1^{\nicefrac{1}{4}}
    \quad\Longleftrightarrow\quad
    \Delta\rmW = \left(\alpha\, n_1^{\nicefrac{1}{4}}\right)\mP_l^\top\rmL\mP_r\rmR,
    \qquad \alpha \in \sRs.
    \label{eq:theorem_rank_stable}
\end{equation}
\end{theorem}

\Cref{eq:theorem_rank_stable} derives the scaling factor (unique up to a constant factor) that simultaneously stabilizes forward output variance and backward gradient magnitude of the trained adapter as $n_1$ grows. (See \cref{sec:app_rank_stable} for the proof and the detailed explanation.)
Using all other factors can lead the signal to either attenuate or diverge (empirically verified in \cref{sec:app_rank_stable_exp}).
Critically, this result is theoretically \emph{derived}; unlike rsLoRA~\citep{dzievski2023rslora}, which proposes $\nicefrac{\alpha}{\sqrt{r}}$ empirically. To our knowledge, no existing block-structured adapter including MoRE~\cite{tan2025more} and QuanTA~\cite{chen2024quanta} provides an analogous stability analysis.

\subsection{Optimization dynamics in structured invariant manifolds}
\label{sec:express_repara}

We also explore optimization behavior of structured adapter parameterizations. Particularly, we investigate how the geometry of invariant manifolds affects parameter space degeneracy and builds a connection with optimization. 

\textbf{Gauge symmetry in matrix factorization} Any factorization $\Delta \rmW = \rmB \rmA$ admits an infinite family of \emph{equivalent} solutions. Inserting an invertible matrix $\rmC \in GL(r)$ yields $\Delta \rmW = (\rmB \rmC)(\rmC^{-1} \rmA)$ with identical loss. This defines an \emph{invariant manifold} indicating a continuous symmetry group under which the optimization objective is flat. For LoRA-style adapters, this manifold spans the full general linear group $GL(r)$, including rotations and shearing transformations in latent space.

\textbf{The optimization bottleneck in invariance manifolds.}
{Adaptive optimizers such as Adam~\citep{kingma2017adam,ilya2019adamw} operate coordinate-wise, implicitly assuming axis-aligned curvature. Rotational freedom in $GL(r)$ violates this assumption. That is, any rotation of the latent basis linearly recombines parameter columns of $\rmB$ and $\rmA$, entangles features that Adam treats as independent~\citep{depavia2025how}. Therefore, Adam cannot efficiently navigate the flat directions carved out by feature-mixing transformations. 
}

We will show that Morpho parameterization does not suffer from this degeneracy, as shown by analyzing the invariant manifold of SaMA.

\begin{theorem}[Invariance manifold of \repname parameterization]
    \label{theorem:main_reparameterization_low_hidden_rank}
    Given a generic matrix $\rmM_{1} \in \mset: \enspace \rmM_1 = \mP_l^\top \rmL_1 \mP_r \rmR_1$, where $\rmL_{1},\rmR_{1}$ are determined block-diagonal matrices.
Let $\mset _g(\rmM_1)$ denote a set of generative factor pairs for $\rmM_1$. If the hidden size $q=n_1m_2 < \min{(M,N)}$, then:
\begin{equation}
    \mset _g(\rmM_1) \equiv \Bigl\{ (\rmL,\rmR) \mid \rmL = \rmL_1 \mD, \,\rmR = \mP_r^\top \mD^{-1} \mP_r \rmR_1;  \mD = \diag{(\vv)} \in \gB\gD, \vv \in (\sR \backslash \{0\} ) ^q \Bigr\}.
\end{equation}
\end{theorem}

\Cref{theorem:main_reparameterization_low_hidden_rank} shows that \ftmt's invariance manifolds belong to the diagonal subgroups $\gB\gD$. This is the consequence of the block-diagonal topology in $\rmL$ and $\rmR$ since dense rotations and shearing are not allowed. The proof can be found in Appendix~\ref{sec:app_repara_morpho}.

\textbf{Implication.} Pure scaling along coordinate axes neither mixes latent features nor disrupts Adam's axis-aligned preconditioner. Cross-feature entanglement is intrinsically suppressed by design, not by external regularization. The block-diagonal structure thus acts as an \emph{implicit structural preconditioner}.
It eliminates channel-mixing components involved in pathological flat directions arising in dense reparameterization, thereby reshaping the loss landscape into a form that adaptive optimizers can traverse more efficiently.
Further discussion of parameter-space degeneracy is provided in \cref{sec:app_repara}.

\subsubsection{Empirical verification of gauge structure.}
To empirically verify the theoretical distinction between LoRA and \repname, we conduct toy experiments on a matrix approximation task.

\begin{figure}[h]
    \centering
  \vspace{-5pt}
  \begin{tikzpicture}
 
  \node at (0,0) {\includegraphics[width=0.35\textwidth]{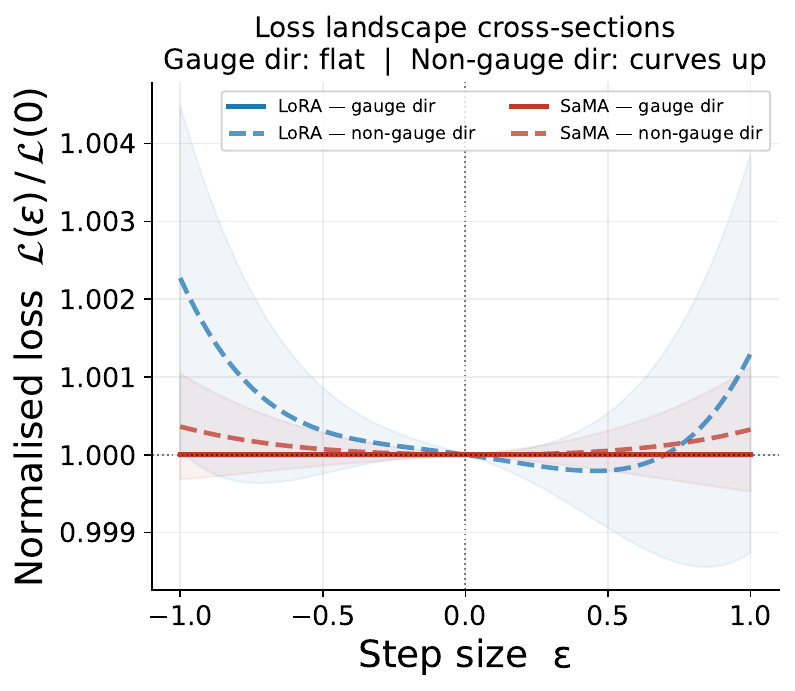}};
  \node at (5, 0) {\includegraphics[width=0.35\textwidth]{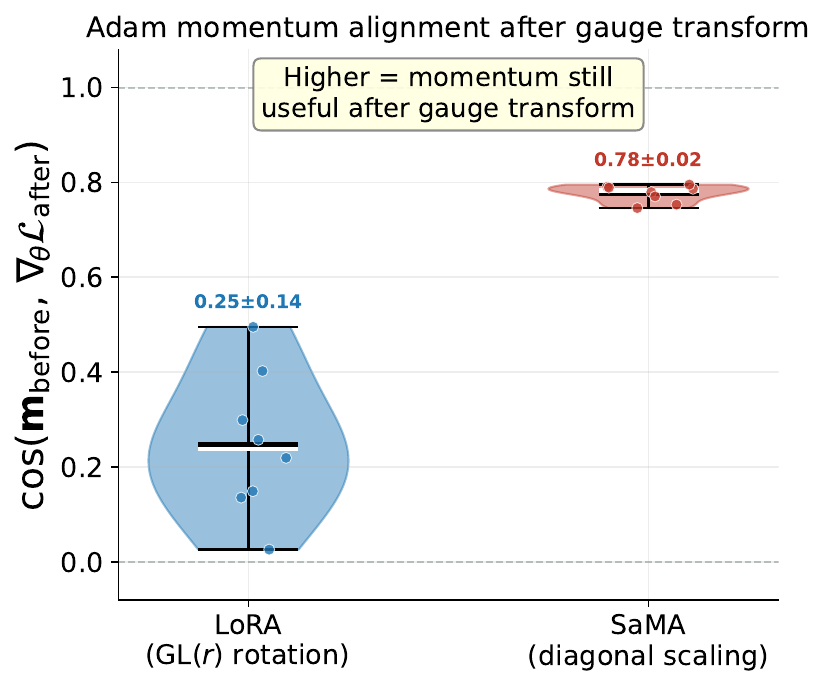}};

  \node at (0, -2.2) {(a)};

  \node at (5, -2.2) {(b)};

  \end{tikzpicture}
  
  \vspace{3pt}
  \caption{(a) Loss landscape flatness between LoRA and SaMA.  (b) Cosine similarity between pre-transform Adam momentum $\vm_{\mathrm{before}}$ and post-transform gradient $\nabla_{\!\theta}\mathcal{L}_{\mathrm{after}}$.}
  \label{fig:momentum_aligment}
  \vspace{-5pt}
    
\end{figure}

\textbf{Loss landscape flatness.}
We partially trained both methods and scanned the normalized loss $\mathcal{L}(\theta + \varepsilon \mathbf{d})\,/\,\mathcal{L}(\theta)$
along two directions: a gauge direction $\mathbf{d}_{\mathrm{gauge}}$, which leaves
$\Delta\mathbf{W}$ exactly unchanged, and a random non-gauge direction.
As shown in~\cref{fig:momentum_aligment}(a), both gauge directions produce
exactly flat loss curves (normalised loss $\equiv 1.0$), confirming the theoretical flat directions.
Crucially, the non-gauge direction of \repname remains nearly flat whereas LoRA's non-gauge
direction exhibits significantly larger curvature.
This indicates that the surrounding loss landscape of \repname is easy to
navigate.

\textbf{Momentum alignment after gauge transform.}
To directly measure the impact of gauge redundancy on adaptive optimizers,
we train each adapter for 400 warm-up steps with Adam to build meaningful
first-moment estimates $\mathbf{m}$. We then apply a gauge transform that
leaves $\Delta\mathbf{W}$ unchanged but modifies the parameter coordinates:
a random permutation matrix $\mathbf{C}$ for LoRA, and an alternating
diagonal scaling for \repname.
We then measure the cosine similarity between the pre-transform momentum
$\mathbf{m}_{\mathrm{before}}$ and the post-transform gradient
$\nabla_{\!\theta}\mathcal{L}_{\mathrm{after}}$, which quantifies how much of
Adam's accumulated directional information remains useful after the
parameter coordinates are rearranged.

As shown in~\cref{fig:momentum_aligment}(b), LoRA yields a cosine similarity
of $0.25 \pm 0.14$ indicating that the rotational freedom in
$\mathrm{GL}(r)$ effectively scrambles Adam's momentum into an uninformative
direction.
In contrast, \repname achieves $0.78 \pm 0.02$ with near-zero variance across seeds,
confirming that diagonal scalings merely rescale the per-coordinate momentum
without mixing features, leaving the optimizer state largely intact.
These results provide direct empirical support for the claim that the $\mathrm{GL}(r)$
gauge freedom of LoRA creates an optimization bottleneck that \repname's
diagonal gauge structure inherently avoids.

\section{Experimental Results}
To evaluate \ftmt against other PEFT methods, we conduct experiments across a broad spectrum of downstream tasks. Specifically, we verify \ftmt effectiveness on three \emph{complex reasoning} domains, namely DROP, Commonsense (CMS), and Arithmetic.
Additional experiments, ablation studies, details of the datasets and hyperparameters are shown in \cref{app_dataset_infor,app_exps,sec:app_further_exps,sec:app_ablation}, respectively.

\textbf{Main baselines:} Across these experiments, we compare \ftmt against a range of strong recent PEFT baselines, including LoRA~\citep{hu2022lora}, PiSSA~\citep{meng2024pissa}, ABBA~\citep{singhal2025abba}, QuanTA~\citep{chen2024quanta}, LoRA-Pro~\citep{wang2024loraproarelowrank} and SMT~\citep{he2025smt}, with full fine-tuning (FFT) included as a reference.

\subsection{DROP Dataset}
\label{sec:exp_drop}
\begin{wrapfigure}{r}{0.42\textwidth}
  \centering
  \vspace{-11pt}
  \includegraphics[width=0.39\textwidth]{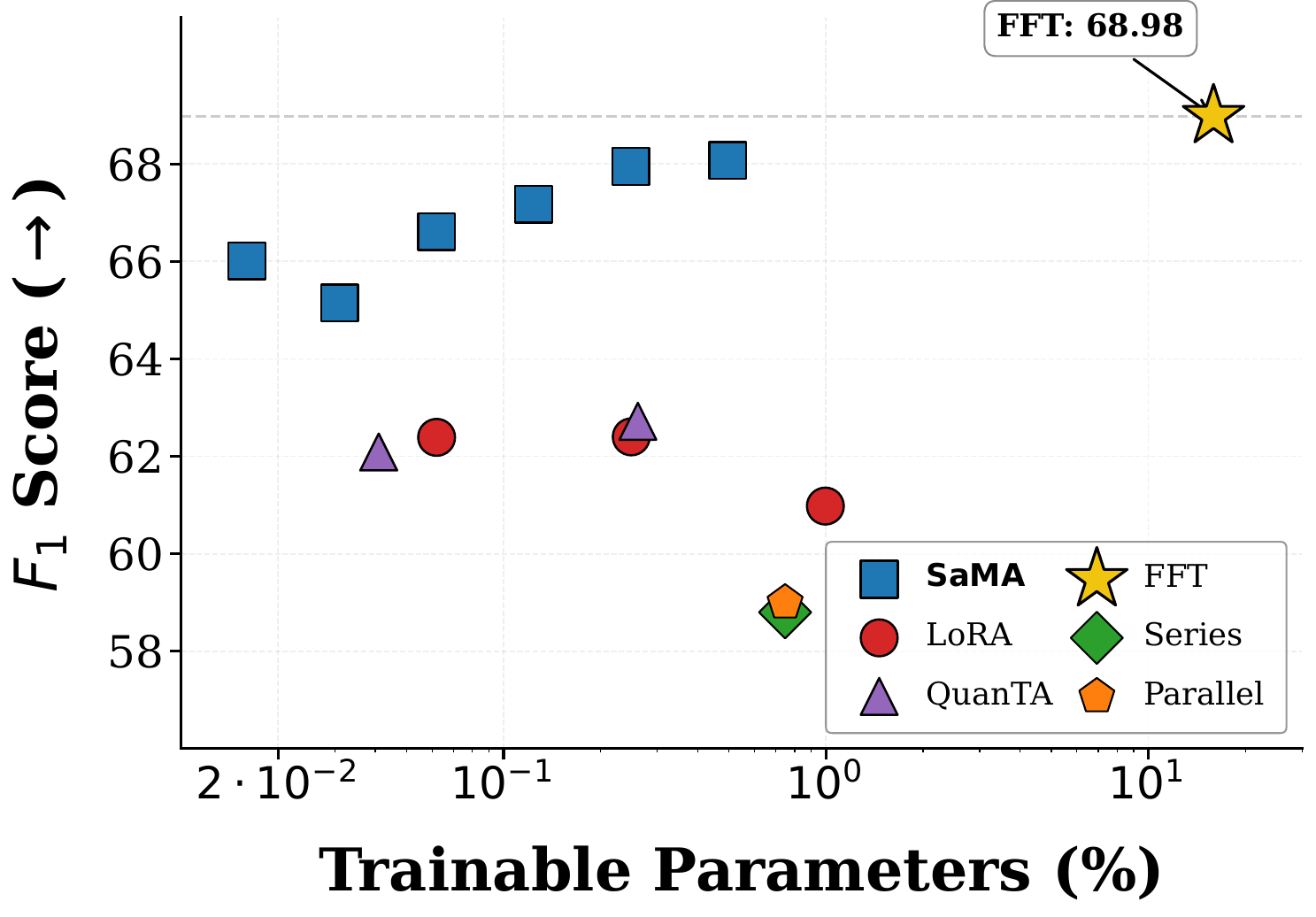}
  \caption{\label{fig:drop_chart}Benchmark of different FT methods on the DROP dataset, using Llama2-7B, as parameter budgets vary.}%that leverage batch matrix multiply and memory efficiency}
  % text beneath
  \vspace{0pt}
\end{wrapfigure}

Following QuanTA~\citep{chen2024quanta}, in order to explicitly clarify the effectiveness of these adapters, we conduct our initial experiments using Llama2-7B on the DROP dataset~\citep{dua2019drop} as a representative case study for high-rank updates. As illustrated in \cref{fig:drop_chart} and further detailed in \cref{tab:llama_drop_bench}, \ftmt consistently outperforms other existing fine-tuning methods across these models, reaching FFT (on Query and Value modules) performance. Notably, while increasing the rank to improve capacity, LoRA exhibits diminishing returns, failing to close the performance gap. Meanwhile, QuanTA -- with high-rank updates -- shows overall improvement, but still suffers from a performance lag.

These results suggest that \ftmt benefits from a mechanism that better navigates mid- and high-rank regimes. By leveraging sparse structures, \ftmt facilitates a favorable optimization landscape, enabling a smoother optimization trajectory compared to dense high-rank alternatives. The omni-rank property might excel at extracting necessary features with less redundancy.

\subsection{Arithmetic reasoning}
\label{exp:arith_main}

\vspace{-12pt}
\begin{table}[htbp]
    \centering
    \small
    \tableArithMainResult
\end{table}

To examine whether \ftmt remains effective on demanding reasoning, we consider and evaluate models on MATH~\cite{hendrycks2021measuringmathematicalproblem} and GSM8K~\cite{cobbe2021gsm8k} as representative challenging benchmarks among popular arithmetic reasoning tasks. We fine-tune pretrained models of Mistral-7B~\cite{jiang2023mistral7b} and Gemma-2 9B~\cite{gemmateam2024gemma2b9} on a 20K-sample subset of MetaMathQA~\cite{yu2023metamath}.

Results on the two test sets in \cref{tab:arith_comparison_results} show a clear overview. Although all PEFT baselines adopt rank 32, equivalent to \oftmto$_{32,32}$, \ftmt consistently outperforms prior methods on both tasks while using fewer parameters; notably, even the lightweight setting $(n_1,m_2)=(8,8)$ uses 4 times fewer parameters than the baselines, yet remains highly competitive and approaches FFT. This suggests that the proposed structured parameterization provides a favorable balance between efficiency and expressivity.
We hypothesize that this effect arises by means of the structured parameterization, combined with a smoother optimization landscape.

\subsection{Commonsense reasoning}
% --- FILE GUARD ---
\ifdefined\MYCRDATALOADED
\else
    \newcommand{\MYCRDATALOADED}{}

    % --- Cr (Commonsense Reasoning) ---
    \newcommand{\myDashLine}{\addlinespace[0.3ex]\cdashline{2-12}\addlinespace[0.5ex]}
    % ChatGPT
    \newcommand{\rCrChatGPT}{ChatGPT$^\ast$ & -- & -- & 73.1 & 85.4 & 68.5 & 78.5 & 66.1 & 89.8 & 79.9 & 74.8 & 77.0 \\}

    % LLaMA-7B (CrSeven)
    \newcommand{\rCrSevenFFT}{& FFT$^\spadesuit$ & 100 & 70.3 & 84.2 & 80.1 & 92.3 & 85.4 & 86.6 & 72.8 & 83.4 & 81.9 \\} % S2FT
    \newcommand{\rCrSevenPrefix}{& Prefix$^\ast$ & 0.11 & 64.3 & 76.8 & 73.9 & 42.1 & 72.1 & 72.9 & 54.0 & 60.6 & 64.6 \\} % LLM
    \newcommand{\rCrSevenSeries}{& Series$^\ast$ & 0.99 & 63.0 & 79.2 & 76.3 & 67.9 & 75.7 & 74.5 & 57.1 & 72.4 & 70.8 \\} % LLM
    \newcommand{\rCrSevenParallel}{& Parallel$^\ast$ & 3.54 & 67.9 & 76.4 & 78.8 & 69.8 & 78.9 & 73.7 & 57.3 & 75.2 & 72.3 \\}
    \newcommand{\rCrSevenLoRA}{& LoRA$^\spadesuit$ & 0.83 & 69.2 & 81.7 & 78.4 & 83.4 & 80.8 & 79.0 & 62.4 & 78.4 & 76.7 \\} % S2FT
    \newcommand{\rCrSevenRepLoRA}{& RepLoRA$^\vee$ & 1.01 & 71.8 & 84.1 & 79.3 & 85.2 & 83.3 & 82.4 & 66.2 & 81.2 & 79.1 \\}
    \newcommand{\rCrSevenRefLoRAl}{& RefLoRA$^\bigstar$ & 0.83 & 69.6 & 82.5 & 79.5 & 88.3 & 82.6 & 81.6 & 66.6 & 80.2 & 78.9 \\} %RepLoRA
    \newcommand{\rCrSevenDoRAh}{& DoRA$_{r=16}^\circ$ & 0.43 & 70.0 & 82.6 & 79.7 & 83.2 & 80.6 & 80.6 & 65.4 & 77.6 & 77.5 \\} % DORA
    \newcommand{\rCrSevenDoRAf}{& DoRA$_{r=32}^\circ$ & 0.84 & 69.7 & 83.4 & 78.6 & 87.2 & 81.0 & 81.9 & 66.2 & 79.2 & 78.4 \\} % DoRA
    \newcommand{\rCrSevenGalore}{& Galore$^\spadesuit$ & 0.83$^\blacktriangle$ & 68.6 & 79.0 & 78.5 & 84.7 & 80.1 & 80.3 & 62.1 & 77.3 & 76.3 \\} % S2FT
    \newcommand{\rCrSevenLaX}{& LaX$_{32}^\circledast$ & 0.83 & 69.6 & 81.9 & 78.9 & 84.7 & 80.8 & 79.8 & 64.8 & 78.0 & 77.3 \\} % LoRA+LaX
    \newcommand{\rCrSevenLISA}{& LISA$^\spadesuit$ & 9.91 & 70.4 & 82.1 & 78.7 & 92.4 & 82.9 & 84.9 & 70.2 & 78.4 & 80.0 \\} % S2FT
    \newcommand{\rCrSevenSSFT}{& S$^2$FT$^\spadesuit$ & 0.81 & 72.7 & 83.7 & 79.6 & 93.4 & 83.5 & 86.1 & 72.2 & 83.4 & 81.8 \\} % S2FT
    \newcommand{\rCrSevenDiReFT}{& DiReFT$^\diamondsuit$ & 0.03 & 69.5 & 83.0 & 79.0 & 92.5 & 80.5 & 82.2 & 68.0 & 77.5 & 79.0 \\} % ReFT
    \newcommand{\rCrSevenLoReFT}{& LoReFT$^\diamondsuit$ & 0.03 & 69.3 & 84.4 & 80.3 & 93.1 & 84.2 & 83.2 & 68.2 & 78.9 & 80.2 \\} % ReFT
    \newcommand{\rCrSevenSMT}{& SMT$^\bullet$ & 0.84 & 68.7 & 81.7 & 78.3 & 91.6 & 78.8 & 84.1 & 68.7 & 77.4 & 78.7 \\} % SMT
    \newcommand{\rCrSevenSMTBest}{& SMT$^\bullet$\textit{(Best)} & 4.91 & 72.0 & 82.9 & 80.7 & 93.3 & 82.4 & 86.1 & 70.6 & 83.0 & 81.4 \\} % SMT
    \newcommand{\rCrSevenSaMAFour}{\rowcolor{tableblue} & $\textbf{SaMA}_{4 \times 4}$ & 0.104 & 69.9 & 84.4 & 81.3 & 94.0 & 83.5 & 86.8 & 71.2 & 84.9 & 82.0 \\}
    \newcommand{\rCrSevenSaMASix}{\rowcolor{tableblue} & $\textbf{SaMA}_{16 \times 16}$ & 0.416 & {71.1} & {84.9} & {81.7} & {94.1} & {84.0} & {87.6} & {72.7} & {86.2} & \textbf{82.8} \\}

    % LLaMA-13B (CrThirteen) -
    % 
    \newcommand{\rCrThirteenFFT}{& FFT$^\spadesuit$ & 100 & 74.5 & 86.3 & 81.3 & 94.4 & 86.9 & 89.7 & 77.9 & 88.8 & 85.0 \\} % S2FT
    \newcommand{\rCrThirteenPrefix}{& Prefix$^\ast$ & 0.03 & 65.3 & 75.4 & 72.1 & 55.2 & 68.6 & 79.5 & 62.9 & 68.0 & 68.4 \\}
    \newcommand{\rCrThirteenSeries}{& Series$^\ast$ & 0.80 & 71.8 & 83.0 & 79.2 & 88.1 & 82.4 & 82.5 & 67.3 & 81.8 & 79.5 \\}
    \newcommand{\rCrThirteenParallel}{& Parallel$^\ast$ & 2.89 & 72.5 & 84.8 & 79.8 & 92.1 & 84.7 & 84.2 & 71.2 & 82.4 & 81.5 \\}
    \newcommand{\rCrThirteenLoRA}{& LoRA$^\ast$ & 0.67 & 72.1 & 83.5 & 80.5 & 90.5 & 83.7 & 82.8 & 68.3 & 82.4 & 80.5 \\}
    \newcommand{\rCrThirteenRepLoRA}{& RepLoRA$^\vee$ & 0.99 & 73.1 & 85.2 & 84.7 & 91.1 & 85.9 & 84.7 & 73.4 & 85.6 & 82.9 \\}
    \newcommand{\rCrThirteenDoRAh}{& DoRA$_{r=16}^\circ$ & 0.35 & 72.5 & 85.3 & 79.9 & 90.1 & 82.9 & 82.7 & 69.7 & 83.6 & 80.8 \\}
    \newcommand{\rCrThirteenDoRAf}{& DoRA$_{r=32}^\circ$ & 0.68 & 72.4 & 84.9 & 81.5 & 92.4 & 84.2 & 84.2 & 69.6 & 82.8 & 81.5 \\}
    \newcommand{\rCrThirteenLaX}{& LaX$_{32}^\circledast$ & 0.67 & 71.3 & 85.4 & 81.3 & 91.3 & 84.1 & 84.4 & 71.8 & 83.1 & 81.6 \\} % LoRA+LaX
    \newcommand{\rCrThirteenSSFT}{& S$^2$FT$^\spadesuit$ & 0.65 & 74.2 & 85.7 & 80.7 & 94.9 & 86.4 & 88.4 & 76.3 & 87.8 & 84.3 \\} % SSFT
    \newcommand{\rCrThirteenSMT}{& SMT$^\bullet$ & 0.68 & 71.1 & 84.4 & 81.7 & 93.7 & 83.2 & 86.7 & 73.7 & 85.2 & 82.4 \\}
    \newcommand{\rCrThirteenSMTBest}{& SMT$^\bullet$\textit{(Best)} & 4.91 & 72.6 & 86.1 & 81.9 & 95.0 & 86.1 & 88.2 & 77.1 & 87.4 & 84.3 \\}
    \newcommand{\rCrThirteenDiaBlo}{& DiaBlo$_{N=64}^\triangle$ & 1.06 & 74.7 & 86.9 & 81.2 & 91.5 & 87.4 & 89.8 & 78.2 & 90.2 & 84.9 \\} % DiaBlo
    \newcommand{\rCrThirteenSaMAFour}{\rowcolor{tableblue} & $\textbf{SaMA}_{4 \times 4}$ & 0.084 & 72.9 & 87.4 & 83.6 & 95.5 & 87.2 & 90.3 & 78.2 & 88.7 & \textbf{85.5} \\}
    \newcommand{\rCrThirteenSaMASix}{\rowcolor{tableblue} & $\textbf{SaMA}_{16 \times 16}$ & 0.336 & {73.5} & {87.6} & {82.6} & {95.4} & {87.7} & {90.2} & {78.0} & {88.3} & {85.4} \\}

    % Llama2-7B (CrTwo)
    \newcommand{\rCrTwoFFT}{& FFT$^\spadesuit$ & 100 & 74.7 & 84.9 & 78.7 & 93.7 & 84.1 & 87.5 & 75.2 & 85.0 & 83.0 \\} %S2FT
    \newcommand{\rCrTwoLoRA}{& LoRA$^\circ$ & 0.83 & 69.8 & 79.9 & 79.5 & 83.6 & 82.6 & 79.8 & 64.7 & 81.0 & 77.6 \\}
    \newcommand{\rCrTwoRefLoRAl}{& RefLoRA$^\bigstar$ & 0.83 & 72.5 & 83.8 & 80.0 & 87.0 & 84.9 & 86.4 & 71.5 & 80.2 & 80.8 \\} % RefLoRA
    \newcommand{\rCrTwoPiSSA}{& PiSSA$^\dagger$ & 0.83 & 67.6 & 78.1 & 78.4 & 76.6 & 78.0 & 75.8 & 60.2 & 75.6 & 73.8 \\}
    \newcommand{\rCrTwoMiLoRA}{& MiLoRA$^\dagger$ & 0.83 & 67.6 & 83.8 & 80.1 & 88.2 & 82.0 & 82.8 & 68.8 & 80.6 & 79.2 \\}
    \newcommand{\rCrTwoLLoRA}{& L2-LoRA & 0.83 & 68.9 & 85.2 & 79.8 & 87.5 & 82.8 & 83.1 & 69.5 & 82.5 & 79.9 \\}
    \newcommand{\rCrTwoKaSA}{& KaSA$^\perp$ & 0.84 & 73.6 & 84.4 & 80.2 & 91.5 & 84.5 & 84.7 & 72.1 & 81.2 & 81.5 \\}
    \newcommand{\rCrTwoDoRAh}{& DoRA$_{r=16}^\circ$ & 0.42 & 72.0 & 83.1 & 79.9 & 89.1 & 83.0 & 84.5 & 71.0 & 81.2 & 80.5 \\}
    \newcommand{\rCrTwoDoRAf}{& DoRA$_{r=32}^\circ$ & 0.84 & 71.8 & 83.7 & 76.0 & 89.1 & 82.6 & 83.7 & 68.2 & 82.4 & 79.7 \\}
    \newcommand{\rCrTwoHiRA}{& HiRA$^\square$ & 0.83 & 71.2 & 83.4 & 79.5 & 88.1 & 84.0 & 86.7 & 73.8 & 84.6 & 81.4 \\} % HiRA
    \newcommand{\rCrTwoSSFT}{& S$^2$FT$^\spadesuit$ & 0.81 & 72.9 & 86.1 & 80.2 & 94.3 & 85.5 & 87.2 & 74.6 & 83.4 & 83.0 \\} % SSFT
    \newcommand{\rCrTwoSMT}{& SMT$^\bullet$ & 0.84 & 72.0 & 83.8 & 80.8 & 93.3 & 82.8 & 86.7 & 74.0 & 81.0 & 81.8 \\} % SMT
    \newcommand{\rCrTwoSMTBest}{& SMT$^\bullet$\textit{(Best)} & 4.91 & 72.6 & 85.2 & 82.0 & 94.4 & 85.7 & 87.8 & 74.5 & 85.0 & 83.4 \\} % SMT
    \newcommand{\rCrTwoDiaBlo}{& DiaBlo$_{N=64}^\triangle$ & 1.04 & 73.9 & 84.8 & 81.7 & 90.0 & 85.0 & 87.9 & 76.8 & 86.8 & 83.4 \\} % DiaBlo
    \newcommand{\rCrTwoSaMAFour}{\rowcolor{tableblue} & $\textbf{SaMA}_{4 \times 4}$ & 0.104 & 73.1 & 85.5 & 81.8 & 94.7 & 85.0 & 88.7 & 73.9 & 86.0 & 83.6 \\}
    \newcommand{\rCrTwoSaMASix}{\rowcolor{tableblue} & $\textbf{SaMA}_{16 \times 16}$ & 0.42 & {73.9} & {86.2} & {81.9} & {94.9} & {86.4} & {89.1} & {77.0} & {86.8} & \textbf{84.5} \\}

    % LLaMA3-8B (CrThree)
    % 
    \newcommand{\rCrThreeFFT}{& FFT$^\spadesuit$ & 100 & 73.9 & 86.2 & 79.1 & 93.1 & 85.8 & 88.1 & 78.2 & 84.0 & 83.6 \\} %S2FT
    \newcommand{\rCrThreeLoRA}{& LoRA$^\circ$ & 0.70 & 70.8 & 85.2 & 79.9 & 91.7 & 84.3 & 84.2 & 71.2 & 79.0 & 80.8 \\}
    \newcommand{\rCrThreeRefLoRAl}{& RefLoRA$^\bigstar$ & 0.70 & 75.5 & 89.7 & 81.1 & 95.6 & 87.3 & 91.0 & 79.8 & 86.0 & 85.8 \\} % RefLoRA
    \newcommand{\rCrThreePiSSA}{& PiSSA$^\dagger$ & 0.70 & 67.1 & 81.1 & 77.2 & 83.6 & 78.9 & 77.7 & 63.2 & 74.6 & 75.4 \\}
    \newcommand{\rCrThreeMiLoRA}{& MiLoRA$^\dagger$ & 0.70 & 68.8 & 86.7 & 77.2 & 92.9 & \textbf{85.6} & 86.8 & 75.5 & 81.8 & 81.9 \\}
    \newcommand{\rCrThreeKaSA}{& KaSA$^\perp$ & 0.71 & 73.6 & 88.1 & 80.4 & 94.7 & 85.5 & 89.7 & 79.4 & 85.6 & 84.6 \\}
    \newcommand{\rCrThreeDoRAh}{& DoRA$_{r=16}^\circ$ & 0.35 & 74.5 & 88.8 & 80.3 & 95.5 & 84.7 & 90.1 & 79.1 & 87.2 & 85.0 \\}
    \newcommand{\rCrThreeDoRAf}{& DoRA$_{r=32}^\circ$ & 0.71 & 74.6 & 89.3 & 79.9 & 95.5 & 85.6 & 90.5 & 80.4 & 85.8 & 85.2 \\}
    \newcommand{\rCrThreeHiRA}{& HiRA$^\square$ & 0.70 & 75.4 & 89.7 & 81.2 & 95.4 & 87.7 & 93.3 & 82.9 & 88.3 & 86.7 \\} % HiRA
    \newcommand{\rCrThreeFlexLoRA}{& FlexLoRA$^\blacktriangledown$ & 0.70 & 72.8 & 89.1 & 80.7 & 96.0 & 86.4 & 90.8 & 81.3 & 87.2 & 85.5 \\} % FlexLoRA
    \newcommand{\rCrThreeSSFT}{& S$^2$FT$^\spadesuit$ & 0.70 & 75.0 & 89.0 & 80.7 & 96.5 & 88.0 & 92.5 & 83.4 & 87.8 & 86.6 \\} % SSFT
    \newcommand{\rCrThreeRandLoRA}{& RandLoRA$_{30}^\curlywedge$  & 0.7 & 76.3 & 88.1 & 80.3 & 95.7 & 86.1 & 90.4 & 80.9 & 87.0 & 85.6 \\}  % RandLoRA 
    \newcommand{\rCrThreeRandLoRABest}{& RandLoRA$_{15}^\curlywedge$  & 1.4 & 72.6 & 88.0 & 81.4 & 95.7 & 87.8 & 91.3 & 80.9 & 89.0 & 85.8 \\}  % RandLoRA
    \newcommand{\rCrThreeSMT}{& SMT$^\bullet$ & 0.71 & 75.7 & 88.4 & 81.4 & 96.2 & 88.2 & 92.7 & 83.2 & 88.6 & 86.8 \\}
    \newcommand{\rCrThreeSMTBest}{& SMT$^\bullet$\textit{(Best)} & 3.01 & 75.1 & 89.9 & 82.4 & 96.3 & 88.8 & 92.6 & 82.8 & 89.6 & 87.2 \\}
    \newcommand{\rCrThreeSaMAFour}{\rowcolor{tableblue} & $\textbf{SaMA}_{4 \times 4}$ & 0.088 & 73.9 & 90.6 & 82.8 & 96.6 & 88.2 & 93.7 & 83.6 & 89.6 & 87.4 \\}
    \newcommand{\rCrThreeSaMASix}{\rowcolor{tableblue} & $\textbf{SaMA}_{16 \times 16}$ & 0.351 & {75.4} & {90.3} & {83.2} & {96.7} & {88.7} & {94.1} & {83.8} & {90.1} & \textbf{87.8} \\}
\fi

% missing ReFT, DiaBlo
% Beam lora

% comment LaX % Nạp dữ liệu vào

\begin{table*}[htbp]
  \vspace{-5pt}
  \centering
  \scriptsize
  \caption{Performance comparison on commonsense reasoning tasks.
  The detailed benchmark with more FT baselines are provided in \cref{tab:exp_cms_full}.}
  \vspace{3pt}
  \resizebox{0.95\textwidth}{!}{%
    \begin{tabular}{llcccccccccc}
      \toprule
      \textbf{Model} & \textbf{PEFT method} & \textbf{\#Params\%} & \textbf{BoolQ} & \textbf{PIQA} & \textbf{SIQA} & \textbf{HellaS.} & \textbf{WinoG.} & \textbf{ARC-e} & \textbf{ARC-c} & \textbf{OBQA} & \textbf{AVG}($\uparrow$) \\
      \midrule
      \multirow{5}{*}{LLaMA-7B} \rCrSevenFFT \rCrSevenLoRA \rCrSevenSMT
      \rCrSevenSaMASix \midrule
      \multirow{5}{*}{LLaMA-13B} \rCrThirteenFFT \rCrThirteenLoRA \rCrThirteenSMT
      \rCrThirteenSaMASix \midrule
      \multirow{5}{*}{LLaMA2-7B} \rCrTwoFFT \rCrTwoLoRA \rCrTwoSMT
      \rCrTwoSaMAFour \rCrTwoSaMASix \midrule
      \multirow{5}{*}{LLaMA3-8B} \rCrThreeFFT \rCrThreeLoRA \rCrThreeSMT
      \rCrThreeSaMAFour \rCrThreeSaMASix
      \bottomrule
    \end{tabular}
  }
  \label{tab:exp_cms_short}
\end{table*}

We continue our evaluation on Commonsense reasoning tasks, including 8 benchmark datasets, training on a single amalgamated dataset of 147k samples from~\cite{hu2023llmadapters}\footnote{Refined from Commonsense-170K because of duplicates. See \Cref{app:data:cms}}. These experiments use several models in the Llama family~\citep{touvron2023llama,touvron2023llama2openfoundation,grattafiori2024llama3herdmodels} to make a comprehensive benchmark.
\cref{tab:exp_cms_short} shows \ftmt performance among other PEFT methods. In most of the models and sub-tasks, \ftmt consistently yields significant improvements, sometimes surpasses FFT. This suggests that by leveraging sparse matrices, \ftmt facilitates a favorable optimization landscape, allowing smoother optimization trajectories. Detailed results are provided in \cref{tab:exp_cms_full}.

\subsection{Training efficiency}
\begin{wrapfigure}{r}{0.44\textwidth}
  % \captionsetup{font=small}
  \centering
  \vspace{-14pt}
  \includegraphics[width=0.4\textwidth]{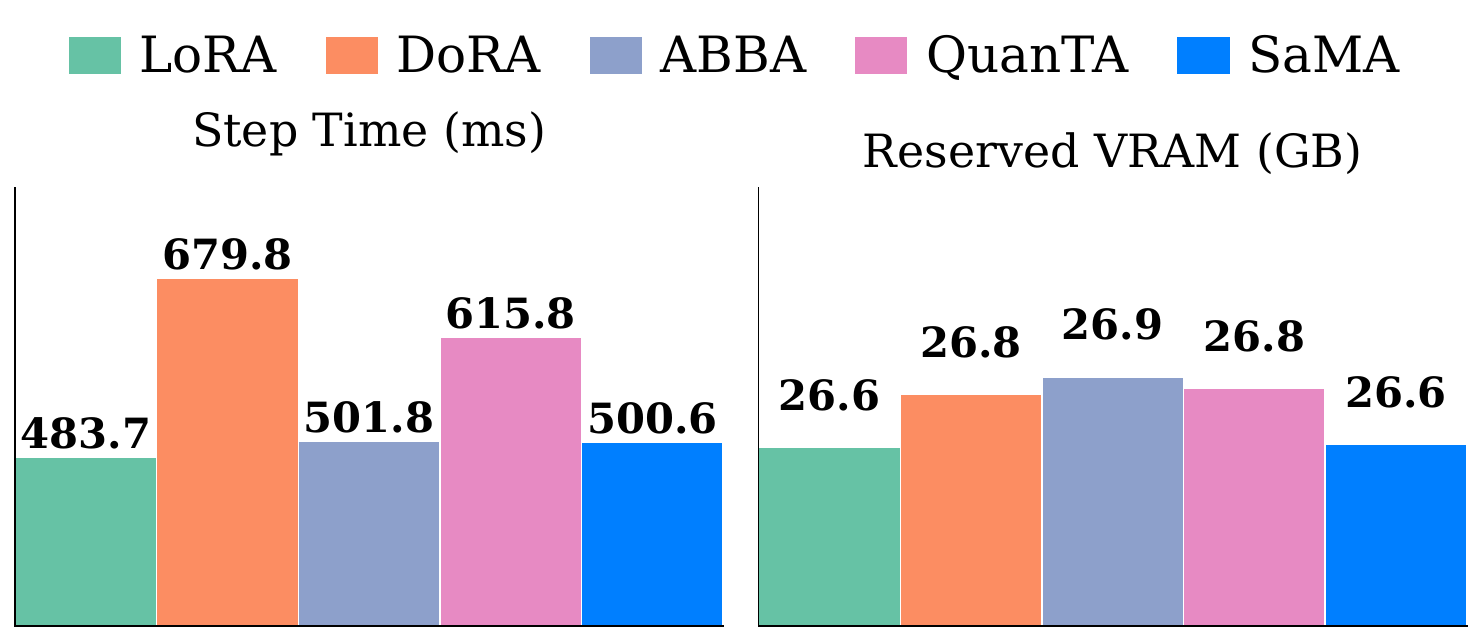}
  % \vspace{3pt}
  \caption{\label{fig:exp_time_memory_llama2b13} Comparison of training time and memory on Gemma2-9B. \ftmt remains hardware-friendly, comparable to LoRA.}
  \label{fig:intro_training_loss}
  % text beneath
  \vspace{-2pt}
\end{wrapfigure}

We validate computational and memory efficiency of \ftmt during the training phase by tracking time per step and reserved memory during the training time on Gemma2-9B (See details and additional results in \cref{sec:app_further_eff_valid}). The results are recorded after 50 warmup steps, averaged 3 times over 25 iterations with comparable parameter counts. 

As shown in \cref{fig:intro_training_loss} and detailed in \cref{sec:app_further_eff_valid}, \ftmt preserves hardware-friendly training behavior: it incurs only marginal runtime overhead of 3.5\% and a comparable memory footprint close to LoRA, and noticeably lower cost than other dense baselines such as DoRA, QuanTA.

\textbf{Summary} Across diverse and challenging NLP tasks and models, \ftmt consistently achieves competitive or superior performance while maintaining all facets of parameter efficiency, scaling stability, computational and memory effectiveness.

\subsection{Discussion on disentangling the \ftmt individual effects}
\label{sec:exp_disentangling}
Given the key paragraphs above, it is obvious to see that \ftmt features several intrisic crucial properties, namely \repname structure, rank capacity as well as reduced gauge freedom of transformation. Although \ftmt adapters simultaneously offer expressivity, structure interaction and diagonal invariant manifolds, it naturally prompts a question on the root cause of SaMA performance:

\begin{tcolorbox}[colback=black!5,colframe=black!15,boxrule=0.4pt,left=5pt,right=5pt,top=3pt,bottom=3pt]
\centering\small\textbf{\textit{What evidence helps disentangle the effects of these factors?}}
\end{tcolorbox}

To thoroughly answer this thought-provoking question, we deem that absolutely isolating these inherently coupled components is mathematically challenging. However, their individual contributions can be observed through structured ablations, optimization dynamics and generalizability. Below is a brief discussion on this problem, the full analysis is provided in \cref{app:sec_meta_evidence}.

\paragraph{i) Beyond nominal rank capacity.} 
Although scaling rank consistently improves \oftmto's expressivity across budgets (\cref{fig:app_sama_vs_lora_drop} (page~\pageref{fig:app_sama_vs_lora_drop})), rank capacity alone does not explain its superiority. In addition, \ftmt preserves a rich, long-tailed singular value spectrum closely aligned with FFT (\cref{fig:ana_spectrum_fft_lora} page~\pageref{fig:ana_spectrum_fft_lora}). See \Cref{app:sec_meta_rank}.

\paragraph{ii) Isolating the \repname topology (Global mixing).} 
A naive diagonal or block-diagonal adapter such as DiaBlo~\cite{gurses2025diablodiagonalblocks} inherently possesses high rank and introduces no parameter degeneracy (since it does not contain any matrix factorization), yet restricts transformations to isolated local subspaces.
Conversely, \ftmt bridges distant feature dimensions to enable global information routing while scaling rank quadratically with budget, yielding better performance. See \cref{app:sec_meta_structure}.

\paragraph{iii) Isolating reduced gauge freedom (Invariant manifolds).} 
Unlike LoRA's dense $\mathrm{GL}(r)$ invariance, which induces optimization concerns under Adam/AdamW (\Cref{sec:express_repara}, page~\pageref{sec:express_repara}), SaMA's structure severely restricts gauge symmetry to scaling only, eliminating shearing and rotations.
In our experiments, gauge-inflating \ftmt by expanding the local gauge group to dense $16\text{-by-}16$ blocks noticeably degrades performance on GSM8K and MATH (see \Cref{app:sec_meta_gauge}), providing evidence for the optimization benefit of reduced gauge freedom.

\subsection{Ablation studies}
\label{sec:main_ablation_init}

\textbf{Sensitivity analysis on initialization of \ftmt factors}

\begin{wraptable}{r}{0.45\textwidth}
  \vspace{-11pt}
  \centering
  % \small 
  \small
  % \scriptsize
  \setlength{\tabcolsep}{8pt}
  \caption{Initialization strategies in \ftmt.}
  \vspace{3pt}
  \label{tab:ablation_init}
        {\renewcommand{\arraystretch}{1.15}  %0.88 \small
        \begin{tabular}{cccc}
            \toprule
            $\sigma^2_\rmL$ &  $\sigma^2_\rmR$ & Acc. & \textbf{$F_1$-scores} \\
            \midrule
            $0$ & $ \num{1e-3}$ & \underline{36.04} &  \textbf{39.77} \\
            $0 $ & $\num{2e-4}$ & \textbf{36.09} & \underline{39.74} \\
            $\num{1e-3}$ & $0$ & 35.47 & 39.41 \\
            $\num{2e-4}$ & $0$ & 35.29 & 39.13 \\
            \bottomrule
        \end{tabular}}
\end{wraptable}
To assess the actual effect of zero-initialization of left and right factors in \ftmt, we conduct experiments on the DROP dataset using Qwen2.5-1.5B~\citep{qwen2023report}, with several choices of standard deviation $\sigma_{init}$ for normal distributions. All other training hyperparameters are fixed across the tests to ensure fair comparison, and the results are reported in \cref{tab:ablation_init} and \cref{sec:app_exp_abinit}.

In general, fine-tuning with \repname parameterization is inherently \emph{resilient} to zero-initialization. Moreover, it is more beneficial to initialize \ftmt factors: $\rmL = 0$ and $\rmR$ from a zero-mean distribution with a small variance: $\rmR \thicksim \mathcal{N}(0,\sigma^2)$. Across both variance schemes of $\sigma^2_{\rmL,\rmR}$, this configuration consistently improves overall performance relative to zero-$\rmR$ -- for both accuracy and $F_1$-score metrics. Hence, we adopt this setting in all main experiments. 

More results on ablation studies on fine-tuning performance on each individual transformer module and asymmetric configurations of $(n_1,m_2)$ are provided in \cref{sec:app_ablation}.

\section{Related Work}
\label{sec:related_work}
\paragraph{Parameter-Efficient Fine-Tuning (PEFT)}
PEFT methods broadly fall into three categories: addition, prompting, and reparameterization. The first two emerged in the early stage of LLM transfer learning, either inserting auxiliary modules~\citep{houlsby2019adapter,he2022sparseadapter} or adapting via task-aware soft tokens~\citep{li2021prefixtuning,lester2021prompttuning}. Both incur additional inference cost or sensitivity to initialization. Our work focuses on \emph{reparameterization}, which preserves the prompt format and base architecture while introducing no inference overhead.

\paragraph{Reparameterization}
Reparameterization methods learn weight updates in a parameterized form and merge them back into the pretrained weights after fine-tuning. LoRA~\citep{hu2022lora} and its variants~\citep{meng2024pissa,liu2024dora,hayou2024loraplusefficient,reploara25truong} have become the dominant baselines in this line, yet a clear performance gap to full fine-tuning (FFT) remains, motivating a recent push toward higher-rank updates.
Representative high-rank approaches include QuanTA~\citep{chen2024quanta}, which uses tensorization with multi-stage factorization; KRAdapter~\citep{albert2025highereffectiverank}, based on the Khatri--Rao product; RandLoRA~\citep{albert2024randlora}, which combines random bases; and SVD-based methods such as KASA~\citep{wang2024kasaknowledgeawaresingularvalue}, Spectral Adapter~\citep{zhang2024spectraladapter}, and SVFT~\citep{lingam2024svft}. ABBA~\citep{singhal2025abba} also achieves rank expansion via Hadamard or Khatri--Rao products, but only up to $r^2/4$, and remains within the LoRA-style dense factorization that suffers from reparameterization ambiguity. In contrast, \ftmt does not commit to either the low- or high-rank regime: a single two-factor parameterization spans both extremes with no preprocessing, while a structured sparse design avoids the dense-factor redundancy of LoRA-family extensions.

\paragraph{Structured sparse matrices}
Structured matrices admit subquadratic cost ($o(n^2)$ for $n \times n$ matrices) and include sparse families and fast transforms. Among them, butterfly~\citep{tridao2019fastltf,parker1995random} and Monarch~\citep{dao2022monarch} matrices have seen renewed interest in deep learning, primarily for parameterizing \emph{square} matrices in pre-training~\citep{thomas2018learning} or full fine-tuning~\citep{dao2022monarch}. \ftmt instead extends these structures to a generic rectangular form and adapts them to additive PEFT, making structured sparsity scalable and rank-flexible for downstream fine-tuning.

\section{Conclusion}
\label{sec:conclusion}
In this study, we present \ftmt together with the \repname parameterization, which bridges the gulf between the Kronecker product and the Monarch matrices with practical applications in reparameterization fine-tuning. We supported this framework with theory, mechanistic analyses and empirical validation. Our extensive experiments substantiate the effectiveness and applicability of the method across various tasks, model architectures and scales. \ftmt consistently achieves competitive results against strong baselines. In particular, compared with FFT, \ftmt substantially narrows the performance gap and is on par with that of FFT, while demonstrating high hardware friendliness and parameter efficiency. Concurrently, the precise conditions under which each rank regime is most beneficial and how far omni-rank PEFT can systematically close the gap to FFT in many settings remain open questions to be better understood.

More broadly, \ftmt suggests that PEFT need not be confined to a fixed rank regime. \ftmt leverages the sparse matrix structure to realize omni-rank adaptation within a unified framework.
\ftmt offers a broader view of PEFT methods beyond regime-specific adapter design. Our findings further suggest that the advantages arise not only from richer expressiveness, but also from a more favorable optimization geometry, without entanglement, under invariant manifolds. 
\paragraph{Limitations} \ftmt implicitly relies on a specific structured parameterization, whose inductive bias may not be equally suitable across tasks, layers, and model families.

\paragraph{Future work} \quad There are several directions for future work to explore:
\begin{enumerate*}[label=(\roman*), font=\bfseries]
\item combining \ftmt with other adaptation strategies, including ensemble-style integration or mixture-of-experts architectures;
\item The core mechanism and underlying role of the sparse structure is not clear so far -- how sparse matrix interactions can improve model generalization and enable the learning of a new task?
\item Extension of the generic rectangular \repname parameterization beyond fine-tuning, including its applicability to pretraining settings where many weight matrices are no longer square.\footnote{Monarch matrices mostly concentrate on square matrices and blocks.}
\end{enumerate*}

\clearpage % \newpage

\section*{Acknowledgements}
This work was supported by the Institute of Information \& Communications Technology Planning \& Evaluation (IITP) grant funded by the Korea government (MSIT) (No. RS-2019-II190079, Artificial Intelligence Graduate School Program (Korea University)), the National Research Foundation of Korea (NRF) grant funded by the Korea government (MSIT) (No. RS-2025-23525509), and the High-Performance Computing Support Project, funded by the Government of the Republic of Korea (Ministry of Science and ICT) (No. RQT-25-070263).

\section*{References}
\medskip
% Disable auto-head
{
    \renewcommand{\bibsection}{}
    \small
    \bibliographystyle{unsrt}
    \bibliography{reference}
}

\clearpage
\appendix
\crefalias{section}{appendix} % latex label for \cref in appendix
\crefalias{subsection}{appendix}

% \appendix

% Common setting

\newcommand{\tableMainCommonInfo}{
    \setlength{\tabcolsep}{7pt}
    \captionof{table}{Default common settings across all experiments, unless otherwise specified.}
    \label{tab:app_hyper_train_common_info}
    \begin{tabular}{lc}
        \toprule
        \textbf \textbf{Hyperparameters} & \textbf{Values} \\
        \midrule
        Learning scheduler & Cosine \\
        Default optimizer & AdamW \\
        AdamW  $(\beta_1, \beta_2)$ & (0.9, 0.99) \\
        Weight Decay & 0.0 \\
        Dropout & 0.0 \\
        Inference sampling & Greedy \\
        \bottomrule
    \end{tabular}
    %   \vspace{-10pt}
}

% DROP
\newcommand{\tableDropMainCommonInfo}{
    \begin{table}[htbp]
    \centering
    \footnotesize
    \caption{Common settings used for main experiments on DROP.}
    \label{tab:app_hyper_drop_train_common_info}
    \begin{tabular}{llc}
        \toprule
        \textbf{Experiment} & \textbf{Hyperparameters} & \textbf{Values} \\
        \midrule
        \multirow{6}{*}{\textbf{Training}} & Modules & (q\_proj, v\_proj) \\
        & Batch size & 4 \\
        & Valid batch & 16 \\
        & Epoch & 2.0 \\
        & Eval step & 25 \\
        & Max token length & 1536 \\
        \midrule
        \multirow{2}{*}{\textbf{Inference}} & LLM max length & 1536 \\
        & New tokens & 50 \\
        \bottomrule
    \end{tabular}
    %   \vspace{-10pt}
    \end{table}
}

\newcommand{\tableDropMainConfig}{
    \begin{table}[htbp]
        \centering
        \footnotesize
        \caption{Learning rate and scaling factor $\evs$ across various Llama2 model scales on the DROP dataset. Symbol $\star$ indicates the configurations from~\citep{chen2024quanta}.}
        \label{tab:app_hyper_drop_config}
        \setlength{\tabcolsep}{8pt}
        \begin{tabular}{llcc}
            \toprule
            \textbf{Model} & \textbf{Configuration} & \textbf{Learning rate} & \textbf{Scaling } \\
            \midrule
            \multirow{8}{*}{Llama2-7B} & LoRA$_{r=8}$ & \num{1e-3},\num{2e-3},\num{8e-4} & 1.0 \\
                                       & LoRA$_{r=32}$ & \num{1e-3},\num{2e-3},\num{8e-4} & 1.0 \\
                                       & LoRA$_{r=128}$ & \num{1e-3},\num{2e-3},\num{8e-4} & 1.0 \\
                                       & QuanTA$_{\text{16-8-8-4}}^\star$ & \num{1e-4} & 1.0 \\
                                       & QuanTA$_{\text{16-16-16}}^\star$ & \num{1e-4} & 1.0 \\
                                       & \ftmt$_\text{2,2}$   & \num{1e-3} & 1.0 \\
                                       & \ftmt$_\text{4,4}$   & \num{1e-3} & 1.0 \\
                                       & \ftmt$_\text{8,8}$   & \num{1e-3} & 1.0 \\
                                       & \ftmt$_\text{16,16}$ & \num{1e-3} & 1.0 \\
                                       & \ftmt$_\text{32,32}$   & \num{1e-3} & 1.0 \\
                                       & \ftmt$_\text{64,64}$   & \num{9e-4} & 1.0 \\
            \midrule
              {Llama2-13B}             & \ftmt$_\text{4-4}$ & \num{8e-4} & 1.0 \\
            \bottomrule
        \end{tabular}
    \end{table}
}

% Commonsense
\newcommand{\tableCmsFullCommonInfo}{
    \begin{table}[htbp]
    \centering
    \footnotesize
    \caption{Common hyperparameters used for main experiments on commonsense reasoning tasks.}
    \label{hyper:cms_full_train_info}
    \begin{tabular}{llc}
        \toprule
        \textbf{Experiment} & \textbf{Hyperparameters} & \textbf{Values} \\
        \midrule
        \multirow{6}{*}{\textbf{Training}} & Modules & (Q, K, V, Up, Down) \\
        & Batch size & 16 \\
        & Valid batch & 16, 32 \\
        & Epoch & 2.0 \\
        & Eval step & 300 \\
        & Max sequence length & 512 \\
        \midrule
        \multirow{2}{*}{\textbf{Inference}} & LLM max length & 1024 \\
            & New tokens & 50 \\
        \bottomrule
    \end{tabular}
    %   \vspace{-10pt}
    \end{table}
}

\newcommand{\tableCmsFullAdapterConfigInfo}{
    \begin{table}[htbp]
    \centering
    \footnotesize
    \caption{Detailed adapter configurations on commonsense reasoning tasks.}
    \label{hyper:cms_info_detailed_configs}
    \begin{tabular}{llccc}
        \toprule
        \textbf{Model} & \textbf{Setting} & \textbf{Learning rate} & \textbf{Scaling ($\evs$)} & \textbf{Config $(n_1,m_2)$} \\
        \midrule
        \multirow{3}{*}{Llama-7B}  & ${\ftmrm}_{4 \times 4}$ & \num{5e-4} & 1.414 & (4,4) \\
                                & ${\ftmrm}_{16 \times 16}$ & \num{5e-4} & 1.414 & (16,16) \\

        \midrule
        \multirow{3}{*}{Llama-13B} & ${\ftmrm}_{4 \times 4}$ & \num{5e-4} & 1.0 & (4,4) \\
                                & ${\ftmrm}_{16 \times 16}$ & \num{5e-4} & 1.0 & (16,16) \\
        \midrule
        \multirow{3}{*}{Llama2-7B} & ${\ftmrm}_{4 \times 4}$ & \num{1e-3} & 0.25 & (4,4) \\
                                & ${\ftmrm}_{16 \times 16}$ & \num{5e-4} & 1.0 & (16,16) \\
        \midrule
        \multirow{3}{*}{Llama3-8B} & ${\ftmrm}_{4 \times 4}$ & \num{2e-4} & 1.0 & (4,4) \\
                                & ${\ftmrm}_{16 \times 16}$ & \num{2e-4} & 1.0 & (16,16) \\
        \bottomrule
    \end{tabular}
    \end{table}
}

\newcommand{\tableCmsAttnInfo}{
    \begin{table}[htbp]
    \centering
    \footnotesize
    \caption{Hyperparameters used for QV-attention experiments on commonsense reasoning, other arguments remain the same as the main experiments. Learning rate and scaling factor correspond respectively to each model.}
    \label{hyper:cms_attn_train_info}
    \begin{tabular}{llc}
        \toprule
        \textbf{Experiment} & \textbf{Hyperparameters} & \textbf{Values} \\
        \midrule
        \multirow{7}{*}{\textbf{\ftmt}} & Modules & (q\_proj, v\_proj) \\
        & Batch size & 16 \\
        & Valid batch & 32 \\
        & Eval step & 300 \\
        & Config $(n_1, m_2)$ & (4,4) \\    
        & Learning rate & \{8e-4, 5e-4, 1e-3, 8e-4, 2e-4\} \\
        & Scaling factor & {1.414, 1.0, 0.5, 1.0, 1.0} \\
        \bottomrule
    \end{tabular}
    %   \vspace{-10pt}
    \end{table}
}

% Arithmetic
\newcommand{\tableArithMainCommonInfo}{
    \begin{table}[htbp]
    \centering
    \footnotesize
    \caption{Common settings used for main experiments on MATH.}
    \label{tab:app_hyper_arith_train_common_info}
    \begin{tabular}{ll>{\centering\arraybackslash}p{4.2cm}}
        \toprule
        \textbf{Experiment} & \textbf{Hyperparameters} & \textbf{Values} \\
        \midrule
        \multirow{5}{*}{\textbf{Training}} & Modules & (q\_proj, v\_proj, k\_proj, o\_proj, gate\_proj, up\_proj, down\_proj) \\
        & Batch size & 4 \\
        & Epoch & 1 \\
        & Max token length & 1536 \\
        \midrule
        \multirow{2}{*}{\textbf{Inference}} & LLM max length & 4096 \\
        & New tokens & 1536 \\
        \bottomrule
    \end{tabular}
    %   \vspace{-10pt}
    \end{table}
}

\newcommand{\tableArithMainConfig}{
    \begin{table}[htbp]
        \centering
        \footnotesize
        \caption{Learning rate and scaling factor $\evs$ across math models on the MATH tasks.}
        \label{tab:app_hyper_arith_config}
        \setlength{\tabcolsep}{9pt}
        \begin{tabular}{llcc}
            \toprule
            \textbf{Model} & \textbf{Configuration} & \textbf{Learning rate} & \textbf{Scaling factor} \\
            \midrule
            \midrule
            \multirow{3}{*}{Mistral-7B} & \ftmt$_\text{2,2}$ & \num{5e-4} & 1.0 \\
                                        & \ftmt$_\text{4,4}$ & \num{5e-4} & 1.0 \\
                                       & \ftmt$_\text{16,16}$ & \num{5e-4} & 1.0 \\
            \midrule
            \multirow{3}{*}{Gemma2-9B}  & \ftmt$_\text{2,2}$ & \num{5e-4} & 1.0 \\
                            & \ftmt$_\text{4,4}$ & \num{5e-4} & 1.0 \\
                            & \ftmt$_\text{16,16}$ & \num{5e-4} & 1.0 \\
            \bottomrule
        \end{tabular}
    \end{table}
}

% Common
\newcommand{\tableAppDatasetInfo}{
    \begin{table*}[htbp]
    \centering
    \footnotesize
    \setlength{\tabcolsep}{5pt}
    \resizebox{0.95\textwidth}{!}{%
        \begin{tabular}{lp{4.05cm}ccccl}
            \toprule
            \textbf{Dataset} & \textbf{Task} & \textbf{\# Train} & \textbf{\# Val} & \textbf{\# Test} & \textbf{Metric} & \textbf{Answer} \\
            \midrule
            DROP~\citep{dua2019drop} & Reading comprehension with discrete reasoning & 2000 & 800 & 1200 & $F_1$-Score & Phrase \\
            \midrule
            COMMONSENSE147K & Commonsense reasoning (Train) & 146627 & 1000 & -- & -- & -- \\
            BoolQ & Commonsense reasoning (Test) & -- & -- & 3270 & Accuracy & Yes/No \\
            PIQA & Commonsense reasoning (Test) & -- & -- & 1838 & Accuracy & Option \\
            SIQA & Commonsense reasoning (Test) & -- & -- & 1954 & Accuracy & Option \\
            HellaSwag & Commonsense reasoning (Test) & -- & -- & 10042 & Accuracy & Option \\
            WinoGrande & Commonsense reasoning (Test) & -- & -- & 1267 & Accuracy & Option \\
            ARC-Easy & Commonsense reasoning (Test) & -- & -- & 2376 & Accuracy & Option \\
            ARC-Challenge & Commonsense reasoning (Test) & -- & -- & 1172 & Accuracy & Option \\
            OBQA & Commonsense reasoning (Test) & -- & -- & 500 & Accuracy & Option \\
            \midrule
            METAMATH (20K) & Arithmetic reasoning (Train) & 20000 & -- & -- & -- & -- \\
            GSM8K & Arithmetic reasoning (Test) & -- & -- & 1319 & Accuracy & Number \\
            MATH & Arithmetic reasoning (Test) & -- & -- & 5000 & Accuracy & Number \\
            \bottomrule
        \end{tabular}
    }
    \caption{Statistics of Datasets for Reading Comprehension, Commonsense, and Arithmetic Reasoning.}
    \label{tab:dataset_stats}
    \end{table*}
}

% Trainingtime commented out
\newcommand{\tableAppTrainingTime}{
    \begin{table}[htbp]
    \centering
    \footnotesize
    \caption{Training time comparison between \ftmt and LoRA across multiple model and task settings. Percentage difference is calculated relative to the LoRA baseline.}
    \label{tab:app_training_time_diff}
    \renewcommand{\arraystretch}{1.3} % Giúp các hàng dãn ra một chút cho dễ đọc
    \begin{tabular}{lccc}
        \toprule
        \multirow{2.5}{*}{\textbf{Model}} & \multicolumn{2}{c}{\textbf{Training Time (H:M:S)}} & \multirow{2.5}{*}{\textbf{Diff. (\%)}} \\
        \cmidrule(lr){2-3}
        & \textbf{LoRA} & \textbf{\ftmt} & \\
        \midrule
        Llama2-7B (DROP) & 2:42:17 & - & +2.47\% \\
        Llama-13B (Commonsense\_15k) & 2:42:17 & - & +2.47\% \\
        Llama3-8B (Commonsense\_15k) & 0:05:43 & - & +1.56\% \\
        Mistral-7B (MetaMathQA\_10k)    & 1:15:55 & - & +3.23\% \\
        Gemma-2 9B (MetaMathQA\_10k)    & 1:45:33 & - & +3.98\% \\
        \bottomrule
    \end{tabular}
    \end{table}
}
%% THIS IS TO GENERATE THE TABLE OF CONTENT
\begingroup
\setcounter{tocdepth}{2}  % Adjust depth: 1=sections, 2=subsections, 3=subsubsections
\renewcommand{\contentsname}{Table of Contents}

\section*{Table of Appendix Contents}
\startcontents[appendix]
\makeatletter
\let\apporiginaldottedtocline\@dottedtocline
\renewcommand{\@dottedtocline}[5]{%
  \ifnum#1=\@ne
    \vspace{2pt}%
  \fi
  \apporiginaldottedtocline{#1}{#2}{#3}{#4}{#5}%
}
\makeatother
\setlength{\parskip}{3.5pt}
\printcontents[appendix]{}{1}{\setcounter{tocdepth}{2}}
\endgroup

\newpage
\section{Mathematical Notations}
\label{app:math_notation}
\textbf{Frequently used notation:} \cref{tab:app_notation_general} summarizes the common symbols used throughout this work.

\begin{table}[H]
    \centering
    \caption{General linear-algebra notation used throughout the paper.}
    \label{tab:app_notation_general}
    \small
    {\renewcommand{\arraystretch}{1.18}
        \begin{tabular}{p{0.18\linewidth} p{0.70\linewidth}}
        \toprule
            \textbf{Symbol} & \textbf{Meaning} \\
        \midrule
            $\rmA,\rmB,\rmC$ & Generic matrices; in the Kronecker formulation, $\rmA \in \sR^{m_1\times n_1}$ and $\rmB \in \sR^{m_2\times n_2}$. \\
            $\otimes$ & Kronecker product. \\
            $\rvx,\rvy$ & Input and output vectors of a linear transformation. \\
            $\rmX,\rmY$ & Matrix forms of $\rvx,\rvy$ obtained via $\unvecop(\cdot)$. \\
            $\vecop(\cdot), \unvecop(\cdot)$ & Vectorization and inverse vectorization operators. \\
            $M,N$ & (Default) Output and input dimensions of the target linear map or an adapter matrix. For a square matrix, we denote its size as $N$ or $N\text{-by-}N$. \\
            $m_1,n_1,m_2,n_2$ & Block dimensions and configuration parameters in the Kronecker/\ftmt construction. \\
            $\rankop(\cdot)$ & Rank operator. \\
            $\erankop(\cdot)$ & Shannon entropy effective rank operator. \\
            $r$ & Rank hyperparameter in LoRA-style baselines. \\
            $\gG_X$ & Gauge group associated with the invariance manifold of a generic parameterization $X$. \\
            $GL(r)$ & General linear group, i.e., the set of all invertible square matrices of size $r \times r$. \\
            $O(r)$ &  Rotational group -- a set of rotational square matrices. $\rmA \in O(r)$ means that $\rmA^{-1} = \rmA^\top$ and $\det(\rmA)= 1.$\\
            $\gB\gD$ & Set of all block-diagonal matrices. \\
            $\sRs$ & $\sRs = \sR \backslash \{ 0 \}$: the set of nonzero real numbers. \\
            $\lVert \cdot \rVert_F$ & Frobenius norm of a matrix. \\
            $\langle \cdot, \cdot \rangle_F$ & Frobenius inner product of two matrices. \\
            $\ve_i$ & Standard basis vector with only a \(1\) in the \(i\)-th position and zeros otherwise. \\
            $\pi()$ & Permutation function - a bijective function that rearranges a set of indices. For instance, $\pi(i) = j$ means that the \(j\)-th element is moved to the \(i\)-th position \\
            $\pi_{m,n}$ & Index map of $\mP_{m,n}$, i.e. $\mP_{m,n}\ve_i = \ve{\pi_{m,n}(i)}$ \\
            $\overline{a..b} $ & Integer interval ranging from $a$ to $b$ -- or $[a,b] \cap \sZ$ \\
        \bottomrule
        \end{tabular}
    }
\end{table}

\textbf{\ftmt-specific notation.} \cref{tab:app_notation_sama} summarizes the symbols used for \ftmt and \repname matrices.

\begin{table}[H]
    \centering
    \caption{\ftmt-specific notation.}
    \label{tab:app_notation_sama}
    \small
    {\renewcommand{\arraystretch}{1.18}
        \begin{tabular}{p{0.18\linewidth} p{0.70\linewidth}}
        \toprule
            \textbf{Symbol} & \textbf{Meaning} \\
        \midrule
            $\Delta \rmW$ & Trainable adapter update added to the pretrained weight matrix in FT \\
            $\rmL, \, \rmR$ & Block-diagonal trainable factors in the \repname parameterization, with constituent blocks as $\rmL_i$ and $\rmR_j$. \\
            $\mP$ or $\mP_{n_1,m_2}$ & A generic form of perfect shuffle permutation. In general, $\mP_{m,n}$ rearranges a vector by reshaping it into a row-major $m$-by-$n$ matrix then transposes and re-vectorizing it. \\
            $\mP_1$ & Permutation matrix for monarch parameterization of all square components. \\
            $\mP_l,\mP_r$ & Fixed permutation matrices used to route and mix block interactions. In \repname parameterization, $\mP_l = \mP_{m_1,m_2}, \mP_r=\mP_{n_1,m_2}$. \\
            $\evs, \evs_\ftmt$ & Scaling factor applied to the \ftmt adapter; for stable scaling, $\evs=\Theta(n_1^{1/4})$. \\
            $q$ & Hidden dimension of the \repname parameterization, defined as $q = n_1 \cdot m_2$ \\
            $\mset, \mset^{(m,n)}$ & \repname manifolds with or without specific primary configuration, parameterized by (\#blocks in $\rmR$, \#blocks in $\rmL$). \\
            $\mset _g, \mset^{(m,n)}_g$ & Generative factor tuples $(\rmL,\rmR)$  associated with the \repname manifold. \\
            $(n_1,m_2)$ & Primary configuration of \ftmt: (\#blocks in $\rmR$, \#blocks in $\rmL$). \\
        \bottomrule
        \end{tabular}
    }
\end{table}

We summarize the recurring symbols used across the main text and appendix.
These are common quantities and operators that appear across multiple sections, although some may be redefined locally to make the context clear. Context-specific notation will be introduced in the corresponding sections.
For the sake of readability, we separate general linear-algebra notation from \ftmt-specific notation.
as displayed in \cref{tab:app_notation_general} and \cref{tab:app_notation_sama}.
\vspace{5pt}

\begin{definition}
    Perfect shuffle permutation matrix of shape $r\times r$~\citep{vanloan2000ubiquitous}:
    \begin{equation}
        \mP_{p,q} = \begin{bmatrix}\ermI_r[1:q:r,:]^\top & \ermI_r[2:q:r,:]^\top & \ldots & \ermI_r[q:q:r,:]^\top\end{bmatrix}^{\!\top},\quad r=pq.
        \label{eq_shuffle}
    \end{equation}
\end{definition}
As explained in the main text, the perfect shuffle permutation $\mP_{p,q}$ reshapes a vector of length $r$ into a $p \times q$ matrix in row-major order, transposes the matrix, then re-vectorizes it.

\paragraph{Terminology explanation}

\emph{Omni-rank:} ability to realize a broad range of matrix ranks, up to the target regime controlled by the configuration, including low-, mid-, high-, and full-rank transformations. Within this work, it is also refered to as flexible rank or quadratic rank property.

In this work, we define \emph{omni-rank} not as the empirical capacity to realize every possible integer rank, but rather as the structural flexibility to span a broad spectrum of adapter ranks while respecting the available parameter budget.

\emph{\repname:} structured parameterization underlying \ftmt, built from sparse block-diagonal  factors and the corresponding perfect shuffle permutations. It can be viewed as a natural extension of the Kronecker product.

\section{Broader Literature Review}
\paragraph{Additional adapter:} The first group can be referred to as Addition approaches, which involves injecting extra adapters into the model architecture. These modules can be integrated in many ways: Serial adapters are interposed between existing attention or multilayer perceptron (MLP) components~\citep{houlsby2019adapter, pfeiffer2021adapterfusion,he2022sparseadapter}, while parallel adapters coexist alongside these modules~\citep{he2021parallel}. % In overall, these approaches intend to increase memory for storage as well as computational load in inference due to the extra components that can not be merged into the original model weights.
\paragraph{Prompting} Secondly, the Prompting group, inspired by the success of textual prompting methods, Prompt tuning~\cite{lester2021prompttuning} and Prefix tuning~\cite{li2021prefixtuning} add extra soft tokens to the original input and solely focus on fine-tuning these trainable tokens, or soft prompts in~\cite{qin2021learning}. Besides, a recent work by~\cite{wang2023twostagellmfinetuning} proposes two-stage fine-tuning, where prompting techniques are used in the first stage.  Overall, these methods commonly encounter efficacy issues due to their sensitivity to initialization affecting their overall robustness and efficiency. These first two categories, either altering model's input sequence or architecture, lead to an increase in model size and introduce additional inference latency and computational overhead compared to the pretrained models. So the research directions were prevalent in early time of PEFT, but less advanced recently, so they are outside of the scope of this paper.

\paragraph{Isometric-based Fine-tuning} Grounded in the metric preservation insight from reparameterization fine-tuning, there are other directions that use orthogonal adapters in the form multiplicative transformations.
Particularly, the orthogonal matrices are parameterized via block-diagonal structures in OFT~\citep{qiu2023controlling} or GS~\citep{gorbunov2024group}, that also utilize sparsity structures; 
Moreover, LoCO~\cite{sdml2026loco} constructs rotations from low-rank skew-symmetric matrices by means of Woodbury identity; some works focus on simple component like Givens rotations in qGOFT~\citep{ma2024parameter}.
Our work, in contrast, takes additive transformations as a core principle because of several prominent advantages. We argue that additive adapters are more effective and expressive in PEFT due to the simplicity and mathematical generality
.

\paragraph{Reparameterization} \textit{Low-rank Adaptation}. Building on the low intrinsic rank hypothesis~\citep{aghajanyan2021intrinsic}, LoRA~\cite{hu2022lora} uses a product of two low-rank matrices to build weight updates, avoiding inference overhead of previous approaches with adapters~\citep{houlsby2019adapter}, or optimization bottleneck in prompt tuning\citep{li2021prefixtuning,lester2021prompttuning}, then LoRA quickly becomes one of the most common techniques. Beyond the vanilla design, subsequent works extend this framework through adaptive rank allocation~\citep{zhang2022adaptive}, alternative initialization~\cite{meng2024pissa,wang2024milora}, spectrum~\citep{liu2025flexloraentropyguidedflexible}, decomposition~\citep{liu2024dora,bini2024delora,lion2025polar}, Hadamard product~\citep{huang2025hira,singhal2025abba}, sharing components~\citep{koohpayegani2023nola, he2024rasa}, random non-trainable matrices~\citep{kopiczko2023vera,koohpayegani2023nola}, fusion~\citep{liu2025seedlora}, training resources (gradient~\citep{wang2024loraproarelowrank,zhang2025reflorarefactoredlowrank,wang2024loraga}, memory~\citep{zhao2024galorememoryefficientllm,liu2025misa}, learning dynamics~\citep{hayou2024loraplusefficient,dzievski2023rslora,zhang2025loram}), attention interaction~\citep{tong2026prism} or reparameterization~\citep{reploara25truong, jung2025gralora}.

Although ABBA~\citep{singhal2025abba} also exhibits a rank-expansion effect, indirectly suggesting a limited form of \emph{Omni-rank} -- $r^2/4$ only. However, this property is not explicitly formulated as a central design principle or motivation in their work. In contrast, \ftmt\ directly builds on the notion of \emph{Omni-rank} adaptation as a primary objective, and develops both the methodology and analyses around this perspective.
Moreover, ABBA remains fundamentally within the LoRA-style family, and thus still inherits the reparameterization ambiguity associated with dense factorized updates.
Conversely, \ftmt is designed to achieve the \emph{Omni-rank} target of $n_1 \cdot m_2$, equivalent to $r^2$, while disentangling gauge symmetry (\cref{sec:express_repara}) and maintaining an implementation-friendly sparse architecture (\cref{sec:method_prac_regime}).

\paragraph{Partial specification and sparse updates} This line of approach aims to reduce the trainable parameters by selecting a subset of pre-trained model weights or sparse matrices in additive adapters. This kind of method attracts much attention in recent research but either sacrifice hardware efficiency~\citep{he2025smt}, even requiring full backward propagation~\citep{song2024sparseenoughfinetuning,xu2024randomask} imposing additional training procedures~\citep{pan2024lisalayerwiseimportance, yang2024s2ftefficientscalable}, or only permit local interaction without dense matrices~\citep{gurses2025diablodiagonalblocks}. In addition, MoRE~\cite{tan2025more} can be conceptually viewed as an ensemble of $b$ LoRA-style block adapters of rank $r_{blk}$, combined with permutation. This partitioned design restricts the global receptive field, leading to a fragmented interaction topology. Apart from those, our work facilitates sparse constituent matrices to construct module-specific dense adapters that are more practical, easy-to-implement while maintaining simplicity and efficiency by virtue of reparameterization for each individual weight matrix.

\paragraph{Structured sparse matrices} Structured matrices are those with more optimal time complexity and space complexity -- below the threshold of quadratic for a square matrix $n\times n$. Many generations of these transformations have been adopted in machine learning to alter dense weight matrices. In~\cite{dao2019kaleidobutterfly}, the expressiveness of butterfly matrices~\cite{tridao2019fastltf} can cover these structured matrices almost optimally regarding runtime and memory, followed by the proposal of Pixelated Butterfly~\citep{chen2021pixelated} to make them hardware-friendly. Moreover, Dao et al. \citep{dao2022monarch} show that monarch representation is strictly more general than butterfly matrices while enhancing hardware utilization; the product of two monarch matrices can represent convolution, Hadamard transform, Toeplitz matrices and AFDF matrices.

\paragraph{Geometric perspectives and Manifold reparameterization in PEFT}
In deep learning, the structural formulation of models inherently introduces continuous gauge symmetries into the parameter space, as highlighted in~\citep{neyshabur2015pathsgd}.
In the context of PEFT, gauge symmetry is implicitly addressed via constraints proposed in recent works. For instance, Riemannian LoRA~\citep{zhang2024riemannian} handles the reparameterization issues by modeling the low-rank adaptation space strictly as a Riemannian submanifold.
LoRA+~\citep{hayou2024loraplusefficient} reveals that parameter space degeneracy in LoRA induces asymmetric optimization dynamics, leading to suboptimal feature learning.
Our work explicitly analyzes the gauge symmetry in the parameter space and resolves it through rigid structural constraints. By adopting sparse structured matrices, our approach directly breaks the transform structure of the continuous gauge invariance inherent in standard dense factorization; replacing them with block-diagonal matrices intrinsically manages parameter entanglement.

\section{Practical Implementation}
\label{sec:method_prac_regime}

This section provides the practical details of \ftmt: the trainable parameter count of a \repname adapter, the initialization scheme inherited from LoRA, the efficient batched block-matmul realization that avoids materializing $\Delta\rmW$, and the absence of both pre-processing and inference-time overhead.

\textbf{Parameters.}\enspace
For a square weight matrix $\rmW \in \sR^{N \times N}$ with $n_1 = m_2 = \nu$,
a \repname matrix $\rmM = \mP_l^\top\rmL\mP_r\rmR$ is described by $2N\nu$
trainable parameters in total: each of $\rmL$ and $\rmR$ contains $\nu$ dense blocks
of size $\nu \times \nicefrac{N}{\nu}$, contributing $N$ parameters per factor.
The permutations $\mP_l, \mP_r$ are fixed and parameter-free.

\textbf{Initialization.}\enspace
Following the reparameterization convention of LoRA~\citep{hu2022lora}, we
initialize $\rmR$ from a zero-mean Gaussian with small standard deviation
$\sigma_{\rmR}$, and set $\rmL = 0$. This ensures $\Delta\rmW = \evs\,\mP_l^\top\rmL\mP_r\rmR = 0$
at the start of training while keeping gradients non-vanishing.
The scalar $\evs_{\ftmrm}$ is a configuration-dependent hyperparameter whose role in
training stability is analyzed in \cref{main:theorem_rank_stable}.
%  \cref{sec:method:rank_property}.

\textbf{Efficient computation.}\enspace
Although $\Delta\rmW$ in \cref{eq_sama_delta} defines a dense linear map,
we never materialize it explicitly. Instead, following~\citep{dao2022monarch},
multiplication $\Delta\rmW\,\rvx$ is carried out in two steps: reshape the
input into a matrix, apply $\rmR$ as a batched block-matrix multiply,
permute via $\mP_r$, apply $\rmL$ similarly, then permute via $\mP_l^\top$.
This exploits batch matrix multiplication — the core primitive of modern
GPU hardware — at a total cost of $O(\nu N)$ FLOPs, with no recursive
graph construction~\citep{tridao2019fastltf}.

\textbf{No pre-processing.}\enspace
Unlike D2S-FT~\citep{dao2022monarch}, which requires projecting pre-trained
weights into a sparse form before training begins, \ftmt requires no such
initialization step. Pre-trained weights are frozen and never modified;
all computation targets the sparse adapter $\Delta\rmW$ alone. This
eliminates the Dense-to-Sparse projection overhead present in related
methods~\citep{pan2024lisalayerwiseimportance,wang2024milora} while
preserving comparable operational complexity.

\textbf{No inference overhead.}\enspace
Like other reparameterization-based methods, \ftmt adapter can be merged
into the pre-trained weight at inference time ($\rmW_\mathrm{ft} = \rmW_0 + \Delta\rmW$),
adding zero latency and preserving the original model architecture for
optimized deployment. Only the lightweight adapter needs to be stored,
not the full fine-tuned model.

\newpage
\section{Extended Theoretical Discussion and Proofs}
\label{sec:app_theoretical}

\subsection{Rank property (Proof of \texorpdfstring{\cref{theo_sama_rank}}{Theorem of rank})}
\label{sec:app_sec_rank_pro}
\begin{theorem*}
    \newcommand{\mylabelprefix}{app}
    \label{app_theo_sama_rank}
    Let $\rmL$ and $\rmR$ be two block-diagonal matrices, as defined in \cref{eq:method_def_LR}, with $n_1 m_2 \leq \min{(M,N)}$. Suppose that within every submatrix, each entry is chosen from independent and identically distributed data (\textit{i.i.d}). Then, the rank of the SaMA adapter matrix:
\begin{equation}
    \rankop({\Delta \rmW}) = \rankop({\evs \mP_l^\top \rmL \mP_r \rmR}) = n_1 m_2,
    \label{\mylabelprefix:eq_theo_rank}
\end{equation}
holds with a probability of 1.

\end{theorem*}

\Cref{theo_sama_rank} characterizes the rank that SaMA can attain during training; it does not describe
the initialization, where $\rmL=\mathbf{0}$ and hence $\Delta\rmW=\mathbf{0}$.
Crucially, this theorem highlights a significant distinction from low-rank adaptation -- the LoRA family~\citep{wang2024milora,meng2024pissa}, and high/full-rank adaptation~\citep{chen2024quanta, albert2025highereffectiverank, huang2025hira}. Whereas LoRA~\citep{hu2022lora} utilizes \emph{`low intrinsic rank'} assumption~\citep{aghajanyan2021intrinsic}, \ftmt further leverages the principle with regarding the current development of large models and complex downstream tasks.
\begin{lemma}
    \label{lemma:app_fr}
    Let ${\vX_1, \vX_2, \ldots, \vX_{m_2}}$ denote $m_2$ vectors in $\sR^{n_2}$, where $m_2 \leq n_2$, drawn i.i.d from a Gaussian or any other distribution. Then with probability 1, the set ${\vX_1, \vX_2, \ldots, \vX_{m_2}}$ will be linearly independent. 
\end{lemma}
\begin{proof}[Proof of the lemma]
    We first note that any measure defined via a Gaussian or any other continuous probability distribution is absolutely continuous with respect to the Lebesgue measure. This implies that such distributions assign zero probability to any set of Lebesgue measure zero.

    Consider the inductive process of drawing the set $\{\vX_1, \vX_2, \ldots, \vX_{m_2}\}$ in $\sR^{n_2}$. Since the vectors are drawn independently, we first consider $\vX_1$. The probability that $\vX_1$ is the zero vector is $0$ with respect to the Lebesgue measure on $\sR^{n_2}$, and consequently for any measure absolutely continuous to it.
    Upon drawing the subsequent vector $\vX_{k+1}$ (where $k < m_2 \leq n_2$), the probability that $\vX_{k+1}$ lies in the subspace $V_k = \text{span}\{\vX_1, \dots, \vX_k\}$ is also $0$. This is because $V_k$ is a subspace of dimension at most $k$, which constitutes a set of zero Lebesgue measure in $\sR^{n_2}$. 

    By induction, it follows that the set $\{\vX_1, \vX_2, \ldots, \vX_{m_2}\}$ is linearly independent with probability $1$ (almost surely). For the general case where vectors may not be restricted in length, we note that normalizing $\vX_i$ by its non-zero norm does not affect linear independence; thus, the result holds for any continuous distribution on $\sR^{n_2}$.
\end{proof}
\paragraph{Proof of \cref{theo_sama_rank}} \enspace %\mbox{}\\
\begin{proof}[Main proof]
    Consider a block-diagonal matrix  $ \rmR = \diag(\rmR_1, \rmR_2, \ldots,\rmR_{n_1}) $, where $\rmR_i \in \sR^{m_2 \times n_2}$.
    
    For any pairs of $(\rmR_i, \rmR_j), i \neq j$, their columns and rows are in different locations in $\rmR$. Due to sparsity of $\rmR$, the blocks $\{\rmR_1, \dots, \rmR_{n-1}\}$ occupy pairwise disjoint set of rows/columns, hence it is the case that these columns/rows in the first submatrix are not a linear combination of columns/rows from the other submatrices and vice versa. It follows that the overall rank of $\rmR$ is simply a summation over the constituent block elements $\rmR_i$.
    \begin{equation}
        \label{eq:app_rd_rcomp}
        \rankop(\rmR) = \sum_{i=1}^{n_1} \rankop(\rmR_i) \quad \text{while:} \quad \rankop({\rmR_i}) \leq \min{(m_2, n_2)},
    \end{equation}
    In fact, \cref{sec:method_prac_regime} describes practical implementation of \ftmt adapters. We usually use small values of $(n_1, m_2)$: $n_1 \leq \sqrt{N}, m_2 \leq \sqrt{M}$, and $n_1 m_2 \leq \min{(M,N)}$, where $N, M$ denote the numbers of input and output features, respectively. Hence, in the field of deep learning applications, it is a common observation that:
    \begin{equation*}
        \min{(m_2, n_2)} = m_2 \quad \text{and} \quad \min{(m_1, n_1)} = n_1,
    \end{equation*}
    Therefore, back to \cref{eq:app_rd_rcomp}:
    \begin{equation*}
        \rankop(\rmR) = \sum_{i=1}^{n_1} \rankop(\rmR_i) \leq \sum_{i=1}^{n_1} m_2 = n_1 m_2,
    \end{equation*}
    Similar analysis for the left factor $\rmL$, note any perfect shuffle permutation matrix $\mP$ is orthogonal:
    \begin{equation}
        \label{eq:rank_property_rl}
        \Rightarrow \left\{
        \begin{aligned}
            \rankop{(\mP_r \rmR)} &= \rankop{(\rmR)} \leq n_1 m_2 \\
            \rankop{(\mP_l^\top \rmL)} &= \rankop{(\rmL)} \leq m_2 n_1,
        \end{aligned}
        \right.
    \end{equation}
    By means of rank property for matrix multiplication, it follows that:
    \begin{equation}
        \label{eq:rank_property_leq_final}
        \rankop({\Delta \rmW}) = \rankop({\evs \mP_l^\top \rmL \mP_r \rmR}) \leq \min{\left(\rankop{(\mP_r \rmR)}, \rankop{(\mP_l^\top \rmL)} \right)} \leq n_1 m_2
    \end{equation}
    \paragraph{Generic block entries} Due to this assumption, almost surely (with probability 1) that the equation of rank matrix holds: $\rankop(\rmR_i) = \min{(m_2, n_2)} = m_2$ as in \cref{lemma:app_fr}. Hence, the overall rank could be roughly expressed as: $\rankop(\rmR) = n_1 m_2$, and similarly $\rankop(\rmL) = m_2 n_1$. That, along with permutation matrix $\mP$, leads in the empirical ranks in \cref{eq:rank_property_rl}:
        \begin{equation}
        \Rightarrow \left\{
        \begin{aligned}
            \rankop{(\mP_r \rmR)} &= \rankop{(\rmR)} = n_1 m_2 \\
            \rankop{(\mP_l^\top \rmL)} &= \rankop{(\rmL)} = m_2 n_1,
        \end{aligned}
        \right.
    \end{equation}

    Applying Sylvester's rank inequality~\citep{horn2012matrix, sylvester1884involutants} then gives:
    \begin{align}
        \label{eq:rank_property_gre_final}
        \rankop({\Delta \rmW}) &= \rankop({ (\mP_l^\top \rmL) (\mP_r \rmR)}) \notag \\ 
        &\geq \rankop({\mP_l^\top \rmL}) + \rankop({\mP_r \rmR}) - q \notag \\
        &= q + q - q = q, \qquad \text{where:} \quad q = n_1 m_2.
    \end{align}
    These two \cref{eq:rank_property_leq_final,eq:rank_property_gre_final} imply that the generic rank of \ftmt adapter is almost surely equal to $n_1 m_2$ in practice: $\qquad \rankop({\Delta \rmW}) = n_1 m_2$.
\end{proof}
\paragraph {Discussion and visualization} We also illustrate the property of flexible hidden rank in \cref{fig:app_theo_monarch_static_rank_analysis} by means of Action space alignment scores (ASA). ASA serves as a rigorous metric that evaluates the structural expressivity of transformation. Unlike point-wise metrics such as Cosine similarity which are sensitive to local block-rank constraints, ASA measures the global geometric overlap between the \repname reachable range and the target FFT subspace. It provides a pure assessment of whether the representation's total degrees of freedom are sufficient to span the required transformation space. In practice, ASA scores are given by the overlap between the subspaces spanned by their respective outputs given the same input distribution from the viewpoint of singular vectors. % Instead of calculating point-wise distance between weight matrices

A drop in the ASA scores below $1.0$ indicates an absolute subspace collapse, signifying the intrinsic rank of the transformation has surpassed the global capacity of the \repname parameterization. At this threshold, the model encounters a hard structural bottleneck: it can no longer satisfy the stationary conditions of the FFT in the experiment of \cref{app:proof_converges_fft}, forcing a lossy projection where critical dimensions become mathematically unrecoverable regardless of the optimization effort. Meanwhile, the threshold rather supports the validation of our rank argument in this section of \cref{sec:app_sec_rank_pro}.
\subsection{Rank stability condition (Proof of \texorpdfstring{\cref{main:theorem_rank_stable}}{Theorem})}
\label{sec:app_rank_stable}
In order to accurately define stability for an adapter with respect to rank configurations, following~\citep{dzievski2023rslora}, we define the term ``rank-stabilized adapter'' as follows:
\begin{definition}
    \label{app_def_rank_stable}
    A \ftmt adapter: $\Delta \rmW = \evs \mP^\top \rmL \mP \rmR$ in \cref{eq_sama_delta} is \textbf{rank-stabilized}\footnote{Because the primary configuration $(n_1,m_2)$ directly pertains to the adapter's hidden rank, we inherit this terminology from the original work for clarity and consistency.} if the following conditions hold:
    \begin{enumerate}
        \item If the inputs to the adapter are i.i.d.\ such that the $m'$-th moment is $\mathbf{\Theta}_{n_1,m_2}(1)$ for each entry, then the $m'$-th moment of the adapter outputs is also $\mathbf{\Theta}_{n_1,m_2}(1)$ for each entry.
        \item If the gradient of the loss with respect to the adapter outputs is $\mathbf{\Theta}_{n_1,m_2}(1)$ for each entry, then the loss gradients with respect to the adapter inputs are also $\mathbf{\Theta}_{n_1,m_2}(1)$ for each entry.
    \end{enumerate}
\end{definition}
In \cref{app_def_rank_stable}, there are two complementary conditions together supporting fine-tuning applications. First, \textbf{Forward-pass stability (Condition 1)} ensures that the scale of hidden representations does not explode or vanish under different rank configurations during inference. If the output moments depend on $(n_1,m_2)$, the activation distribution can shift substantially, requiring recalibration of subsequent layers or learning rates.

Second, \textbf{Backward-pass stability (Condition 2)} is crucial for gradient flow and helps avoiding gradient collapse during training. It guarantees that the gradient signal passing through the adapter remains at a consistent magnitude $\mathbf{\Theta}_{n_1,m_2}(1)$. Without this property, instability in gradient flow makes it difficult to find a universal hyperparameter setting across architectural scales.
Both of the two conditions play significant roles that preserve signal energy magnitudes and stabilize computational graphs throughout all layers during training and inference time.

Unlike rsLoRA~\citep{dzievski2023rslora} that uses analysis in the limiting regime when $r \rightarrow \infty$, we only examine \ftmt adapters in normal settings and prove that the scaling of $\evs$ (sometimes being denoted as $\evs_{\mathrm{\oftmto}}$ for clarity) that could achieve \emph{rank-stabilized} adapters is:
\begin{equation}
    {\evs}_{\mathrm{\oftmto}} = n_1^{0.25},
\end{equation}
for some scalar $\alpha \in \sR$.

\begin{theorem*}[Detailed statement of \cref{main:theorem_rank_stable}]
    Consider \ftmt adapters which are of the form $\Delta \rmW = \evs \mP_l^\top \rmL \mP_r \rmR$, where $\mP_{l,r}$ is the non-trainable perfect shuffle permutation matrix, two trainable factors $\rmL$ shape ${m_2 m_1 \times m_2  n_1}$, $\rmR$ shape ${n_1 m_2\times n_1 n_2}$ are block-diagonal matrices as defined in \cref{eq:method_def_LR} -- where $(M,N)$ are large and do not depend on the configuration $(n_1,m_2)$, initialized such that $\rmL$ is a zero matrix $\mathbf{0}_{m_2 m_1 \times m_2  n_1}$, entries of $\rmR$ are i.i.d with mean $\mathbf{0}$ and variance $\mathbf{\sigma_{\rmR}^2}$ and scaling factor $\evs$ (or $\evs_{n_1,m_2}) \in \sR$; and train the adapter for finite $n$ SGD updates.

    In expectation over the initialization method, all adapters are rank-stabilized if and only if:
    \begin{equation}
    {\evs}_{\ftmrm}= \Theta\left(n_1^{\nicefrac{1}{4}}\right),
        \label{app_eq_scaling}
    \end{equation}
\end{theorem*}

The above theorem is a detailed description and explanation of \cref{main:theorem_rank_stable}. It suggests that the there exists a feasible way to ensure stability through the adapters regardless of abitrary rank configurations, is to adapt the scaling factor: ${\evs}_{\mathrm{\oftmto}} \propto n_1^{0.25}$

\paragraph{The role of scaling and multi-layer stability} A critical insight from \cref{main:theorem_rank_stable} is that maintaining stability across arbirtary ``rank configuation'' of $(n_1,m_2)$ necessitates scaling the adapter by: ${\evs}_{\mathrm{\oftmto}} \propto n_1^{0.25}$. While it may initially seem that activations and gradients in middle layers would inevitably depend on $(n_1,m_2)$ due to the influence of preceding (forward pass) or succeeding (backward pass) adapters, this dependency is mitigated through an inductive flow:
\begin{itemize}
    \item \textbf{Forward propagation} Since the intial input data is independent of $(n_1,m_2)$, applying the rank-stablization theorem layer-by-layer from input to output guarantees that the activation magnitude remains stable ($\mathbf{\Theta}_{n_1,m_2}(1)$) throughout the entire forward pass.
    \item \textbf{Backward propagation} Similarly, by starting from the loss gradient at the output and moving backward, the theorem ensures that gradient magnitudes do not explode or vanish, preserving a consistent scale of $\mathbf{\Theta}_{n_1,m_2}(1)$ across all adapters.
\end{itemize}
\paragraph{Limitation note} It is also important to distinguish between numerical stability and feature quality. The theorem only ensures that the learning process does not fall into a collapse mode in the limiting regime of ``rank configuration'' $(n_1,m_2)$, but it does not concentrate and establish a connection between heavy-mode configuration and better features. To this extent -- the quality of trained features and the practical benefits of the corrected scaling factor ${\evs}_{\mathrm{\oftmto}}$ may require further analysis and validation.

\begin{lemma}
    \label{lemma:rosenthal_absolute}
    Given a set $\rmX = \{\rmX_K\}$ contains $K$ independent and identically distributed (i.i.d) random variables with mean $\E[\rmX_k] = 0, \Var(\rmX_k) = \sigma_X^2 > 0, \forall k$, and high-order absolute moments $\E[\lvert \rmX_k \rvert^m], \forall m \geq 2$ are finite.

    Let $\rmS = \sum_{m=1}^K \rmX_k$ be the sum of these random variables, as K grows large, the \textit{m-th} absolute moment of $\rmS $ follows the asymptotic scaling:
    \begin{equation}
        \E \left[\lvert \rmS \rvert^m \right] = \Theta\left(K^{m/2} \Var(\rmX_k)^{m/2}\right) = \Theta\left(K^{m/2} \sigma_X^m\right).
    \end{equation}
\end{lemma}
\begin{proof}[Proof of \cref{lemma:rosenthal_absolute}]
    By Rosenthal inequality~\citep{Rosenthal1970OnTS} for independent zero-mean random variables, there exist a constant $C_m$ such that the upper bound:
    \begin{equation}
        \label{eq:rosenthan_inquality}
        \E \left[  \lvert \rmS \rvert^m  \right] \leq C_m \, \max{ \left\{\sum_{k=1}^K \E \left[ \lvert \rmX_k \rvert^m \right], \left( \sum_{k=0}^K \E \left[ \rmX_k^2 \right] \right)^{m/2} \right\} },
    \end{equation}
    Since the variances are i.i.d, we analyze the scaling rate with respect to $K$ for both components inside the $\max$ function.
    \begin{itemize}
        \item First component: $ \sum_{k=1}^K \E \left[ \lvert \rmX_k \rvert^m \right] = K \, \E \left[ \lvert \rmX_1 \rvert^m \right] = \Theta(K) $.
        \item Second component: $\left( \sum_{k=0}^K \E \left[ \rmX_k^2 \right] \right)^{m/2} = (K \sigma_X^2)^{m/2} = K^{m/2} \sigma^m$. \\
        As $K \rightarrow \infty$, given $p \geq 2$, the $K^{m/2} \sigma^m$ term strictly dominates the linear term $\Theta(K)$, apply back to \cref{eq:rosenthan_inquality}, we we establish the exact asymptotic bound as stated. \qedhere
    \end{itemize}
\end{proof}
\begin{proof}[Proof of \cref{main:theorem_rank_stable}]
    \textbf{Proof sketch} With a constant initialization variance $\sigma_R$, we estimate the variance in $\rmL$ update. Deriving the scaling factor $\evs$ is fundamentally an algebraic balancing act to neutralize this structural discrepancy, bounding both flows at $\Theta(1)$. Finally, under independence assumptions and \cref{lemma:rosenthal_absolute}, successfully stabilizing the 2nd-order moment (variance) serves as a sufficient condition to generalize and bound high-order absolute moments.

    \textbf{R-branch Forward pass} Consider the intermediate activation vector $\vz = \rmR \vx$ Each vector feature $\vz_k$ is a sum of $n_2$ independent elements.
    
    For simplicity, we first take the $2^{nd}$ moment into account then make a generalization of high-order moments. Apply \cref{lemma:rosenthal_absolute}, given $\Theta_{n_1,m_2}(\sigma_R) = \Theta(1)$:
    \begin{gather}
        \E \left[ \lvert \vz_k \rvert^m \right] = \Theta \left( n_2^{p/2} (\sigma_R)^m \right) \\
        \Rightarrow\E[\vz_k^2] = \Theta(n_2) = \Theta(n_1^{-1}) = \E[\vz^2].
    \end{gather}
    \textbf{L-block update:} Due to sparsity constrain, only intra-block boundaries accumulate gradients, at time step $n$. $\rmL_{i,k}^{(n)} \propto \sum_{t=1}^n \evs \cdot \nabla \vy_i^{(t)} \cdot \vz_k^{(t)}$. With finite $n$ steps and $\Theta(1)$ error gradients, zero-mean distribution, the variance of a \emph{valid element} in $\rmL$ is:
    \begin{align}
        \sigma^{2(n)}_L =  \E [\lvert \rmL_{i,k}^{(n)} \rvert^2] &= \E \left[ \lvert n \cdot s \cdot \nabla \vy_i \cdot \vz_k \rvert^2 \right] \notag \\ 
        &= \Theta \left( n^2 \cdot s^2 \cdot \E [\lvert \nabla \vy_i \rvert^2] \cdot \E [\lvert \vz \rvert^2] \right), \notag \\
        \Rightarrow \sigma^{2}_L &= \Theta(s^2 \cdot \sigma_z^2) = \Theta(s^2 n_1^{-1})
    \end{align}

    \textbf{Output condition for $\vy$: } With assumption that $\vy, \vz_k$ are independent, combined with zero mean distributions: $ \Var(\rmL_{i,k} \vz_k) =  \Var(\rmL_{i,k}) \Var(\vz_k)$.
    
    Each generic formula of output feature $\vy_i = \evs \sum_{k=1}^{n_1} \rmL_{i,k} \vz_k$\footnote{It is indeed a span of $n_1$ elements, we denote from $1$ to $n_1$ for simplicity.} is sum of $n_1$ components. Due to accumulated updates, assume that $\rmL, \vz$ are independent of each other. The output variance is:
    \begin{align}
        \E [\lvert \vy_i \rvert^2] &= \evs^2 \cdot \E \left[ \rmL_{i,k} \vz_k \right] \notag\\ 
            &= \Theta\left( \evs^2 \cdot n_1 \Var(\rmL_{i,k} \vz_k) \right) \notag \\
        \Rightarrow \sigma_y^2 &= \Theta \left( \evs^2 n_1 \cdot  \Var(\rmL_{i,k})  \Var(\vz_k) \right)
    \end{align}
    Since: $\quad \sigma^{2}_L = \Theta(s^2 \cdot \sigma_z^2) = \Theta(s^2 n_1^{-1}), \quad \Var(\vz) = \E[\vz^2]-0 = \E[\vz^2] = \Theta(n_1^{-1})$:
    \begin{equation}
        \sigma_y^2 = \E [\lvert \vy_i \rvert^2] = \Theta(\evs^2 n_1 \cdot \evs^2 n_1^{-1} \cdot n_1^{-1}) = \Theta(\evs^4 \cdot n_1^{-1})
    \end{equation}
    To satisfy the stabilization constrain $\E[\vy_i^2] = 1$, we strictly require $\evs^4 \cdot n_1^{-1} = \textit{const}$, leading to the amplification factor: $\evs \propto n_1^{0.25}$.
    
    \textbf{Backward gradient condition for $\nabla \vx:$}
    With assumption that $\rmR, \vv_k$ are independent, combined with zero mean distributions: $ \Var(\rmR_{i,k} \vv_k) =  \Var(\rmR_{i,k}) \Var(\vv_k)$.

    Let $\vv_k = \sum_{i=1}^{m_1} \rmL_{k,i}^\top \nabla \vy_i$. The input gradient is:
    \begin{equation}
        \nabla \vx_j  = \evs \sum_{k=1}^{m_2} \rmR_{j,k}^\top \vv_k,
    \end{equation}
    With $\Theta(\nabla_{\vy_i}) = 1$, The $2^{nd}$-moment, equivalent to the variance of $\vv_k$ is:
    \begin{equation}
        \sigma_v^2 = \E[\vv_k^2] = \Theta(m_1 \cdot \sigma_\rmL^2 \cdot 1) = \Theta(m_1 n_1^{-1}\evs^2)
    \end{equation}
    The $2^{nd}$-moment of input gradient, summed over $m_2$ terms:
    \begin{align}
        \sigma_{\nabla \vx}^2 &= \E[\nabla \vx_i^2] = \Theta(\evs^2 \cdot m_2 \cdot \sigma_R^2 \cdot \sigma_v^2) \notag \\
        &= \Theta(\evs^2 \cdot m_2 \cdot 1 \cdot m_1 n_1^{-1} \evs^2) = \Theta(\evs^4 n_1^{-1}), \text{ since: } m_1m_2 = M.
    \end{align}
    Similarly, to satisfy the stabilization constrain, the amplification factor: $\evs \propto n_1^{0.25}$.
    
    \textbf{Generalization to $m-th$ moments:} \quad
    Because the internal tensors are formed by summing multiple independent mean-zero random variables, Lemma 1 establishes that the asymptotic scaling of all $m$-th absolute moments ($m \ge 2$) is strictly governed by a power of the variance: $\mathbb{E}[|S|^m] = \Theta(K^{m/2} \sigma^m)$. 
    
    Since we have proved that the variance (2nd moment) of all evaluated tensors ($\vy$ and $\nabla \vx$) is strictly stabilized at $\Theta(1)$, all higher-order $m$-th absolute moments are inherently bounded at $\Theta(1)$ as well. 
\end{proof}

% ==============================================================
\subsection{Expressivity in low effective rank. Proof of \texorpdfstring{\cref{theo:express_approx}}{Expressivity Crossover Theorem}}
\label{sec:app_proof_express}
% ==============================================================

We restate the two competing approximations precisely.
Let $\mset^{(r)}$ denote the manifold of \repname matrices with configuration $(n_1, m_2) = (r, r)$,
and define:
\begin{align}
    \rmW_L &\coloneqq \argmin_{\substack{U = \rmB\rmA \\ \rmB, \rmA^\top \in \sR^{d \times r}}}
               \|\rmW - U\|_F,
    \qquad\qquad\quad
    \rmW_M \coloneqq \argmin_{U \in \mset^{(r)}} \|\rmW - U\|_F.
\end{align}
$\rmW_L$ is the Eckart--Young rank-$r$ truncation; $\rmW_M$ is the optimal \repname approximation
at the same parameter budget.
The proof proceeds in four steps:
\begin{enumerate*}[label=(\roman*)]
    \item show that $\mset^{(r)}$ contains a canonical family of block-structured matrices
        (\cref{lemma:app_exp_mor_manifold});
    \item establish that the residual of any block-rank-1 approximation is orthogonal to the
        approximation itself (\cref{lemma:app_exp_mor_projection});
    \item derive a rank-dependent lower bound on the energy captured by $\rmW_M$;
    \item identify the spectral crossover threshold via the Marchenko--Pastur law\citep{Marchenko_1967}
        (\cref{lemma:exp_rdm_matrix_ppt,corol:app_crossover}). % lemma:mp_covergence,
\end{enumerate*}
% --------------------------------------------------------------
\subsubsection*{Step 1: \repname Contains All Block-Rank-1 Matrices}
% --------------------------------------------------------------

\begin{lemma}[Block-rank-1 inclusion]
\label{lemma:app_exp_mor_manifold}
Let $\rmW \in \sR^{d \times d}$ be partitioned into a uniform $r \times r$ grid of blocks
$\rmW_{jk} \in \sR^{s \times s}$, $s = d/r$.
Denote by $\rmM_{jk} = \vu_{jk}\vv_{jk}^\top$ the optimal rank-$1$ approximation of each block.
Then the block-assembled matrix $\rmM^{(s)} = \sum_{j,k} \rmM_{jk}$ lies exactly in $\mset^{(r)}$:
\begin{equation}
    \rmM^{(s)} \in \mset^{(r)}, \qquad (n_1, m_2) = (r, r).
\end{equation}
\end{lemma}

\begin{proof}
We construct $(\rmL, \rmR)$ explicitly.
Let $\vu'_{jk} = \sigma_{jk}^{1/2}\vu_{jk}$ and $\vv'_{jk} = \sigma_{jk}^{1/2}\vv_{jk}$,
where $\sigma_{jk} = \sigma_1(\rmW_{jk})$.

\paragraph{Factor assignment.}
Assign the $j$-th block of $\rmL \in \gB\gD$ (of size $s \times r$) and the $k$-th block of
$\rmR \in \gB\gD$ (of size $r \times s$) as:
\begin{equation}
    \rmL^{(j)}_{:,k} \leftarrow \vu'_{j,k}, \qquad
    \rmR^{(k)}_{j,:} \leftarrow (\vv'_{j,k})^\top,
    \qquad \forall\, 1 \leq j, k \leq r.
    \label{eq:app_expr_LR_assign}
\end{equation}
Each block of $\rmL$ pools the $r$ left singular vectors associated with its grid row;
each block of $\rmR$ pools the $r$ right singular vectors associated with its grid column.

\paragraph{Inner routing via $\mP_r$.}
The permutation $\mP_r$ acts as a bipartite switchboard: it pairs precisely the $k$-th column
of $\rmL^{(j)}$ with the $j$-th row of $\rmR^{(k)}$, synthesizing the outer product
$\vu'_{jk}(\vv'_{jk})^\top = \rmM_{jk}$.
Over all $r^2$ pairs, $\rmL\mP_r\rmR$ simultaneously generates all $r^2$ rank-1 blocks.

\paragraph{Spatial routing via $\mP_l$.}
The outer permutation $\mP_l$ is a pure reindexing: it relocates the $r^2$ rank-1 blocks
to their correct $(j,k)$ positions in the $r \times r$ grid, recovering $\rmM^{(s)}$ exactly.

Since $\rmL$ and $\rmR$ are unconstrained beyond their block-diagonal structure,
and the permutations are fixed linear bijections, any choice of $\{\vu'_{jk}, \vv'_{jk}\}$
is realizable, establishing $\rmM^{(s)} \in \mset^{(r)}$.
\end{proof}

% --------------------------------------------------------------
\subsubsection*{Step 2 : Orthogonality of Block-Rank-1 Residuals}
% --------------------------------------------------------------

\begin{lemma}[Block-wise Pythagorean identity]
\label{lemma:app_exp_mor_projection}
With $\rmM^{(s)}$ as in \cref{lemma:app_exp_mor_manifold}, the approximation residual
$\rmW - \rmM^{(s)}$ is Frobenius-orthogonal to $\rmM^{(s)}$:
\begin{equation}
    \langle \rmW - \rmM^{(s)},\, \rmM^{(s)} \rangle_F = 0
    \quad\Longrightarrow\quad
    \|\rmW - \rmM^{(s)}\|_F^2 = \|\rmW\|_F^2 - \|\rmM^{(s)}\|_F^2.
    \label{eq:app_exp_pythagoras}
\end{equation}
\end{lemma}

\begin{proof}
The inner product decomposes blockwise:
$\langle \rmW - \rmM^{(s)}, \rmM^{(s)} \rangle_F = \sum_{j,k} \langle \rmW_{jk} - \rmM_{jk}, \rmM_{jk} \rangle_F$.
Each term vanishes because $\rmM_{jk} = \sigma_{jk}\vu_{jk}\vv_{jk}^\top$ is the optimal rank-1
approximation of $\rmW_{jk}$: by Eckart--Young, the residual $\rmW_{jk} - \rmM_{jk}$ lies in
the orthogonal complement of $\mathrm{span}(\vu_{jk}) \otimes \mathrm{span}(\vv_{jk})$, which contains $\rmM_{jk}$.
Summing over all blocks yields \cref{eq:app_exp_pythagoras}.
\end{proof}

% --------------------------------------------------------------
\subsubsection*{Step 3 : Energy Comparison and the Crossover Threshold Existence}
% --------------------------------------------------------------

\textbf{Ideas} We show that each strip of $r$ rows in $\rmW$ contains at leat an energy unit under this specific condition $R > d - s (s = d/r)$, and \repname matrices are able to capture the energy.

\begin{proof}[Proof of \cref{theo:express_approx}]
Define the \emph{energy} of a matrix as its squared Frobenius norm:
$\rmE(\rmW) = \|\rmW\|_F^2 = \sum_i \sigma_i^2(\rmW)$.
Minimizing approximation error is equivalent to maximizing captured energy:
$\|\rmW - \rmW_Q\|_F^2 = \rmE(\rmW) - \rmE(\rmW_Q)$.

\paragraph{Energy ceiling for low-rank.}
By the Eckart--Young--Mirsky theorem, the rank-$r$ truncation captures exactly the $r$
largest singular values.
Since $\rmW$ has rank $R$ with all unit singular values, $\sigma_i(\rmW) = \mathbf{1}[i \leq R]$, so:
\begin{equation}
    \rmE(\rmW_L) = \sum_{i=1}^{r} \sigma_i^2(\rmW) = r.
    \label{eq:app_exp_Elow}
\end{equation}
This is a hard ceiling: no rank-$r$ matrix can capture more than $r$ units of energy,
regardless of the target structure.

\paragraph{Energy floor for \repname.}
Partition $\rmW$ into its $r \times r$ grid of $s \times s$ blocks ($s = d/r$),
and let $\rmM^{(s)}$ be the block-rank-1 approximation from \cref{lemma:app_exp_mor_manifold}.
Since $\rmW_M$ is the global optimum over $\mset^{(r)} \ni \rmM^{(s)}$:
\begin{equation}
    \rmE(\rmW_M) \geq \rmE(\rmM^{(s)}) = \sum_{j=1}^{r}\sum_{k=1}^{r} \sigma_1^2(\rmW_{jk}).
\end{equation}

Consider $\rmW$ as $r$ row strips (by merging horizontally $r$ blocks of $s\text{-by-}s$), where each strip $\rmW^{(j)}_{\text{strip}}$ consists of $s$ rows. Let $\rmW^{(j)}_{\text{rest}}$ denote the remaining $d-s$ rows of $\rmW$. We analyze the energy captured by \repname within a single strip.

Define two vector space in $\sR^d$:
\begin{itemize}
    \item $\ermA = \row{(\rmW)}$. Since $\rankop(\rmW) = R$, we have $\dim(A)=R.$
    \item $\ermB = \nul(\rmW^{(j)}_{\text{rest}}).$ By the Rank-Nullity theorem:
    \begin{equation}
        \dim(\ermB)=d-\rankop(\rmW^{(j)}_{\text{rest}}) \geq d - (d-s) = s.
    \end{equation}
\end{itemize}
    By Grassman's formula, the intersection of these subspaces satisfies:
    \begin{equation}
        \dim(\ermA \cap \ermB) \geq \dim(\ermA) + \dim(\ermB) - \dim(\ermA \cup \ermB) \geq R + s - d.
    \end{equation}
    If we impose the condition $R > d -s$, then $\dim(\ermA \cap \ermB) \geq 1$. Thus, there exists a non-zero vector $\vx \in \ermA \cap \ermB$. Assume $\vx$ is normalized such that $\lVert \vx \rVert = 1$.

    Since $\vx \in \ermA = \row(\rmW)$, is is strictly orthogonal to $\nul(\rmW)$. Because all non-zero singular values of $\rmW$ are equal to 1, $\rmW$ acts as an isometry on its row space, preversing the vector energy completely: $\lVert \rmW \vx \rVert = \lVert x \rVert = 1$.

    Simultaneously, since $\vx \in \ermB = \nul(\rmW^{(j)}_{\text{rest}})$, we have $\rmW^{(j)}_{\text{rest}}\vx = 0$.

    The squared norm of the full transformation $\rmW\vx$ decomposes exactly over the strip the the rest of the matrix:
    \begin{equation}
        \lVert \rmW^{(j)}_{\text{strip}} \vx \rVert^2 + \lVert \rmW^{(j)}_{\text{rest}} \vx \rVert^2 = \lVert \rmW \vx \rVert^2 \Rightarrow \lVert \rmW^{(j)}_{\text{strip}} \vx \rVert^2 + 0 = 1.
    \end{equation}
    By the definition of the spectral norm, this implies $\sigma_1^2(\rmW^{(j)}_{\text{strip}}) \ge 1$.

\textit{Regarding \repname parameterization.\quad } Now partition the strip  $\rmW^{(j)}_{\text{strip}}$ intos $r$ constituent blocks of $(s\times s)$: $\rmW_{j1}, \rmW_{j2}, \dots, \rmW_{jr}$.
    \begin{lemma}[Block spectral norm]
    \label{lemma:block_spectral_norm_ine}
    Let $\rmW_j$ is a matrix and its partitioned horizontally into $r$ sub-matrix: $\rmW_j = [\rmW_{j1}, \rmW_{j2}, \dots, \rmW_{jr}]$ (as being pre-defined in the section above).

    The squared spectral norm of the concatenated matrix $\rmW_j$ is bounded the by sum of the squared spectral norm of its individual blocks:
    \begin{equation}
        \sigma_1^2{\left( \rmW^{(j)}_{\text{j}} \right)}  \leq \sum_{k=1}^{r}\sigma_1^2(\rmW_{jk}).
    \end{equation}
    \end{lemma}
    
    \textit{Proof schetch:} We show that by spectral norm properties and Cauchy-Schwarz inequality.

    \begin{proof}[Proof of \cref{lemma:block_spectral_norm_ine}]
        By the definition of the spectral norm, there exists a unit vector $\vy$ with $\lVert \vy\rVert = 1$ such that $\sigma_1^2(\rmW_j) = \lVert \rmW_j \vy \rVert^2$. We partition $\vy$ into $r$ segments $\vy = [\vy_1^\top, \vy_2^\top, \dots, \vy_r^\top]^\top$ conforming to the structure of $\rmW_j$, which implies $\sum_{k=1}^r \Vert \vy_k\Vert^2 = 1$.

        Decomposing the linear transformation and applying the triangle inequality, we obtain:
        \begin{equation}
            \lVert \rmW_j \vy\rVert = \lVert \sum_{k=1}^r \rmW_{jk} \vy_k \rVert \le \sum_{k=1}^r \lVert \rmW_{jk} \vy_k\Vert \le \sum_{k=1}^r \sigma_1(\rmW_{jk}) \Vert \vy_k\rVert.
        \end{equation}

        Squaring both sides and applying the Cauchy-Schwarz inequality yields:
        \begin{equation}
            \begin{aligned}
            \lVert \rmW_j \vy \rVert^2 &\le \left( \sum_{k=1}^r \sigma_1^2(W_k) \right) \left( \sum_{k=1}^r \lVert \vy_k\rVert^2 \right),\\
            \Rightarrow \sigma_1^2(\rmW_j) &= \sigma_1^2{(\rmW^{(j)}_{\text{strip}})} \leq \sum_{k=1}^{r}\sigma_1^2(\rmW_{jk}),
            \end{aligned}
        \end{equation}
        Since $\sum_{k=1}^r \lVert \vy_k\rVert^2 = 1$. Here we complete the proof
    \end{proof}

By applying \cref{lemma:block_spectral_norm_ine}:
\begin{equation}
    \sum_{k=1}^{r}\sigma_1^2(\rmW_{jk}) \geq \sigma_1^2{(\rmW^{(j)}_{\text{j}})}.
\end{equation}

Since the \repname approximation $\rmW_M$ can independently formulate an optimal rank-1 approximation for every block (established in \cref{lemma:app_exp_mor_manifold}), it captures at least this much energy per strip. Summing over all $r$ strips yields:
\begin{equation}
    \rmE(\rmW_M) \ge \sum_{j=1}^r \sum_{k=1}^r \sigma_1^2(\rmW_{jk}) \ge r.
\end{equation}

\paragraph{Crossover condition.} Combining the ceiling and the floor, we conclude that $\rmE(\rmW_M) \ge \rmE(\rmW_L)$ almost surely hold whenever $R>d-s$. This establishes a deterministic worst-case threshold $R_0 \leq d -s$, confirming existence of a nontrivial finite crossover.

However, because this worse-case bound is still extremely close to d, we next employ random matrix theory to derive the average-case crossover threshold, which governs practical deep learning scenarios.
\end{proof}

% --------------------------------------------------------------
\subsubsection*{Step 4 : Quantifying the Crossover via Random Matrix Theory}
% --------------------------------------------------------------
The proof above establishes existence of $R_0$, but for the worse-case.
We now investigate on practical application on random matrix using the Marchenko--Pastur law~\citep{Marchenko_1967} and the Bai-Yin theorem~\citep{bai1988necessary}.

\begin{lemma}[Recall of matrix norms for random matrix]
    \label{lemma:exp_rdm_matrix_ppt}
    Let $\rmS \in \sR^{s \times s}$ have i.i.d.\ $\mathcal{N}(0, \sigma_S^2)$ entries. The following statements hold:
    \begin{itemize}
        \item $\lVert \rmS \rVert^2_F = s^2 \sigma_S^2$.
        \item $\sigma_1^2(\rmS) \xrightarrow{\,s\to\infty\,} C(s)\sigma_S^2$, where $C(s)$ is a constant depending on size $s$ only. When $s$ is sufficiently large, $C(s) \to 4$.
    \end{itemize}
\end{lemma}

\begin{proof}[Proof of \cref{lemma:exp_rdm_matrix_ppt}]
    By the law of large numbers, $\|\rmS\|_F^2 = \sum_{ij} S_{ij}^2 \to s^2 \sigma_S^2$ as $s \to \infty$.

    For the spectral norm, treat $\rmG = s^{-1}\rmS\rmS^\top$ as a sample covariance matrix
with aspect ratio $\lambda = \nicefrac{s}{s} = 1$. The Marchenko--Pastur law~\citep{Marchenko_1967} places eigenvalues $\lambda_i(\rmG)$ in
$[\sigma_S^2(1-\sqrt{\lambda})^2,\, \sigma_S^2(1+\sqrt{\lambda})^2]$; at $\lambda=1$ the
upper edge is $4\sigma_S^2$, so $\lambda_{\max}(\rmG) \to 4\sigma_S^2$, giving:
    \begin{equation}
    \sigma_1^2(\rmS) = s\,\lambda_{\max}(\rmG) \xrightarrow{\,s\to\infty\,} 4s\sigma_S^2.
    \end{equation}
    From now on, we refer $C(s) \leftrightarrow C$ exchangably for simplicity.

    \textbf{Remark 1 (Finite-size Corrections and Tracy-Widom Fluctuations).} 
    While the Bai-Yin theorem establishes $C(s) \to 4$ in the asymptotic limit ($s \to \infty$), practical \repname configurations employ finite block sizes (e.g., $s \in [64, 512]$ for $d=4096$). For finite-dimensional random matrices, the largest eigenvalue exhibits statistical fluctuations around the Marchenko-Pastur upper edge, governed by the Tracy-Widom distribution \cite{johnstone2001distribution}.
\end{proof}

\paragraph{Step 4a: Asymptotic Gaussianity of Local Windows.}
While the worst-case bound $R_0 \le d - s$ assumes an adversarial energy distribution, we know concentrate on matrices exhibitting random-like properties. Let $\rmW = \rmU\rmV^\top$ where $\rmU, \rmV \in \mathbb{R}^{d \times R}$ are drawn from the Haar measure on the Stiefel manifold. Consider a local $s \times s$ sub-block $\rmW_{jk} = \rmU_j \rmV_k^\top$, where $\rmU_j, \rmV_k \in \sR^{s \times R}$ are the corresponding row slices, $j,k \in \overline{1..r}$. Each entry of this block is computed as an inner product of $R$-dimensional vectors:

\begin{equation}
    (\rmW_{jk})_{ab} = \left\langle \rmU_{j,a,:}, \rmV_{k,b,:} \right\rangle.
\end{equation}

Because the full matrices $\rmU$ and $\rmV$ have orthonormal columns distributed uniformly in $\mathbb{R}^d$, the expected squared magnitude of any single element is $\E[\rmU_{j,ac}^2] = \E[\rmV_{k,bc}^2] = 1/d$. Since $\rmU$ and $V$ are independent, the variance of their scalar product is $\E[\rmU\rmV] = 1/d^2$. Summing over $R$ independent components yields the theoretical variance for the entries of the sub-block:
\begin{equation}
    \sigma_H^2 = \Var\left( (\rmW_{jk})_{ab} \right) = R \left( \frac{1}{d^2} \right) = \frac{R}{d^2}.
\end{equation}

\begin{figure}[htbp]
	\centering
	\includegraphics[width=0.98\linewidth]{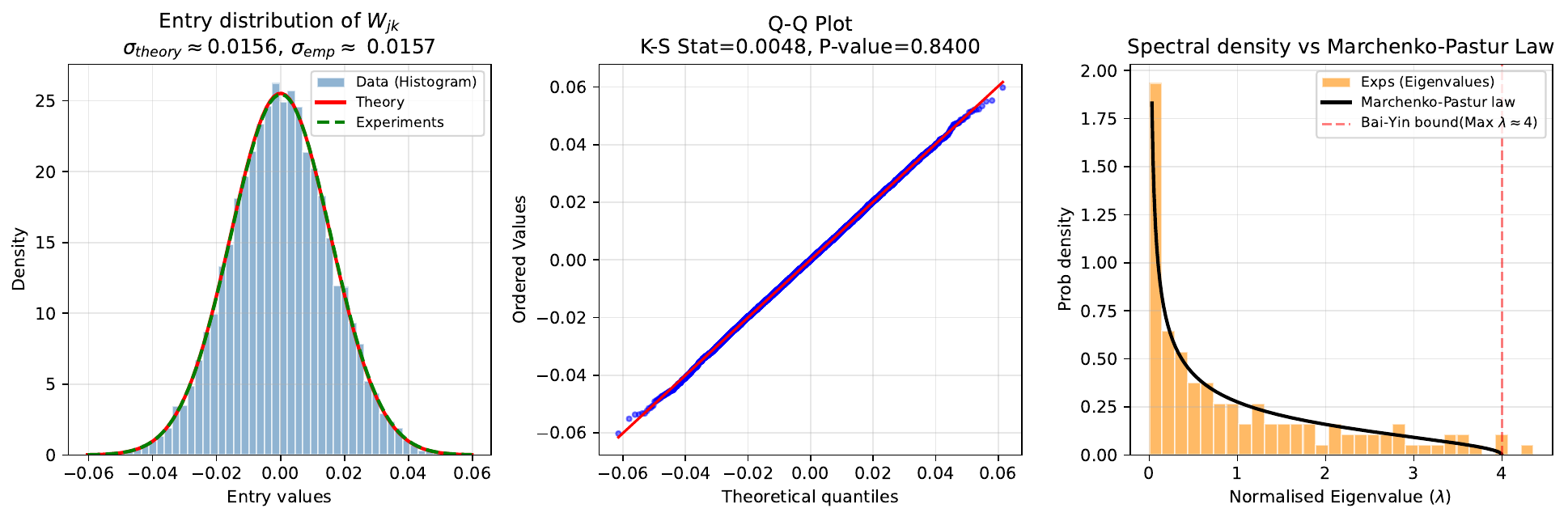}
	\caption{\label{fig:exp_sbys_3subfigs} {Statistical and spectral validation of local sub-blocks $\rmW_{jk}$ extracted from a Haar-distributed target matrix.} \textbf{(Left)} Empirical probability density of the entries from an $s \times s$ sub-block  $\rmW_{jk}$. The empirical distribution closely aligns with the theoretical Gaussian density $\mathcal{N}(0, R/d^2)$. \textbf{(Middle)} The Q-Q plot confirms the asymptotic normality of the entries, corroborated by a high Kolmogorov-Smirnov test p-value. \textbf{(Right)} The empirical spectral density of the normalized sample covariance matrix perfectly follows the theoretical Marchenko-Pastur law. The maximum eigenvalue strictly adheres to the Bai-Yin upper bound at $\lambda \approx 4$, establishing the rigorous mathematical foundation for Morpho's average-case energy capture threshold at $R_0 \approx d/4$.}
\end{figure}

By classical results on Haar-distributed matrices and the Central Limit Theorem, when $R$ increases, the entries of $\rmW_{jk}$ asymptotically converge to independent and identically distributed (i.i.d.) Gaussian variables $\mathcal{N}(0, \sigma_H^2)$.
The deviation from the true Gaussian measure in total variation distance diminishes rapidly when $R,d$ sufficiently large. We empirically show this effect on a local window in \cref{fig:exp_sbys_3subfigs}, confirming our evidence for random-like matrix analysis.

\textbf{Step 4b: Marchenko-Pastur Law and Average-Case Crossover.}
Because $W_{jk}$ behaves as a random Gaussian matrix, the spectrum of its normalized sample covariance follows the Marchenko-Pastur law with an aspect ratio $\lambda = s/s = 1$. According to the Bai-Yin theorem~\citep{bai1988necessary}, the maximum singular value squared of the block, $\sigma_1^2(W_{jk})$, converges almost surely to the upper edge of the Marchenko-Pastur support, with respect to \cref{lemma:exp_rdm_matrix_ppt}:
\begin{equation} 
    \sigma_1^2(\rmW_{jk}) \approx C s \sigma_H^2 = C s\sigma_H^2 = C s \frac{R}{d^2}.
\end{equation}
where constant $C \xrightarrow{\,s\to\infty\,} (1+\sqrt{\lambda})^2 = 4$. 

Recall from \cref{lemma:app_exp_mor_manifold} that the Morpho parameterization can independently formulate an optimal rank-1 approximation for every block. Consequently, the total expected energy captured by Morpho across all $r^2$ blocks is:

\begin{equation}
    \rmE(\rmW_M) \geq \sum_{j=1}^r \sum_{k=1}^r \sigma_1^2(\rmW_{jk}) \approx r^2 \left( Cs \frac{R}{d^2} \right) = C r \frac{R}{d}.
\end{equation}

For Morpho to strictly outperform the LoRA baseline, its captured energy must exceed LoRA's deterministic energy ceiling ($\rmE(\rmW_L) = r$). Solving $\rmE(\rmW_M) > \rmE(\rmW_L)$ yields:
\begin{equation}
    \label{eq:exp_R_threshold_ine}
    Cr \frac{R}{d} > r \Longleftrightarrow R > \frac{d}{C} % \implies R_0 \approx \frac{d}{4}.
\end{equation}

This average-case threshold demonstrates that under standard random initialization or well-regularized weight distributions, Morpho efficiently surpasses LoRA at approximately $25\%$ of the full rank. This theoretical derivation tightly mirrors our empirical observations of the crossover point in practical settings in \cref{fig:main_express_2bl,fig:combined_experiments_uv_compress}.

\begin{corollary}[Universal crossover bound]
\label{corol:app_crossover}
With respect to the above settings, the crossover threshold $R_0$ satisfies:
\begin{equation}
    R_0 = \lceil \frac{d}{C} \rceil,
\end{equation}
and is \emph{independent of the budget rank $r$}.
Consequently, for any fixed matrix dimension $d$, the fraction of targets for which \repname
dominates low-rank approximation is at least $1 - 1/C$, converging to $0.75$, as $d \to \infty$.
\end{corollary}

\begin{proof}
This is a direct corollary from \cref{eq:exp_R_threshold_ine} and \cref{lemma:exp_rdm_matrix_ppt} by means of the Marchenko--Pastur law~\citep{Marchenko_1967} and the Bai-Yin theorem~\citep{bai1988necessary}.

There is no dependence on budget $r$, making $R_0$ a function of $d$ and $C$ alone. As a result of \cref{lemma:exp_rdm_matrix_ppt}, the actually threshold might vary since $C$ follows Tracy-Widom distribution~\citep{johnstone2001distribution}.
\end{proof}

\paragraph{Synthesis.}
\Cref{corol:app_crossover} reveals a structurally important universality: the rank at which
\repname overtakes low-rank approximation is determined entirely by the ambient dimension $d$
and the spectral concentration constant $C$—not by the parameter budget $r$.
This means that as models scale (larger $d$), the crossover threshold grows proportionally,
but the \emph{fraction} of high-eRank targets for which \repname dominates remains constant.
In the regime of LLMs, where weight matrices are large and their updates are empirically
observed to have broadly distributed singular spectra~\citep{sharma2023truth},
this universality directly predicts the practical advantage of \repname observed in
\cref{sec:sim_exp_exp}.

\subsection{Numerical stability of \ftmt over ranks}
\label{sec:app_rank_stable_exp}
% ==============================================================

\Cref{main:theorem_rank_stable} establishes the optimal scaling factor for \ftmt as:
\begin{equation}
    \evs_{\ftmrm} = \Theta\!\left(n_1^{\nicefrac{1}{4}}\right),
\end{equation}
which ensures numerical stability of output and gradient magnitudes as the hidden dimension
grows.
We now empirically validate this prescription against a grid of different scaling strategies.

\paragraph{Experimental setup.}
We fix matrix dimensions $M = N = 2048$ and a constant learning rate, then independently vary
$n_1$ (with $m_2$ fixed) and $m_2$ (with $n_1$ fixed), monitoring output norms and gradient
magnitudes under six scaling schemes parameterized by the free variable $v \in \{n_1, m_2\}$:
\begin{equation}
    \evs_{\ftmrm} \in \left\{
        \Theta(1),\;
        \Theta\!\left(v^{-1}\right),\;
        \Theta\!\left(v^{-\nicefrac{1}{2}}\right),\;
        \Theta\!\left(v^{-\nicefrac{1}{4}}\right),\;
        \Theta\!\left(v^{\nicefrac{1}{4}}\right),\;
        \Theta\!\left(v^{\nicefrac{1}{2}}\right)
    \right\}.
\end{equation}

\begin{figure}[htbp]
    \centering
    \includegraphics[width=0.99\linewidth]{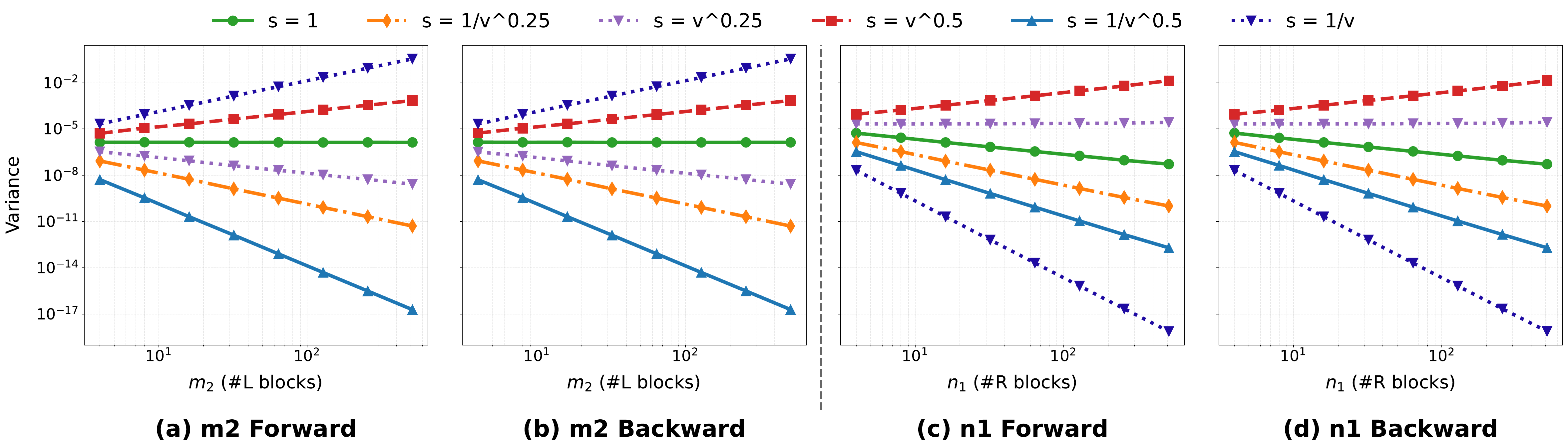}
    \caption{Controlled study of output and gradient magnitudes under six scaling schemes
    for $\evs_{\oftmto}$, across two hyperparameter axes.
    \textbf{(a,b):} $n_1$ fixed, $m_2$ varying —
    \textbf{\textcolor{SpringGreen4}{$\evs = \Theta(1)$ (green)}} is the unique stable scheme.
    \textbf{(c,d):} $m_2$ fixed, $n_1$ varying —
    \textbf{\textcolor{MediumPurple3}{$\evs = \Theta(n_1^{0.25})$ (purple)}} is the unique
    stable scheme.
    All other schemes exhibit monotone attenuation or divergence.
    }
    \label{fig:app_rs_m2n1_example}
\end{figure}

\paragraph{Results.}
\Cref{fig:app_rs_m2n1_example} reports the outcome across both variation axes, with y-axes represent variance of output $\vy$ (forward pass) and gradient $\nabla_\gL\vx$ (backward pass), respectively.
Two scaling schemes maintain stable signal propagation; all others exhibit either
systematic attenuation or divergence as the free parameter grows:
\begin{itemize}[leftmargin=1.5em]
    \item \textbf{Panels (a,b) — varying $m_2$, fixed $n_1$:}
    the constant scheme
    \textbf{\textcolor{SpringGreen4}{$\evs = \Theta(1)$}} (green)
    is the unique stable choice, consistent with the theoretical prediction that
    $\evs$ has no dependence on $m_2$.
    \item \textbf{Panels (c,d) — varying $n_1$, fixed $m_2$:}
    the quarter-power scheme
    \textbf{\textcolor{MediumPurple3}{$\evs = \Theta(n_1^{\nicefrac{1}{4}})$}} (purple)
    is the unique stable choice, directly confirming \cref{main:theorem_rank_stable}.
\end{itemize}

\begin{figure}[htbp]
    \centering
    \includegraphics[width=0.85\linewidth]{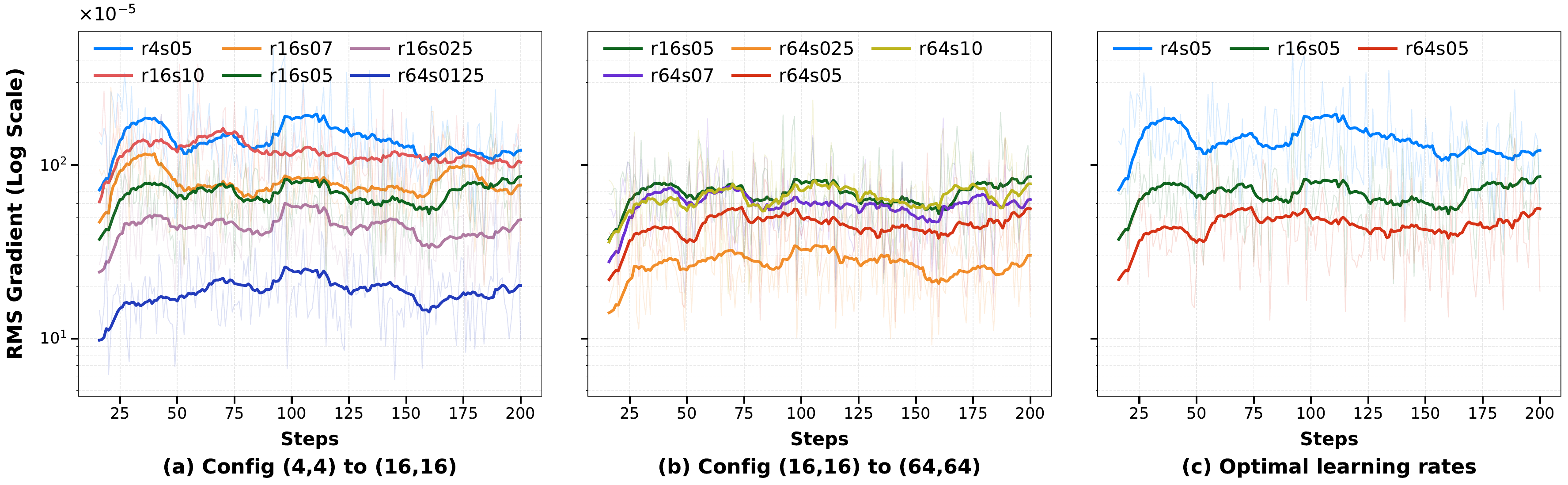}
    \caption{Observation of moving average of gradient norms over settings, where $(n_1=m_2=r)$, $s$ denotes scaling factors. \textbf{(a,b)} illustrate gradient norms when $(m_1,n_1)$ changes from $(4,4)$ to $(16,16)$ and $(16,16)$ to $(64,64)$, respectively. The rule in \cref{main:theorem_rank_stable} are closely approximated. \textbf{(c)} The gradients of the empirical optimal training configurations for each rank. Higher parameter counts tend to lower the gradient norms. This further illustrate the point to distuinguish between stability and feature learning.}
    \label{fig:app_rs_grad}
\end{figure}

The asymmetry between the two axes clearly show that stability is governed by $n_1$ alone,
not by $m_2$, which corroborates the structural interpretation of \cref{main:theorem_rank_stable}. That is,
the $n_1^{\nicefrac{1}{4}}$ factor compensates for the growth in the operator norm of
$\mP_l^\top\rmL\mP_r\rmR$ as the number of right-factor blocks increases,
while $m_2$ affects only the rank of the adapter (\cref{theo_sama_rank}) and not its
initialization scale.

We also provide practical obervation of gradient norms in \cref{fig:app_rs_grad}. The rule for the scaling factor suggest a means to stablize gradient norms at close values when $n_1$ varies. However, as the limitation note in \cref{sec:app_rank_stable}, while this theoretical rule strictly aligns the gradient scales, it acts primarily as a mathematical safeguard. In optimal empirical scenarios, strictly equating the gradient norm is not always desirable.

% ==============================================================
\subsection{\ftmt Converges to the FFT Solution}
\label{app:proof_converges_fft}
% ==============================================================
\paragraph{Convergence to a full fine-tuning solution.}
To understand the solution quality of \ftmt, we analyze a canonical linear
least-squares (LSQ) setting.  Given input $\rmX \in \sR^{b \times N}$,
target $\rmZ \in \sR^{b \times M}$, and frozen weights $\rmW_0$, the
fine-tuning objective is:
\begin{equation}
  \min_{\Delta\rmW}\;
  \frac{1}{2}
  \bigl\|\rmZ - \rmX\rmW_0^\top - \rmX\Delta\rmW^\top\bigr\|_F^2.
  \label{eq:method_full_lsq}
\end{equation}

\begin{theorem}
    \label{theorem:main_lsq}
    [Sparsity] Given any generic rank-$\rho$ matrix $\mX$, if \ftmt configuration: $n_1 \geq \rho$, then any corresponding solution $\Delta \rmW_{\ftmrm}$ to the LSQ problem also solves the full-LSQ in \cref{eq:method_full_lsq}.

\end{theorem}

The theorem asserts that, whenever $n_1 \geq \rankop(\rmX)$, the \ftmt-constrained LSQ
problem admits solutions that are simultaneously optimal for the unconstrained (FFT) problem.
We prove this by reducing the block-diagonal structure of \ftmt to a collection of independent
standard LSQ problems, then showing that their closed-form solutions produce a zero gradient
in the full unconstrained objective.

% --------------------------------------------------------------
\subsubsection*{Proof Setup}
% --------------------------------------------------------------

Within the scope of this theorem, $\rmX \in \sR^{b \times N}$ and $\rmZ \in \sR^{b \times M}$
are batches of row vectors.
We optimize the \repname factors $(\rmL, \rmR)$ directly, absorbing the scaling $\evs$ into
the initialization:
\begin{align}
    \label{eq:app_lsq_morpho}
    \min_{\Delta\rmW \in \mset} \;
    \frac{1}{2}\left\lVert \rmZ - \rmX\rmW_0^\top - \rmX\Delta\rmW^\top \right\rVert_F^2
    \;\equiv\;
    \min_{(\rmL,\rmR)\,\in\,\mset_g} \;
    \frac{1}{2}\left\lVert \rmZ - \rmX\rmW_0^\top - \rmX\rmR^\top\mP_r^\top\rmL^\top\mP_l \right\rVert_F^2.
\end{align}

Fix an generic $\rmR_t \coloneqq \rmR^\top, \rankop(\rmR_t) = n_1m_2$ and introduce the substitutions:
\begin{equation}
    \rmY \coloneqq (\rmZ - \rmX\rmW_0^\top)\mP_l^\top \in \sR^{b \times M},
    \qquad
    \rmT \coloneqq \rmX\rmR_t\mP_r^\top \in \sR^{b \times Q},
    \qquad
    \rmL_t \coloneqq \rmL^\top \in \sR^{Q \times M},
\end{equation}
where $Q = n_1 m_2$.
The problem in \cref{eq:app_lsq_morpho} reduces to:
\begin{equation}
    \label{eq:app_lsq_rewrite_morpho}
    \min_{\rmL_t \,\in\, \gB\gD} \;
    \frac{1}{2}\left\lVert \rmY - \rmT\rmL_t \right\rVert_F^2.
\end{equation}

% --------------------------------------------------------------
\subsubsection*{Step 1: Block Decomposition into Independent Subproblems}
% --------------------------------------------------------------

Since $\rmL_t = \diag(\rmL_1, \ldots, \rmL_{m_2})$ with $\rmL_j \in \sR^{n_1 \times m_1}$,
partition all matrices column-wise into $m_2$ vertical slices:
\begin{equation}
    \rmY = \begin{bmatrix}\rmY_1 & \cdots & \rmY_{m_2}\end{bmatrix},
    \quad
    \rmT = \begin{bmatrix}\rmT_1 & \cdots & \rmT_{m_2}\end{bmatrix},
    \quad
    \rmY_j \in \sR^{b \times m_1},\;
    \rmT_j \in \sR^{b \times n_1}.
\end{equation}
The coupled problem in \cref{eq:app_lsq_rewrite_morpho} decouples exactly into $m_2$
independent standard LSQ problems:
\begin{equation}
    \label{eq:app_lsq_block}
    \min_{\rmL_j} \; \frac{1}{2}\left\lVert \rmY_j - \rmT_j\rmL_j \right\rVert_F^2,
    \qquad j = 1, \ldots, m_2,
\end{equation}
each admitting the closed-form minimizer~\citep{golub2013matrix}:
\begin{equation}
    \label{eq:app_lsq_simplified}
    \rmL_j^\ast = \rmT_j^\dagger\rmY_j.
\end{equation}
Assembling these gives $\rmL_t^\ast = \diag(\rmL_1^\ast, \ldots, \rmL_{m_2}^\ast)$,
the solution to \cref{eq:app_lsq_rewrite_morpho}.

% --------------------------------------------------------------
\subsubsection*{Step 2 : The Assembled Solution is a Global Minimizer of the Full LSQ}
% --------------------------------------------------------------

Define $\Delta\rmW^\ast \coloneqq \rmR_t\mP_r^\top\rmL_t^\ast\mP_l$.
The gradient of the full unconstrained objective with respect to $\Delta\rmW^\top$ is:
\begin{equation}
    \rvg_{\Delta\rmW^\ast}
    = -\rmX^\top(\rmZ' - \rmX\rmR_t\mP_r^\top\rmL_t^\ast\mP_l)
    = -\rmX^\top(\rmY - \rmT\rmL_t^\ast)\mP_l,
\end{equation}
where $\rmZ' = \rmZ - \rmX\rmW_0^\top$.
It suffices to show $\rmG_\Delta \coloneqq -\rmX^\top(\rmY - \rmT\rmL_t^\ast) = \zeromatrix_{N \times M}$.

\paragraph{Rank factorization of $\rmT$.}
Let $\rho = \rankop(\rmX)$.
Since $\mP_r^\top$ is orthogonal (full rank) and $\rankop(\rmR_t) = n_1 m_2$
by \cref{theo_sama_rank}, the rank of $\rmT$ is determined by the bottleneck:
\begin{equation}
    \rankop(\rmT) = \rankop(\rmX\rmR_t\mP_r^\top) = \rankop(\rmX) = \rho.
\end{equation}
By SVD, write $\rmT = \rmO\rmQ$ where $\rmO \in \sR^{b \times \rho}$ has orthonormal
columns ($\rmO^\top\rmO = \mI_\rho$, so $\rmO^\dagger = \rmO^\top$) and $\rmQ \in \sR^{\rho \times Q}$
has orthonormal rows.
Slice column-wise: $\rmT_j = \rmO\rmQ_j$, $\rmQ_j \in \sR^{\rho \times n_1}$.
Since $\rho \leq n_1$ by assumption, $\rmQ_j$ is full row-rank due to generic matrices, giving:

\begin{equation}
    \label{eq:app_theo3_rightinverse}
    \rmQ_j\rmQ_j^\dagger = \mI_\rho.
\end{equation}
Because $\rmO$ is full column-rank and $\rmQ_j$ is full row-rank, the pseudoinverse distributes
over their product~\citep{greville1966productofpseudo}:
\begin{equation}
    \label{eq:app_theo3_distrib}
    \rmT_j^\dagger = (\rmO\rmQ_j)^\dagger = \rmQ_j^\dagger\rmO^\top.
\end{equation}

\paragraph{Gradient vanishes blockwise.}
Partition $\rmX$ into $n_2$ horizontal slices $\rmX_i \in \sR^{b \times n_1}$.
The $(i,j)$-th block of $\rmG_\Delta$ is:
\begin{align}
    \rmG_{\Delta_{i,j}}
    &= \rmX_i^\top\rmT_j\rmL_j^\ast - \rmX_i^\top\rmY_j \notag \\
    &= \rmX_i^\top(\rmO\rmQ_j)(\rmQ_j^\dagger\rmO^\top)\rmY_j - \rmX_i^\top\rmY_j \notag \\
    &= \rmX_i^\top\rmO(\rmQ_j\rmQ_j^\dagger)\rmO^\top\rmY_j - \rmX_i^\top\rmY_j \notag \\
    &= \rmX_i^\top\rmO\mI_\rho\rmO^\top\rmY_j - \rmX_i^\top\rmY_j \notag \\
    &= \rmX_i^\top(\rmO\rmO^\top)\rmY_j - \rmX_i^\top\rmY_j.
    \label{eq:app_theo3_grad0}
\end{align}
Since $\rmO$ has orthonormal columns spanning $\mathrm{col}(\rmT) = \mathrm{col}(\rmX)$,
the projection $\rmO\rmO^\top$ is the orthogonal projector onto $\mathrm{col}(\rmX)$.
Because $\rmX_i^\top$ maps into $\mathrm{col}(\rmX)^\perp$'s annihilator,
$\rmX_i^\top\rmO\rmO^\top = \rmX_i^\top$, and thus:
\begin{equation}
    \rmG_{\Delta_{i,j}} = \rmX_i^\top\rmY_j - \rmX_i^\top\rmY_j = \zeromatrix_{n_1 \times m_1},
    \qquad \forall\, i, j.
\end{equation}
Hence $\rmG_\Delta = \zeromatrix_{N \times M}$, and $\Delta\rmW^\ast$ is a stationary point—
and therefore a global minimizer—of the full LSQ objective.

% --------------------------------------------------------------
\subsubsection*{Step 3 : Extension to the Full Solution Set}
% --------------------------------------------------------------

The argument above uses the minimum-norm solution $\rmL_j^\ast = \rmT_j^\dagger\rmY_j$.
When $\rho < n_1$, the system $\rmT_j\rmL_j = \rmY_j$ is underdetermined and the complete
solution set is~\citep{golub2013matrix}:
\begin{equation}
    \rmL_j^\ast = \rmT_j^\dagger\rmY_j + (\mI_{n_1} - \rmT_j^\dagger\rmT_j)\mU,
    \qquad \mU \in \sR^{n_1 \times m_1}.
\end{equation}
The null-space term drops out of the gradient:
\begin{align}
    \rmX_i^\top\rmT_j\rmL_j^\ast
    &= \rmX_i^\top\rmY_j + \rmX_i^\top\rmT_j(\mI_{n_1} - \rmT_j^\dagger\rmT_j)\mU \notag \\
    &= \rmX_i^\top\rmY_j + \rmX_i^\top(\rmT_j - \rmT_j\rmT_j^\dagger\rmT_j)\mU \notag \\
    &= \rmX_i^\top\rmY_j,
    \label{eq:app_theo3_ex_grad0}
\end{align}
using $\rmT_j\rmT_j^\dagger\rmT_j = \rmT_j$.
The gradient vanishes identically across the entire solution set. \qed

% --------------------------------------------------------------
\subsubsection*{Empirical Verification}
% --------------------------------------------------------------

\begin{figure}[htbp]
    \centering
    \includegraphics[width=0.65\linewidth]{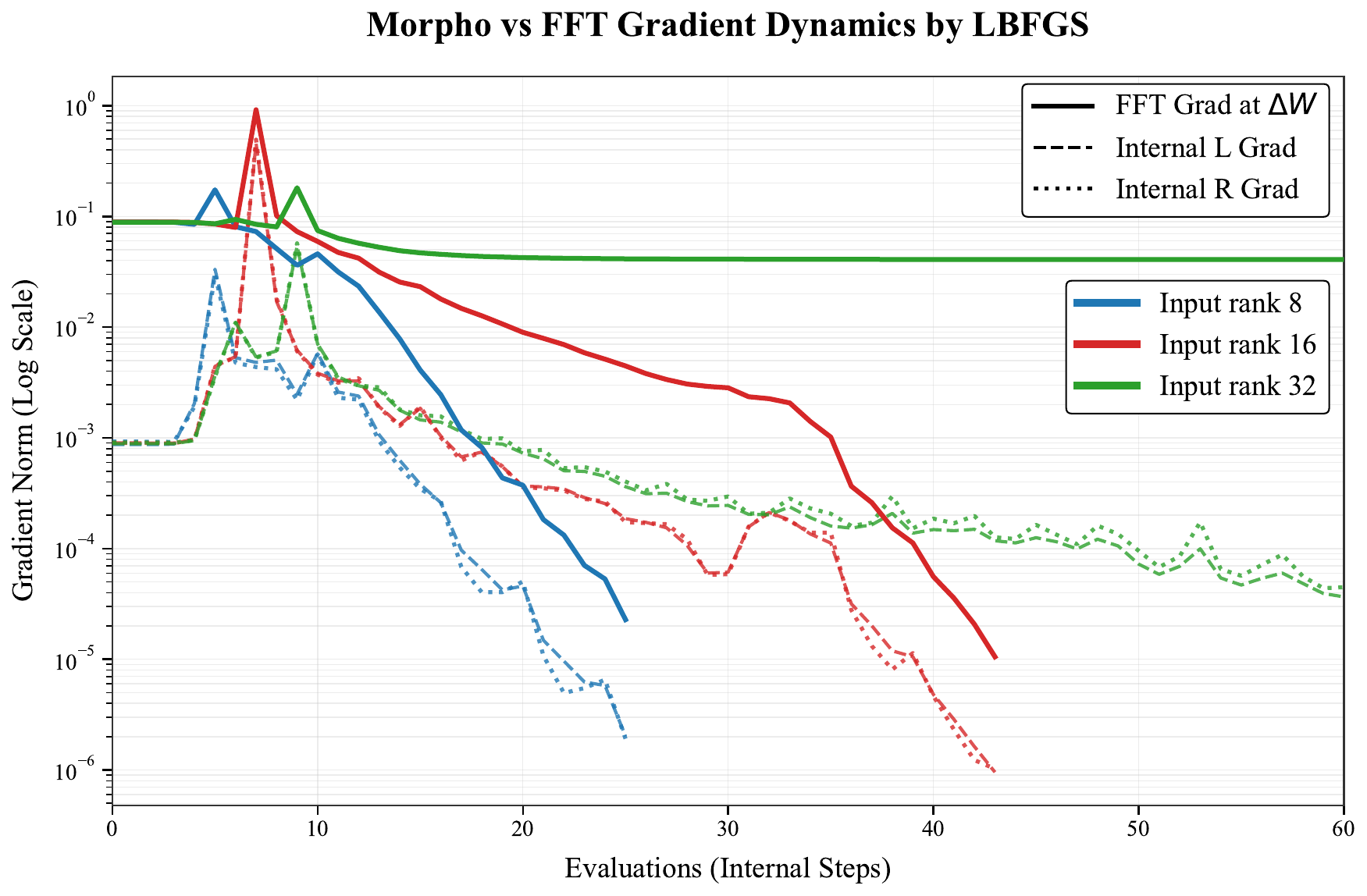}
    \caption{Gradient dynamics of \repname factors $(\rmL, \rmR)$ and the FFT baseline
    over training, at adapter ranks $\{8, 16, 32\}$.
    When $n_1 \geq \rankop(\rmX)$, the \repname gradient norm tracks the FFT gradient
    norm closely throughout optimization, confirming \cref{theorem:main_lsq}.
    Divergence emerges only when $n_1 < \rankop(\rmX)$, marking the rank threshold
    predicted by the theorem.}
    \label{fig:app_theo_morphor_fft_grad_dynamic}
\end{figure}

\begin{figure}[htbp]
    \centering
    \includegraphics[width=0.65\linewidth]{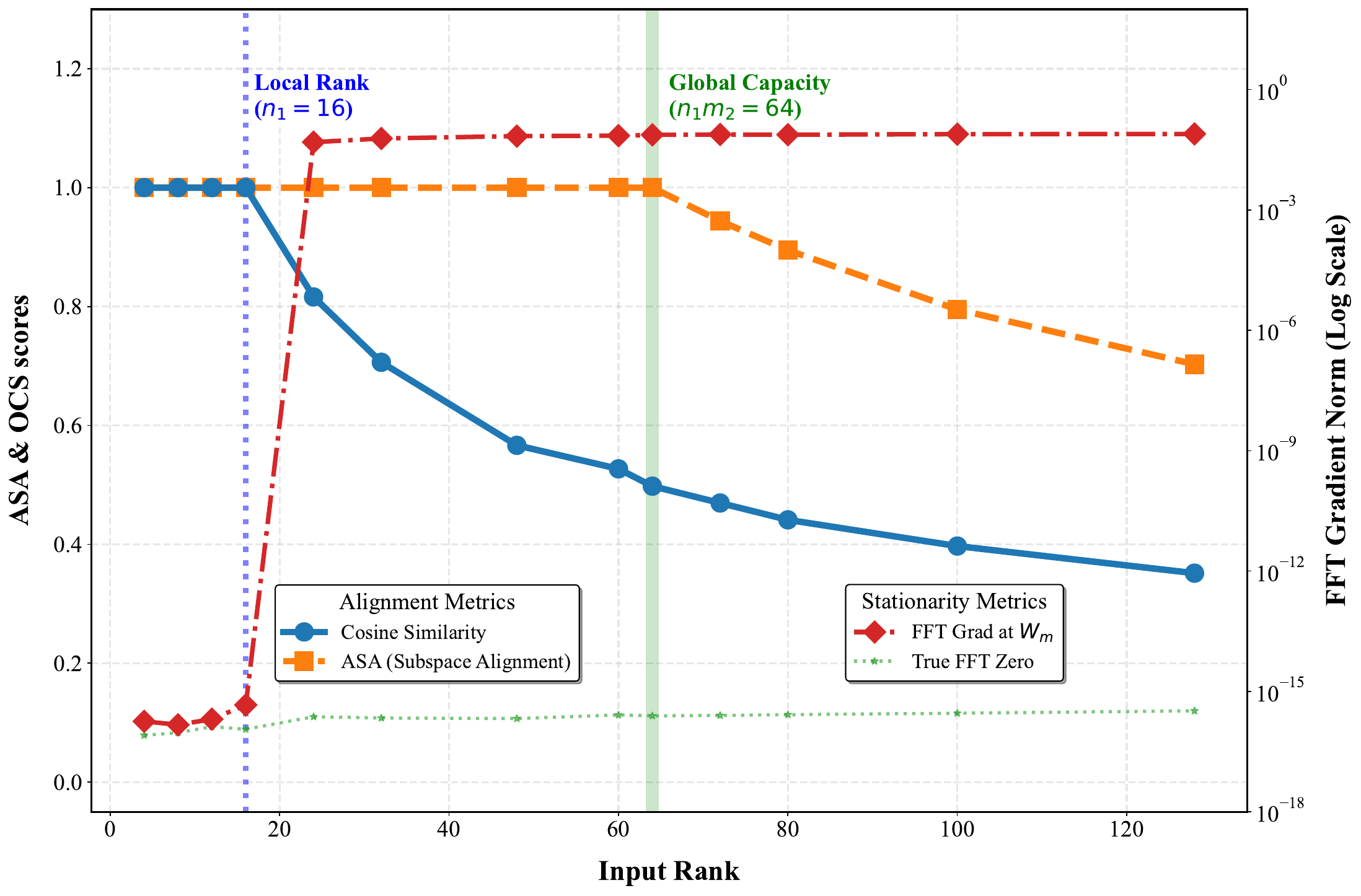}
    \caption{Structural capacity of \repname across input ranks.
    \textbf{Output Cosine Similarity (OCS)} measures point-wise directional alignment between
    \repname and FFT outputs—a precision metric for exact numerical reconstruction.
    \textbf{Averaged Subspace Alignment (ASA)} measures how well the column space of
    $\Delta\rmW_{\ftmrm}$ covers that of $\Delta\rmW_{\text{FFT}}$—a coarser but more
    robust structural metric.
    Two critical thresholds emerge: at $n_1$ (the left-factor rank), OCS collapses as individual
    blocks lose sufficient degrees of freedom for exact reconstruction; at $n_1 m_2$ (the
    \repname adapter rank), ASA begins to gradually decay from $1.0$, showing that \repname
    preserves the fundamental geometric subspace of the target transformation well beyond the
    point-wise precision threshold.}
    \label{fig:app_theo_monarch_static_rank_analysis}
\end{figure}

\Cref{fig:app_theo_morphor_fft_grad_dynamic} confirms the gradient alignment predicted by
\cref{theorem:main_lsq}: when $n_1 \geq \rankop(\rmX)$, \repname gradient norms track those
of FFT throughout training.
\Cref{fig:app_theo_monarch_static_rank_analysis} reveals an important nuance: point-wise
precision (OCS) and subspace alignment (ASA) degrade at \emph{different} rank thresholds.
OCS collapses at $n_1$—the local block capacity—while ASA remains near $1.0$ until the
global adapter rank $n_1 m_2$.
This separation demonstrates that \repname maintains the correct geometric subspace of the
target transformation long after exact numerical reconstruction becomes infeasible,
a property not captured by point-wise metrics alone.

% ==============================================================
\subsection{Gauge Invariance in PEFT: A Unified Perspective and the Parameter-Space Uniqueness of \repname}
\label{sec:app_repara}
% ==============================================================

A central but underappreciated obstacle in factorized fine-tuning is \emph{gauge invariance} which is known as the existence of a continuous family of internal parameterizations that all produce the identical adapter matrix.
It induces flat directions in the loss landscape, destabilizes gradient-based optimizers.
We first expose this phenomenon across representative PEFT families, revealing it as a pervasive structural defect rather than an isolated LoRA artifact.
We then prove that the invariant manifold of \repname parameterization reduces to \emph{diagonal scalings}.

% --------------------------------------------------------------
\subsubsection{Gauge Invariance: A Taxonomy of Failure Modes}
\label{sec:app_repara_audit}
% --------------------------------------------------------------

We unify the analysis under a single framework.
Let $\Phi(\vtheta)$ denote any parameterized adapter, mapping a tuple of trainable tensors $\vtheta$ to an update matrix $\Delta\rmW$.
We say $\Phi$ admits a \emph{gauge group} $\mathcal{G}$ if there exists a non-trivial group action $g \cdot \vtheta \mapsto \vtheta'$ such that $\Phi(\vtheta') = \Phi(\vtheta)$ for all $g \in \mathcal{G}$.
The larger $\mathcal{G}$ is, the more severe the redundancy becomes.

\paragraph{LoRA: $\mathcal{G} = \mathrm{GL}(r)$.}
LoRA~\citep{hu2022lora} writes $\Delta\rmW = \rmB\rmA$, $\rmB \in \sR^{M \times r}$, $\rmA \in \sR^{r \times N}$.
For any invertible $\rmC \in \sR^{r\times r}$:
\begin{equation}
    \Delta\rmW = \rmB\rmA = (\rmB\rmC)(\rmC^{-1}\rmA) = \rmB'\rmA'.
\end{equation}
The gauge group is the full general linear group $\mathrm{GL}(r)$, which is $r^2$-dimensional.
Every point in parameter space lies at the intersection of an $r^2$-dimensional flat valley.
This symmetry is broken only at the boundary of the feasible set (e.g.\ when $\rmA$ or $\rmB$ becomes rank-deficient), which is measure-zero but attracting for gradient descent near convergence, producing the well-documented \emph{rank collapse} phenomenon~\citep{biderman2024loralearnsforgets}.
The same $\mathrm{GL}(r)$ redundancy propagates verbatim to DoRA~\citep{liu2024dora}, PiSSA~\citep{meng2024pissa}, MiLoRA~\citep{wang2024milora}, HiRA~\citep{huang2025hira}, RandLoRA~\citep{albert2024randlora}, and ABBA~\citep{singhal2025abba}, since all of them retain the $\rmB\rmA$ formulation at their core.

\paragraph{QuanTA: Multi-linear $\mathcal{G} = \mathrm{GL}(d_1) \times \cdots \times \mathrm{GL}(d_K)$.}
QuanTA~\citep{chen2024quanta} contracts a chain of parameter tensors $\{\gT_k\}_{k=1}^{K}$ over shared latent dimensions $\{d_k\}$.
The weight update (suppressing input indices) is:
\begin{equation}
    \Delta\mathcal{W}_{g,h,i,a,b,c} = \sum_{d,e,f} (\gT_1)_{g,h,d,e}\,(\gT_2)_{d,i,a,f}\,(\gT_3)_{e,f,b,c}.
    \label{eq:app_repara_quanta_einsum}
\end{equation}
Inserting identity resolutions $\rmC_d^{-1}\rmC_d = \mI$ along each shared latent dimension $d$ independently:
\begin{equation}
    \gT_1 \times_d \gT_2 = (\gT_1 \times_d \rmC_d^{-1}) \times_d (\rmC_d \times_d \gT_2) = \tilde{\gT}_1 \times_d \tilde{\gT}_2, \qquad \rmC_d \in \mathrm{GL}(d).
\end{equation}
Since this freedom is independent across every shared axis, the total gauge group is $\prod_k \mathrm{GL}(d_k)$, whose dimension is $\sum_k d_k^2$.
For typical QuanTA configurations with $K=3$ and $d_k \sim 8$–$64$, this amounts to hundreds of redundant degrees of freedom per adapter, making the loss landscape exponentially more degenerate than LoRA.

\paragraph{Orthogonal fine-tuning (OFT / BOFT): $\mathcal{G} = O(r)$.}
Orthogonal Fine-Tuning~\citep{qiu2023controlling} and its butterfly variant BOFT~\citep{boft} parameterize the adapter as an orthogonal matrix $\rmQ$, updating the weight as $\rmW_{ft} = \rmQ \rmW_0$.
Because $\rmQ$ is constrained to $O(N)$, a direct dense gauge group is not available; however, within the block structure of BOFT, each block admits an $O(b)$ ambiguity ($b$ being the block size) in its parameterization via skew-symmetric generators, yielding a gauge group $\prod_k O(b)$.
For skew-symmetric parameterizations of the form $\rmA = \rmU\rmV^\top - \rmV\rmU^\top$, the substitution $\rmU \leftarrow \rmU\rmC$, $\rmV \leftarrow \rmV\rmC$ for any $\rmC \in O(r)$ leaves $\rmA$ unchanged, confirming an $O(r)$-valued gauge symmetry.

\medskip
\noindent
\textbf{Summary.}
Across the PEFT landscape, gauge invariance is the rule, not the exception.
The severity is determined by the dimensionality of the gauge group: $\dim \mathcal{G} = r^2$ for LoRA; $\sum d_k^2$ for QuanTA; $\binom{r}{2}$ for orthogonal methods.

% --------------------------------------------------------------
\subsubsection{Invariant Manifold of \repname Parameterization}
\label{sec:app_repara_morpho}
% --------------------------------------------------------------

We build the theory in three steps as follows:

\paragraph{Step 1: The Kronecker Baseline}

We first establish uniqueness for the most structured case—a pure Kronecker product—as the theoretical bedrock.

\begin{theorem}[Uniqueness of Kronecker factorization]
    \label{theorem:kronecker_uni}
    Let $\rmK = \rmA_1 \otimes \rmB_1 \neq \mathbf{0}$, with $\rmA_1 \in \sR^{m_1 \times n_1}$, $\rmB_1 \in \sR^{m_2 \times n_2}$.
    If $\rmC \otimes \rmD = \rmK$ for some $\rmC \in \sR^{m_1 \times n_1}$, $\rmD \in \sR^{m_2 \times n_2}$, then:
    \begin{equation}
        \exists\, k \in \sRs : \quad \rmC = k^{-1}\rmA_1, \quad \rmD = k\,\rmB_1.
    \end{equation}
\end{theorem}

\begin{proof}
Since $\rmK \neq \mathbf{0}$, at least one block entry $a_{ij}\rmB_1 \neq \mathbf{0}$ of the first Kronecker expansion equals $c_{ij}\rmD$ of the second.
With $a_{ij}, c_{ij} \neq 0$ (otherwise the block is zero in both), it follows that $\rmD = k\,\rmB_1$ where $k = a_{ij}/c_{ij}$.
Substituting back and applying the cancellation law for Kronecker products:
\begin{equation}
    \rmA_1 \otimes \rmB_1 = \rmC \otimes (k\,\rmB_1) = (k\,\rmC) \otimes \rmB_1 \quad \Rightarrow \quad \rmA_1 = k\,\rmC.
\end{equation}
Hence $\rmC = k^{-1}\rmA_1$, and the scalar $k$ is the sole degree of freedom.
\end{proof}

\Cref{theorem:kronecker_uni} establishes a striking contrast with LoRA: the Kronecker product $\rmA \otimes \rmB$ uniquely determines the pair $(\rmA, \rmB)$ up to a \emph{single scalar}.
The gauge group is $\mathcal{G} = \sRs \cong (\sRs)^1$, reducing $r^2$ degrees of freedom to exactly $1$.

\paragraph{Step 2: Square Monarch Matrices}

Monarch parameterization extends Kronecker products to square matrices via two block-diagonal factors and a permutation.
We characterize its full invariant manifold for a $N\text{-by-}N$ square matrix, factorized by $n: n^2=N$

\begin{lemma}[Permutation conjugation of diagonal matrices]
    \label{lemma:fw_block_requirements}
    Let $\mD_L = \diag(\vv)$ be an invertible diagonal matrix and $\mP_1 = \mP_{n,n}$ the perfect shuffle permutation.
    Define $\rmL = \rmL_0\mD_L$ and $\rmR = \mP_1^\top\mD_L^{-1}\mP_1\rmR_0$.
    Then $(\rmL, \rmR)$ are block-diagonal with the same block structure as $(\rmL_0, \rmR_0)$.
\end{lemma}

\begin{proof}
Right-multiplying a block-diagonal matrix $\rmL_0$ by a diagonal $\mD_L$ scales each row independently, preserving the block structure: $\rmL_{ij} = (\rmL_0)_{ij} (D_L)_{jj}$.
For $\rmR$, observe that $\mQ := \mP_1^\top\mD_L^{-1}\mP_1$ satisfies $\mQ_{ij} = d_{\pi(j)}[\ve_{\pi(i)}^\top\ve_{\pi(j)}]$, which is zero for $i \neq j$ and equals $d_{\pi(j)}$ for $i = j$.
Thus $\mQ$ is diagonal with entries permuted by $\pi$, and left-multiplying a block-diagonal $\rmR_0$ by a diagonal matrix preserves its block structure.
\end{proof}

\begin{lemma}[Reverse direction: block-diagonality forces diagonality]
    \label{lemma:rv_block_requirements}
    Let $\mD_L, \mD_R$ satisfy $\rmL = \rmL_0\mD_L$ and $\rmR = \mD_R^{-1}\rmR_0$ with $(\rmL,\rmR,\rmL_0,\rmR_0) \in \gB\gD^{(n)}$ (block-diagonal with $n \times n$ blocks), and $\mD_R = \mP_1^\top\mD_L\mP_1$.
    Then $\mD_L$ and $\mD_R$ are diagonal.
\end{lemma}

\begin{proof}
Since $(\rmL, \rmL_0)$ and $(\rmR, \rmR_0)$ are pairwise block-diagonal, $\mD_L, \mD_R \in \gB\gD$.
Write the permutation function $\pi$ of $\mP_{n,n}$ explicitly (1-indexed):
\begin{equation}
    \pi: i \mapsto n\bigl((i-1) \bmod n\bigr) + \bigl\lceil i/n \bigr\rceil.
\end{equation}
The $(i,j)$-entry of $\mD_R = \mP_1^\top\mD_L\mP_1$ is $(\mD_L)_{\pi(i),\pi(j)}$.
For a principal entry $(i,j)$ within the $k$-th diagonal block of $\mD_R$ (so $(k-1)n < i,j \leq kn$ or $\lceil i/n\rceil = \lceil j/n\rceil = k$), write $i = (k-1)n + l_i$, $j = (k-1)n + l_j$ with $l_i, l_j \in \overline{1..n}$. The relocated index satisfies $\lvert i - j\rvert_{\mD_R} = \lvert l_i - l_j\rvert n \geq n$ whenever $i \neq j$, placing it strictly outside any diagonal block of $\mD_L$.
Since $\mD_L \in \gB\gD$ requires off-diagonal-block entries to vanish, we conclude $(\mD_R)_{ij} = 0$ for all $i \neq j$, i.e.\ $\mD_R$ is diagonal, and then so is $\mD_L$ by the same argument applied in reverse.
\end{proof}

\begin{theorem}[Invariant manifold of square Monarch / \repname]
    \label{theorem:reparameterization_square_matrix}
    Let $\rmM_0 = \mP_1^\top\rmL_0\mP_1\rmR_0 \in \mset^{(N)}$ be a square Monarch matrix.
    The set of all generative factor pairs is:
    \begin{equation}
        \mset_g(\rmM_0) = \Bigl\{\, (\rmL,\rmR) \;\Big|\; \rmL = \rmL_0 \mD,\; \rmR = \mP_1^\top\mD^{-1}\mP_1\rmR_0;\; \mD = \diag(\vv),\; \vv \in \sRs^N \,\Bigr\}.
    \end{equation}
    The gauge group is isomorphic to $\sRs^N$, the $N$-dimensional multiplicative torus.
\end{theorem}

\begin{proof}
Let $(\rmL, \rmR) \in \mset_g(\rmM_0)$, so $\mP_1^\top\rmL\mP_1\rmR = \mP_1^\top\rmL_0\mP_1\rmR_0$.
Cancelling the leading $\mP_1^\top$ and right-multiplying by $\rmR^{-1}$ (which exists almost surely in the random-initialization regime; see \cref{sec:app_sec_rank_pro}):
\begin{equation}
    \rmL_0^{-1}\rmL\mP_1 = \mP_1\rmR_0\rmR^{-1} \quad \Rightarrow \quad \mD_L\mP_1 = \mP_1\mD_R,
\end{equation}
where $\mD_L = \rmL_0^{-1}\rmL$ and $\mD_R = \rmR_0\rmR^{-1}$, giving the conjugation relation $\mD_R = \mP_1^\top\mD_L\mP_1$.
By \cref{lemma:rv_block_requirements}, $\mD_L$ is diagonal; by \cref{lemma:fw_block_requirements}, $(\rmL, \rmR)$ are valid block-diagonal factors.
The forward direction (any diagonal $\mD$ yields a valid reparameterization) is immediate from \cref{lemma:fw_block_requirements}.
\end{proof}

\paragraph{Step 3: General \repname (Rectangle)}

\begin{lemma}[Pseudoinverse block-preservation for rectangular factors]
    \label{lemma:rv_block_rec}
    Let $\rmL \in \sR^{M \times q}$ be full column-rank block-diagonal and $\rmR \in \sR^{q \times N}$ be full row-rank block-diagonal with $q = n_1m_2 < \min(M,N)$.
    If $\rmL = \rmL_1\mD_{Lr}$ and $\rmR = \mD_{Rr}^{-1}\rmR_1$ with $\mD_{Rr} = \mP_r^\top\mD_{Lr}\mP_r$, then $\mD_{Lr}$ and $\mD_{Rr}$ are diagonal.
\end{lemma}

\begin{proof}
Since block-diagonal matrices with full-rank blocks admit pseudoinverses that decouple blockwise (Moore–Penrose uniqueness~\citep{penrose1955generalizedinverse}), both $\mD_{Lr} = \rmL_1^\dagger\rmL$ and $\mD_{Rr} = \rmR_1\rmR^\dagger$ lie in $\gB\gD$, specifically as $q \times q$ square matrices with $m_2$ blocks of size $n_1 \times n_1$ and $n_1$ blocks of size $m_2 \times m_2$, respectively.
The conjugation $\mD_{Rr} = \mP_r^\top\mD_{Lr}\mP_r$ with permutation $\mP_r = \mP_{n_1,m_2}$ (with
$\pi(i) = \pi_{n_1,m_2}(i) = n_1\left[(i-1) \bmod m_2 \right] + \lceil i/m_2\rceil$
) relocates each off-diagonal principal entry of $\mD_{Lr}$ to a position outside any block of $\mD_{Rr}$.
Since $\mD_{Rr} \in \gB\gD$ forces those entries to zero, $\mD_{Lr}$ must be diagonal, and hence so is $\mD_{Rr}$.
\end{proof}

\begin{theorem}[Gauge invariance of parameter-efficient \repname]
    \label{theorem:reparameterization_low_hidden_rank}
    Let $\rmM_1 = \mP_l^\top\rmL_1\mP_r\rmR_1 \in \mset^{(M,N)}$ with hidden dimension $q = n_1m_2 \leq \min(M,N)$.
    The invariant manifold of generative factor pairs is:
    \begin{equation}
        \mset_g(\rmM_1) = \Bigl\{\, (\rmL,\rmR) \;\Big|\; \rmL = \rmL_1 \mD,\; \rmR = \mP_r^\top\mD^{-1}\mP_r\rmR_1;\; \mD = \diag(\vv),\; \vv \in \sRs^q \,\Bigr\}.
    \end{equation}
    The gauge group is $\mathcal{G}_{\repname} \cong (\sRs)^q$.
\end{theorem}

\begin{proof}
For a randomly initialized adapter, $\rmL_1$ is full column-rank and $\rmR_1$ is full row-rank almost surely.
From $\mP_l^\top\rmL\mP_r\rmR = \mP_l^\top\rmL_1\mP_r\rmR_1$, left-multiply by $\rmL_1^\dagger\mP_l$ and right-multiply by $\rmR^\dagger$:
\begin{equation}
    \mD_{L1}\mP_r = \mP_r\mD_{R1}, \quad \mD_{L1} = \rmL_1^\dagger\rmL,\quad \mD_{R1} = \rmR_1\rmR^\dagger,
\end{equation}
yielding the conjugation $\mD_{R1} = \mP_r^\top\mD_{L1}\mP_r$.
By \cref{lemma:rv_block_rec}, $\mD_{L1}$ is diagonal; by \cref{lemma:fw_block_requirements}, any diagonal $\mD$ generates a valid block-diagonal reparameterization.
\end{proof}

% --------------------------------------------------------------
\subsubsection{Discussion: Why Diagonal Gauge Structure Matters for Optimization}
\label{sec:app_repara_discussion}
% --------------------------------------------------------------

\paragraph{The geometry of degeneracy.}
\Cref{theorem:reparameterization_low_hidden_rank} establishes that the invariant manifold of \repname is a $q$-dimensional torus denoted as $\mathcal{T}^q = (\sRs)^q$.
In contrast, LoRA's gauge group $\mathrm{GL}(r)$ contains not only diagonal scalings but also rotations, shearings, and reflections. All of these are invisible to the loss function yet create curved, degenerate valleys that optimizers may navigate poorly.

\paragraph{Decoupling magnitude from direction.}
The $\mathrm{GL}(r)$ symmetry of LoRA conflates magnitude and directional updates in the parameter space.
DoRA~\citep{liu2024dora} addresses this empirically by explicitly decomposing weight updates into magnitude and direction components.
\repname resolves the same issue structurally: since the only reparameterization freedom is diagonal scaling, the $i$-th hidden feature's magnitude is encoded by $\mD_{ii}$, while its direction is fixed by $\rmL_i$ and $\rmR_i$.
Magnitude and direction are \emph{structurally decoupled} without requiring any post-hoc decomposition.

% ============================
%More
\subsection{Distinction from MoRE Fine-tuning}
\label{sec:app_more_dist}

Although both \ftmt and MoRE~\cite{tan2025more} leverage perfect shuffle permutations in their designs, they differ fundamentally in their algebraic factorization, resulting in distinct interaction topologies.

Consider a MoRE adapter $\rmW \in \sR^{M \times N}$, defined as $\rmW = \mP_2 \rmL_B \cdot \mP_z \cdot \rmR_B$, where $b$ is the number of blocks in both $\rmL_B$ and $\rmR_B$, $r_{blk}$ is the local rank of each block, and $\mP_z, \mP_2$ are the stride permutation matrices as prescribed in the MoRE configuration. The matrices $\rmL_B$ and $\rmR_B$ consist of submatrices of size $m \times r_{blk}$ and $r_{blk} \times n$, respectively, where $m = M/b$ and $n = N/b$. In the context of reparameterization, we omit the final permutation matrix $\rmP_2$ in the analysis and simply write: $\rmW = \rmL_B \cdot \mP_z \cdot \rmR_B$, as it cancels out when comparing two adapters of the same configuration.

\begin{gather*}
    \mP_z = \mP_{r_{blk},b}, \\
    \rmL_B = \diag(\rmL_1, \ldots, \rmL_b), \quad \rmL_i \in \sR^{m \times r_{blk}}, 1 \leq i \leq b, \\
    \rmR_B = \diag(\rmR_1, \ldots, \rmR_b), \quad \rmR_j \in \sR^{r_{blk} \times n}, 1 \leq j \leq b.
\end{gather*}

\subsubsection{Different Objectives in Structured Matrix Design}

\emph{Core Motivation:} MoRE prioritizes sparsity preservation to minimize parameters, whereas \ftmt extends Kronecker factorization to gain flexibility, scalability, and global interactions.

\begin{figure}[htbp]
    \centering
    \includegraphics[width=0.87\linewidth]{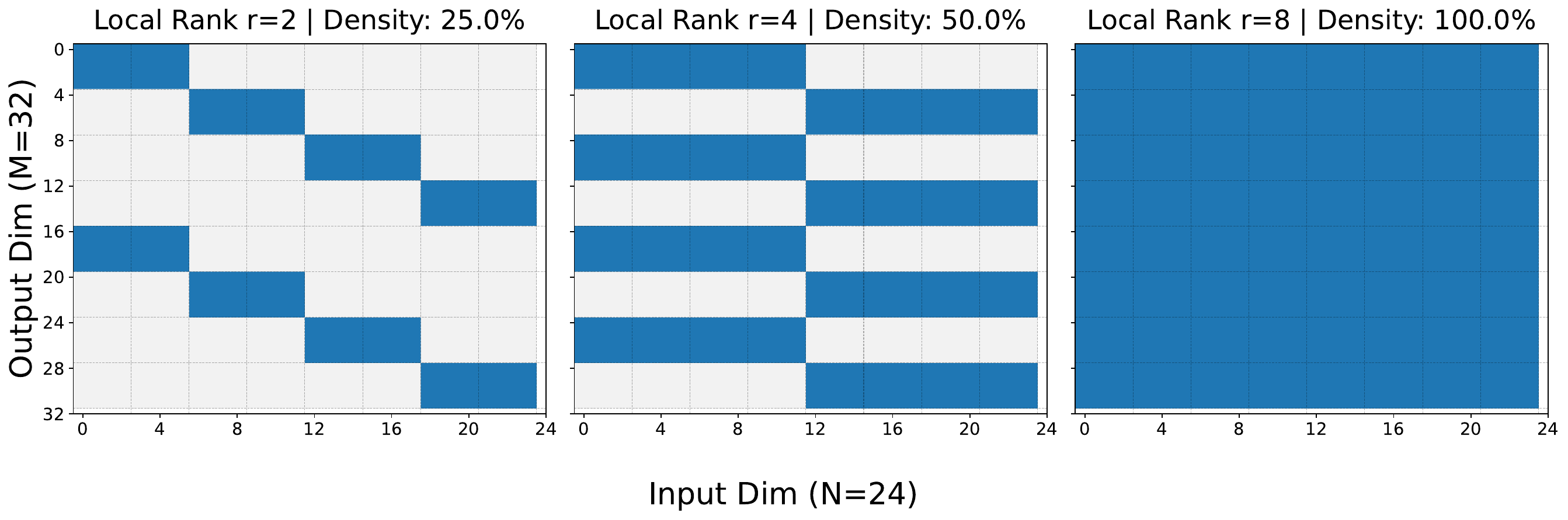}
    \caption{\label{fig:more_connection} Binary masks of MoRE's equivalent output (when $b=8$). Blue regions indicate trainable cross-neuron connections. The visualization suggests that MoRE inherently enforces a fragmented, sparse interaction topology when $r_{blk} < b$. Global dense connectivity is exclusively conditionally triggered when the local rank matches the number of blocks ($r_{blk}=b$).}
\end{figure}

\begin{figure}[htbp]
    \centering
    \includegraphics[width=0.92\linewidth]{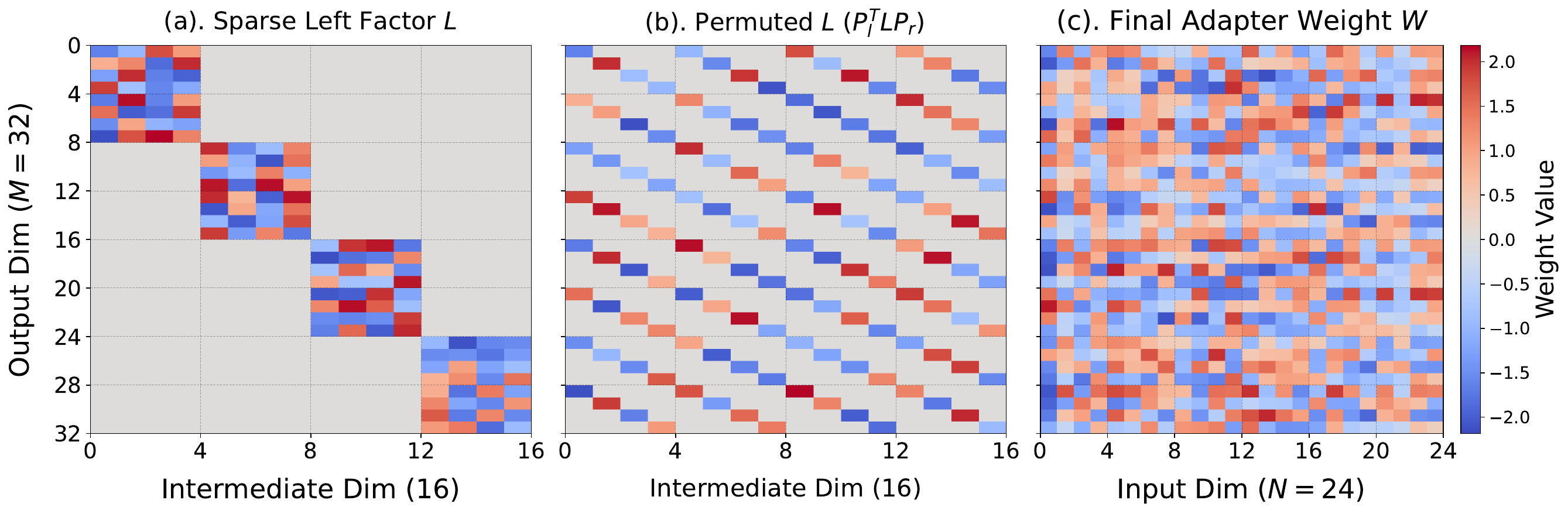}
    \caption{\label{fig:sama_heatmap} The structural transformation pipeline of SaMA. The heatmaps illustrate the deterministic expansion of the parameterization space. The sparse, block-diagonal left factor (a) is reorganized by modulo-based stride permutations into a tiled sub-diagonal pattern (b), effectively executing a Kronecker operand swap. This scattered representation is then integrated by the right factor to produce a fully dense global weight matrix (c), achieving complete input-output feature mixing regardless of adapter configurations.}
\end{figure}

\emph{Architectural Principle:} MoRE follows a partitioned low-rank paradigm: it splits the representation space into isolated, group-wise low-rank blocks and relies on permutations to mix them. In dense adapter regimes, this design suffers from information isolation (\cref{fig:more_connection}). In contrast, \ftmt applies a bilateral structured permutation $(\mP_l^\top \rmL \mP_r)$ that converts the factors into block-diagonal matrices, yielding a globally distributed interaction grid inherited from the Kronecker product (\cref{fig:sama_heatmap}).

\emph{Design Strategy:} Consistent with its partitioning principle, MoRE divides both factors into $b$ low-rank blocks, so its parameter count is dictated solely by the block rank $r_{blk}$. \ftmt instead derives the number of blocks per factor from block shapes -- a choice rooted in the Kronecker view -- which establishes mutual connections across partitions.

\begin{table}[htbp]
    \centering
    \caption{Summary of fundamental structural and algebraic differences between MoRE and SaMA, where $b$ and $r_{blk}$ denote the number of blocks and the block rank, respectively.}
    \label{tab:more_vs_sama}
    \renewcommand{\arraystretch}{1.4} 
    {\small 
        \begin{tabular}{@{} p{3.2cm} p{4.5cm} p{4.5cm} @{}}
            \toprule
            \textbf{Feature} & \textbf{MoRE} & \textbf{SaMA (Ours)} \\ 
            \midrule
            \textbf{Algebraic Form} & Multi-LoRA + Permutation & Kronecker Product Generalization \\ 
            \textbf{Interaction Topology} & Fragmented (Isolated subspaces) & Guaranteed 100\% Dense \\ 
            \textbf{Factor Shapes}
                & \shortstack[l]{$\rmL_B: (b, M/b, r_{blk})$ \\ $\rmR_B: (b, r_{blk}, N/b)$}
                & \shortstack[l]{$\rmL: (m_2, m_1, n_1)$ \\ $\rmR: (n_1, m_2, n_2)$} \\
            \textbf{Parameter Count} $\theta$ & $r_{blk}(M + N)$ & $n_1M + m_2N$ \\ 
            \textbf{Invariant Manifold} & Allows off-diagonal interactions & Diagonal matrices only \\
            \bottomrule
        \end{tabular}
    }
\end{table}

\subsubsection{MoRE fails to restrain the invariant manifold to diagonal transformations}
\begin{theorem}[Reparameterization in MoRE]
    \label{theorem:more_invariance}
    Assume $r_{blk} > b > 1$, there exists a non-diagonal invertible matrix $\mC \in \sR^{br_{blk} \times br_{blk}}$ that reparameterizes the MoRE factors as:
    \begin{equation*}
        \rmL_B^\prime = \rmL_B \,(\mP_z \mC \mP_z^\top), 
        \qquad 
        \rmR_B^\prime = \mC^{-1} \rmR_B,
    \end{equation*}
    while leaving the resulting weight $\rmW$ invariant.
\end{theorem}

\begin{proof}[Constructive Proof]
For any invertible $\mC \in \sR^{br_{blk} \times br_{blk}}$, inserting $\mP_z \mC \mP_z^\top \cdot \mP_z \mC^{-1} \mP_z^\top = \mI$ between $\rmL_B$ and $\mP_z$ gives:
\begin{equation*}
    \rmW \;=\; \rmL_B \,\mP_z\, \rmR_B
    \;=\; \underbrace{\rmL_B (\mP_z \mC \mP_z^\top)}_{\rmL_B^\prime} \,\mP_z\, \underbrace{\mC^{-1} \rmR_B}_{\rmR_B^\prime}.
\end{equation*}
It remains to construct a non-diagonal $\mC$ such that $(\rmL_B^\prime, \rmR_B^\prime)$ are valid MoRE factors, i.e.\ both retain the prescribed block-diagonal structure.

\paragraph{Construction.} 
Take $\mC = \mI + \alpha\, \rmE_{u,v}$, with $\alpha \in \sRs$ and $u \neq v$, where $\rmE_{u,v}$ has a single $1$ at position $(u,v)$. Such $\mC$ is automatically non-diagonal and invertible (with the inverse $\mC^{-1} = \mI - \alpha \rmE_{u,v}$). Validity then reduces to choosing $(u,v)$ so that the block-sparsity patterns of $\rmR_B$ and $\rmL_B$ are preserved:
\begin{enumerate}
    \item \textbf{Right preservation.} $\mC^{-1}$ acts on $\rmR_B$ by row mixing. Confining the mixing within a single block of $\rmR_B$ requires
    $\lceil u / r_{blk} \rceil = \lceil v / r_{blk} \rceil$.
    \item \textbf{Left preservation.} The conjugated factor $\mC_L = \mP_z \mC \mP_z^\top$ acts on $\rmL_B$ by column mixing. Letting $\pi$ denote the index permutation induced by $\mP_z$, hence $\mP_z \ve_u = \ve_{\pi(u)} $, leads to $ \mP_z \rmE_{u,v}\mP_z^\top = \rmE_{\pi(u),\pi{v}}$, therefore the off-diagonal entry must remain inside one block of $\rmL_B$ if:
    $\lceil \pi(u) / r_{blk} \rceil = \lceil \pi(v) / r_{blk} \rceil$.
\end{enumerate}

\paragraph{Existence of valid $(u,v)$.} Write an index in $\rmR_B$ given by $u = (k-1) \cdot b + l$ with $k \in \overline{1..r_{blk}}$ and $l \in \overline{1..b}$. The stride permutation map $u \mapsto \pi(u) = (l-1) \cdot r_{blk} + k$, i.e.\ $u$ is sent to the $k$-th position of block $l$ of $\rmL_B$.

By selecting adjacent coefficients $k_u=1, k_v=2$ and offset $(l=1)$, the linear indices $u=1$ and $v=b+1$ both belong to block 1 of $\rmR_B$ since $r_{blk} > b$. Their images under $\pi$ are $\pi(u) = 1$ and $\pi(v)=2$, both mapped indices remain in block 1 of $\rmL_B$. Thus $(u,v) = (1, b+1)$ satisfies both preservation conditions.

\paragraph{Verification.}
By construction, $\mC$ is non-diagonal and invertible. Condition 1 confines the mixing of $\mC^{-1}$ to a single block of $\rmR_B$, so $\rmR_B^\prime \in \gB\gD$. Condition 2 places the off-diagonal entry of $\mC_L$ inside one block of $\rmL_B$, so $\mC_L \in \gB\gD$ and consequently $\rmL_B^\prime = \rmL_B \mC_L \in \gB\gD$. Hence $(\rmL_B^\prime, \rmR_B^\prime)$ is a valid MoRE reparameterization of $\rmW$. \qedhere
\end{proof}

The construction generalizes immediately to $\mC = \mI + \sum_{(u,v) \in \gS} \alpha_{u,v} \rmE_{u,v}$ for any set $\gS$ of pairs satisfying both conditions, provided $\mC$ is invertible (e.g.\ $\gS = \{(1,b+1),(b+1,1)\}$ with $\alpha_{1,b+1}\alpha_{b+1,1} = 1$ yields $\det \mC = 0$). Since our purpose is only to show that the invariant manifold of MoRE strictly extends beyond the diagonal subspace $\gD$, we leave a complete characterization of this extended subspace to future work.

As a direct corollary of \cref{theorem:more_invariance}, the MoRE design admits mutual interactions across its invariant manifold, at least when $r_{blk}>b$, which can introduce entanglement in the parameter space -- a behavior fundamentally distinct from the diagonal-only gauge structure of \ftmt (\cref{theorem:reparameterization_low_hidden_rank}).

\subsubsection{Pseudo code for \ftmt implementation vs. MoRE implementation}
As the permutations $\mP_l$ and $\rmP_r$ are not explicitly constructed but we use matrix tranpose operation insead, we provide a pseudocode to demonstrate their implemetation in \cref{alg:sama}. We further provide pseudocode for MoRE~\citep{tan2025more} in \cref{alg:more} to help better distinguish the two methods.

\begin{figure}[htbp]
    \centering
    \begin{minipage}{0.95\linewidth} 
    
    % --- THUẬT TOÁN 1: MoRE ---
    \begin{algorithm}[H]
        \caption{MoRE Forward Pass }
        \label{alg:more}
        \begin{algorithmic}[1]
            \Require Input $x \in \mathbb{R}^{B \times (b \times s)}$
            \Require Factors $W_1 \in \mathbb{R}^{b \times r \times s}$, $W_2 \in \mathbb{R}^{b \times s \times r}$

            \Statex % --- Dòng trống giãn cách ---
            \State \textbf{Phase 1: Local Block Projection}
            \State \Comment{Reshape $x$ to $(B, b, s)$ and apply Batched MatMul}
            \State $y_1 \gets \textsc{BlockDiagMult}(x, W_1)$ \Comment{Shape: $(B, b, r)$}

            \Statex % --- Dòng trống giãn cách ---
            \State \textbf{Phase 2: Middle Permutation (Stride Routing)}
            \State \Comment{Flatten, view as $(r, b)$, then transpose}
            \State $y_{p} \gets \textsc{Rearrange}(y_1, \text{'(}r~b\text{)} \rightarrow \text{(}b~r\text{)'})$ \Comment{Shape: $(B, b, r)$}

            \Statex % --- Dòng trống giãn cách ---
            \State \textbf{Phase 3: Final Block Projection}
            \State $y_2 \gets \textsc{BlockDiagMult}(y_p, W_2)$ \Comment{Shape: $(B, b, s)$}

            \Statex % --- Dòng trống giãn cách ---
            \State \textbf{Phase 4: Output Permutation}
            \State $y_o \gets \textsc{Rearrange}(y_2, \text{'(}b~s\text{)} \rightarrow \text{(}s~b\text{)'})$ \Comment{Shape: $(B, s, b)$}

            \Statex % --- Dòng trống giãn cách ---
            \State \Return $\textsc{Reshape}(y_o, [B, s \times b])$
        \end{algorithmic}
    \end{algorithm}

    \vspace{0.3cm} % Tăng khoảng cách giữa 2 thuật toán lên một chút
    
    % --- THUẬT TOÁN 2: SaMA ---
    \begin{algorithm}[H]
        \caption{SaMA Forward Pass (Ours)}
        \label{alg:sama}
        \begin{algorithmic}[1]
            \Require Input $x \in \mathbb{R}^{B \times (c_L \times c_R)}$
            \Require Factors $L \in \mathbb{R}^{r_R \times (r_L \times c_L)}$, $R \in \mathbb{R}^{c_L \times (r_R \times c_R)}$ (3D-tensors) 
            \Require Scaling factor $\alpha$
            
            \Statex % --- Dòng trống giãn cách ---
            \State \textbf{Phase 1: Right Factor Projection}
            \State $y_r \gets \textsc{BlockDiagMult}(x, R)$ \Comment{Shape: $(B, c_L, r_R)$}
            
            \Statex % --- Dòng trống giãn cách ---
            \State \textbf{Phase 2: Bilateral Permutation I (Mid-routing)}
            \State $y_{p1} \gets \textsc{Rearrange}(y_r, \text{'(}c_L~r_R\text{)} \rightarrow \text{(}r_R~c_L\text{)'})$ \Comment{Shape: $(B, r_R, c_L)$}
            
            \Statex % --- Dòng trống giãn cách ---
            \State \textbf{Phase 3: Left Factor Projection}
            \State $y_l \gets \textsc{BlockDiagMult}(y_{p1}, L)$ \Comment{Shape: $(B, r_R, r_L)$}
            
            \Statex % --- Dòng trống giãn cách ---
            \State \textbf{Phase 4: Bilateral Permutation II (Missing in MoRE)}
            \State \Comment{\textbf{Restores spatial alignment via Kronecker commutation}}
            \State $y_{p2} \gets \textsc{Rearrange}(y_l, \text{'(}r_R~r_L\text{)} \rightarrow \text{(}r_L~r_R\text{)'})$ \Comment{Shape: $(B, r_L, r_R)$}
            
            \Statex % --- Dòng trống giãn cách ---
            \State \Return $\textsc{Reshape}(y_{p2}, [B, r_L \times r_R]) \times \alpha$
        \end{algorithmic}
    \end{algorithm}

    \end{minipage}

    \vspace{0.2cm}
    \caption{Pseudo-code comparison detailing the tensor shape transformations. By explicitly tracking the dimensions, it becomes algebraically evident that SaMA incorporates a mandatory second bilateral permutation (Phase 4) to ensure proper output spatial alignment, a step entirely omitted in the MoRE architecture.}
    \label{fig:pseudocode_compare}
\end{figure}

%just a try
\newpage
\section{What evidence supports disentangling the effects of \ftmt adapter features}
\label{app:sec_meta_evidence}

\subsection{Expressivity via rank capacity}
\label{app:sec_meta_rank}
Here we discuss the role of rank capacity when considering 3 criteria:
\begin{itemize}
    \item Low-rank regimes (when this trait is not dominant), 
    \item \oftmto's performance over ranks,
    \item \oftmto's ability to overcome the bottleneck of high-rank regimes. 
\end{itemize}

\begin{figure}[htbp]
    \centering
    \includegraphics[width=0.85\linewidth]{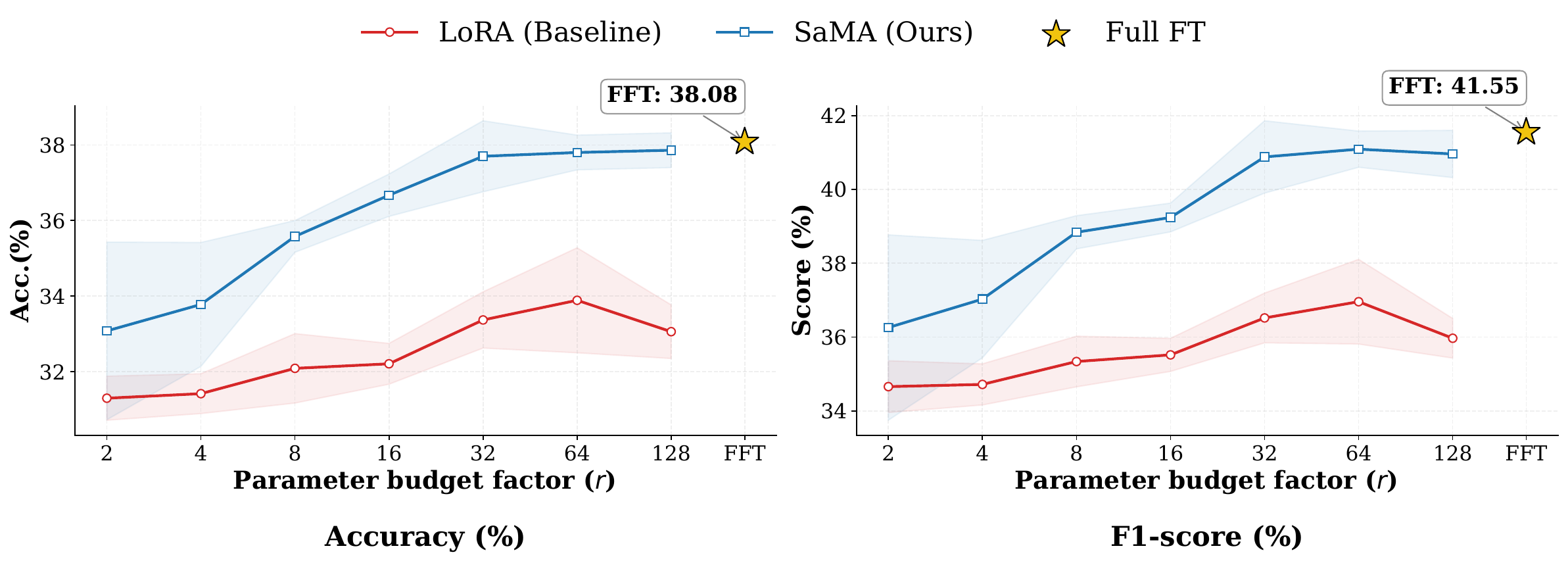}
    \vspace{3pt}
    \caption{\label{fig:app_sama_vs_lora_drop} Performance comparison between \ftmt and LoRA; FFT (on Query and Value modules) serves as the upper-bound baseline. The plots show Accuracy (\%) and F1-score (\%) across various parameter budget factor $r$. Shaded regions represent standard deviations. }
\end{figure}

To isolate the rank effect of \ftmt, we examine the model’s performance on low-rank regions and take the regimes into account.
In \cref{fig:app_sama_vs_lora_drop} (page~\pageref{fig:app_sama_vs_lora_drop}), we demonstrate LoRA and \oftmto's performance over varying parameter counts (associated with adapter ranks).  It is clear that \ftmt gains consistently increase as ranks increase.
This basically suggests that their rank capacity contributes to the overall performance of adapters.

Meanwhile, the overall performance patterns of \ftmt's is higher than LoRA’s, especially in equal-rank comparison of low-rank regimes (for example $\text{LoRA}_{r = 4}$ versus \oftmto$_{2,2}$). It means that even in low-rank regimes, the recommended regions for LoRA, \ftmt still performs well. 
This further suggests positive effects of \ftmt which are beyond rank capacity. Here we tentatively postulate them as inductive biases of the \repname structure as well as gauge symmetry.

Having said that, although rank capacity theoretically means a richer representation space of the adapter, the actual performance might not come with a straightforward translation into gains.
The issue behind high-rank performance in dense matrices like LoRA is also illustrated in \cref{fig:app_sama_vs_lora_drop} (page~\pageref{fig:app_sama_vs_lora_drop}) along with the previous studies from~\citep{he2025smt}.
In these experiments, in high-rank regimes, LoRA either introduces a plateau region while leaving a significant gap to FFT or encounters a rank-collapse mode, making it suffer from poor behavior for complicated tasks.

Apart from that, given \cref{fig:ana_spectrum_fft_lora} (page~\pageref{fig:ana_spectrum_fft_lora}), we can further analyze this phenomenon.
Regarding effective ranks and singular value spectrum patterns, it is usually expected that low-rank assumption underlying LoRA causes the adapter’s singular values to concentrate in the most prominent directions.
However, low-rank patterns in \cref{fig:ana_spectrum_fft_lora}b-d, whose spectrum depletes quickly in high budgets, are not in alignment with FFT (as the reference).

In contrast, \ftmt adapters tend to emulate the long-tailed spectrum patterns of FFT, while maintaining richer effective feature spaces.
This consolidates the contribution of other mechanisms that support training dynamics rather than only static expressivity. Moreover, \oftmto's performance in this case closely reaches the FFT adapter gains.

From these perspectives, it is reasonable to emphasize the contributions of expressivity as well as other remaining factors both directly and indirectly to the overall performance.

\subsection{Morpho structure}
\label{app:sec_meta_structure}
Considering a naive diagonal matrix as a canonical example, this adapter (scaling transform only) theoretically possesses full rank and reduced parameter degeneracy space, but yields a hollow capacity as well as poor scalability. 

To further make a comprehensive analysis, we compare \ftmt with DiaBlo~\citep{gurses2025diablodiagonalblocks} (a standard single block diagonal matrix).
Fundamentally, the distinction lies on the topology of information routing.
An inherent limitation of block-sparse matrices is isolation of far features, which does not facilitate global interaction in the high-dimensional representation spaces of LLMs (but small local groups instead). \Cref{tab:exp_cms_full} (page~\pageref{tab:exp_cms_full}) shows this effect on overall performance (Llama-13B and Llama2-7B).

Moreover, it is still a theoretical question of the reason for choosing the diagonal form for the block, rather than arbitrary arrangements (except for implementation convenience).
They might introduce other paradigms of group-interacting. 
The Morpho structure specifically overcomes this blind spot by permutation, allowing global mixing. These empirical results advocate the contributions of the Morpho structure to the overall effectiveness.

Although general high-rank matrices do not absolutely mean improvement in expressivity, we further notice that: apart from intrinsic high-rank in diagonal matrices or DiaBlo adapter, which may lead in hollow high-ranks in extreme parameter budgets; \ftmt takes this quantity into account by design, making its rank grow as a quadratic function of the budgets.
Thus we endorse this property due to its flexibility and logicality. 

\begin{table}[htbp]
    \centering
    \caption{Performance comparison of \ftmt and DiaBlo on MetaMath tasks.}
    \label{tab:meta_effects_b}
    \vspace{4pt}
    \begin{tabular}{lccc}
        \toprule
        \textbf{Method} & \textbf{\#Params (M)} & \textbf{GSM8K} & \textbf{MATH} \\
        \midrule
        $\text{DiaBlo}_{N=128}$ & 46.4 & 72.18 & 28.92 \\
        $\text{DiaBlo}_{N=64}$ & 92.9 & 72.33 & 29.54 \\
        \oftmto$_{16,16}$ & 54.0 & 78.45 & 39.45 \\
        \bottomrule
    \end{tabular}
\end{table}

To support this perspective, we further design experiments and compare performance of \ftmt and DiaBlo in the MetaMath benchmark (a subset of 12K samples) as we consider the two constituent tasks as demanding NLP reasoning benchmarks that require rich representation capacity.
All schemes were conducted on the same experimental platform.
The results are reported in \cref{tab:meta_effects_b}. Although these are not really comprehensive ablations, the prominent pattern here is that the overall performance degrades in sparse adapters regardless of parameter counts.
Moreover, the gap is particularly considerable in the demanding tasks of MATH.

\subsection{Invariant manifolds}
\label{app:sec_meta_gauge}

Regarding the reduced gauge freedom, it serves as theoretically supportive evidence of (Adam/AdamW) optimizer-friendly adapter.
In our toy experiment, \cref{fig:momentum_aligment} (\cref{sec:express_repara}, page~\pageref{fig:momentum_aligment}) shows empirical support for the claim that the $GL(r)$ gauge freedom of LoRA creates an optimization bottleneck that Morpho’s diagonal gauge structure inherently avoids.
Furthermore, in \cref{fig:app_sama_vs_lora_drop}, we conjecture that the smooth performance improvement towards FFT points might have contributions from the decoupling effect by reduced gauge freedom, which becomes severe when the budgets increase in LoRA.
Without this, high-rank LoRA adapters are shown to not be able to manage the bottleneck as well as closing the gap with FFT despite the fact that LoRA has dense components, allowing more freedom in internal interactions.

\begin{table}[htbp]
    \centering
    \caption{Performance comparison of gauge-inflated and vanilla \ftmt.}
    \vspace{4pt}
    \label{tab:meta_effects_c}
    \begin{tabular}{lccc}
        \toprule
        \textbf{Method} & \textbf{\#Params (M)} & \textbf{GSM8K} & \textbf{MATH} \\
        \midrule
        Gauge-Inflated \ftmt & 56.4 & 77.24 & 38.10 \\
        \oftmto$_{16,16}$ & 54.0 & 78.45 & 39.45 \\
        \bottomrule
    \end{tabular}
\end{table}

To make things clear, we conduct additional comparison experiments where we artificially inflate gauge symmetry into the \ftmt structure. 
In particular, rather than $\Delta \rmW = \mP_l \rmL \mP_r \rmR$, we introduce $(\rmL \leftarrow \rmL @ \rmL^{\prime}, \rmR \leftarrow \rmR^{\prime} @ \rmR)$, where $(\rmL^{\prime}, \rmR^{\prime})$ are square block diagonal matrices of conformable shapes, which add degrees of freedom within each block components.
The rank capacity and the \ftmt structure are relatively preserved. The results are reported in \cref{tab:meta_effects_c}.

This setting (\ftmt$_{16,16}$) introduces a little overhead of 4.4\% on parameter counts compared to the vanilla \ftmt.
The invariant space for each factor now is a merely small batch of 16-by-16 matrices.
All schemes are done on the same experimental platform. As we can see in this experiment results, the performance slightly degrades regardless of a marginal increase in terms of budgets.

\newpage 
\section{Further Experimental Result and Analysis}
\label{sec:app_further_exps}
\subsection{Matrix approximation -- experimental settings}
\label{sec:app_detail_express}
This section describes experiments in \cref{sec:sim_exp_exp}.
\paragraph{Baseline and setup} In order to make a comprehensive understanding and fair evaluation, \ftmt is compared to other reparameterization adapters including: FFT, LoRA family~\cite{hu2022lora,liu2024dora}, QuanTA~\cite{chen2024quanta} with high-rank update. In low-rank target matrix. We then simplify the target to a square matrix and set all its nonzero singular values to $1$ to guarantee that it is low-rank but strictly of $R$, isolating the effect of spectral decay. In other words, $R$ is not only algebraic rank, but also the exact target effective rank~\citep{roy2007effective}.
\begin{figure}[htbp]
	\centering
	\includegraphics[width=0.99\linewidth]{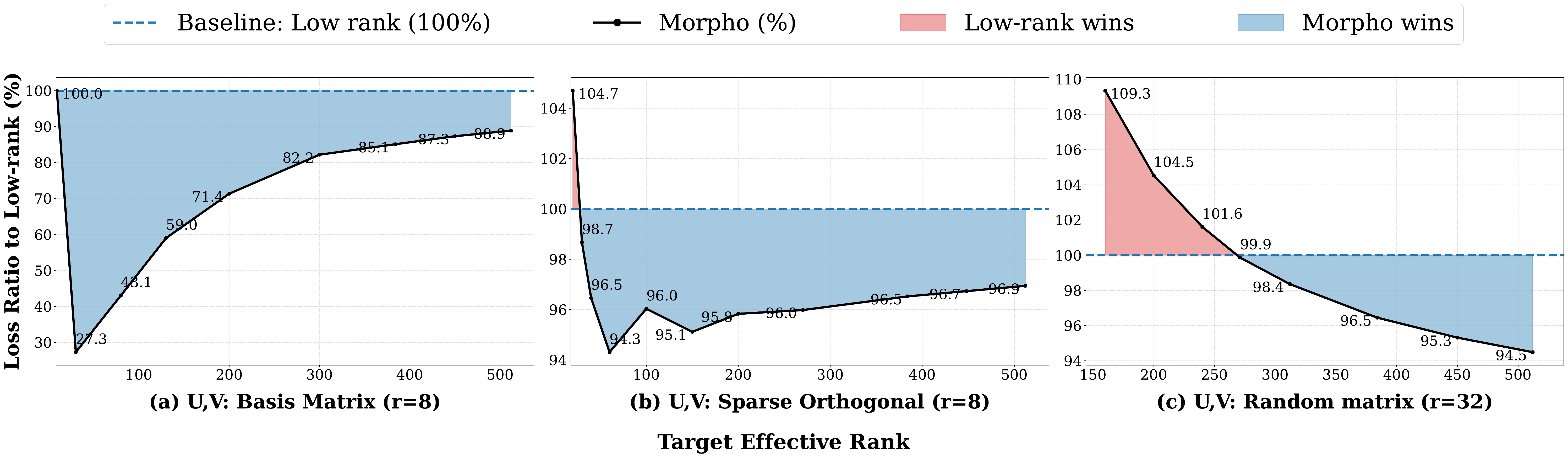}
	\caption{\label{fig:combined_experiments_uv_compress} Relative loss comparison: \repname vs Low-rank. (a) $U,V$ are sampled from the orthogonal bases; (b) $U,V$ are sampled from sparse orthogonal blocks; (c) $U,V$ are sampled from random matrices. The crossover points $R_{th}$ vary on settings. It is obvious that random matrix-based $(U,V)$ is the worst case among all, requires $R_{th} \approx 270$ in Fig. (c) ${r=32}$; or $r=217$ in \cref{fig:main_express_2bl}${r=8}$.}
\end{figure}

In the experiments of matrix approximation, we use the AdamW optimizer\citep{ilya2019adamw} with the same learning rates of $\num{1e-2}$, except for DoRA with high learning rate of $\num{4e-2}$ due to its slow convergence speed in such simulation, running 300 training iterations, then evaluate the average over 10 independent simulations with different random seeds. The main configurations are low-rank $r=8$, and \repname $(n_1,m_2)=(r,r)$, resulting in comparable parameter budgets in the compression problems. Moreover, we also compare low-rank final optimal points to SVD solutions by the Eckart–Young–Mirsky theorem~\citep{Eckart1936TheAO} to make sure that the training processes lead to all points close to the theoretical approximation. The target matrices have dimensions of $1024 \times 1024$.
\paragraph{Entropy effective rank} The standard algebraic rank of a matrix is highly sensitive to noise as infinitesimally small perturbations can easily lead to full apparent rank. In addition, it does not take into account and reflect the full singular value spectrum. To overcome these limitations, Roy and Vetterli~\citep{roy2007effective} introduced an alternative continuous measurement of underlying information -- effective rank (erank). Given a nonzero matrix $\rmA$ of shape $(M \times N), K = \min{M,N}$, with all singular values $\{\sigma_i(\rmA)\}$, the erank is calculated as follows:
\begin{align}
    p_i &= \frac{\sigma_i}{\sum_{j=1}^K \sigma_j}, \quad \text{for } i = 1, 2, \dots, K. \notag \\
    H(p) &= -\sum_{i=1}^K p_i \ln(p_i) \qquad  \text{-- \quad Shannon Entropy,} \notag \\
    \text{erank}(\rmA) &= r_{\text{eff}}(\rmA) = \exp[H(p)] = \exp \left( -\sum_{i=1}^K p_i \ln(p_i) \right).
\end{align}
The effective entropy rank provides a continuous capacity measure bounded by $1 \leq r_{\text{eff}} \leq K$. It reaches its minimum $r_{\text{eff}}=1$ when the matrix is strictly rank-1, and its maximum $r_{\text{eff}}=K$ when all singular values are equal (an isotropic spectrum). In the context of machine learning, erank quantifies the true intrinsic capacity and information bottleneck of weight matrices, filtering out negligible noise directions that do not meaningfully contribute to the representation power. In our matrix approximation setting, the target matrix is already constructed to be algebraic low-rank, given that reason, we tighten the design by imposing low erank as a complementary metric without additional spectral collapse as well as rank deficiency.

\begin{figure}[htbp]
	\centering
	\includegraphics[width=0.99\linewidth]{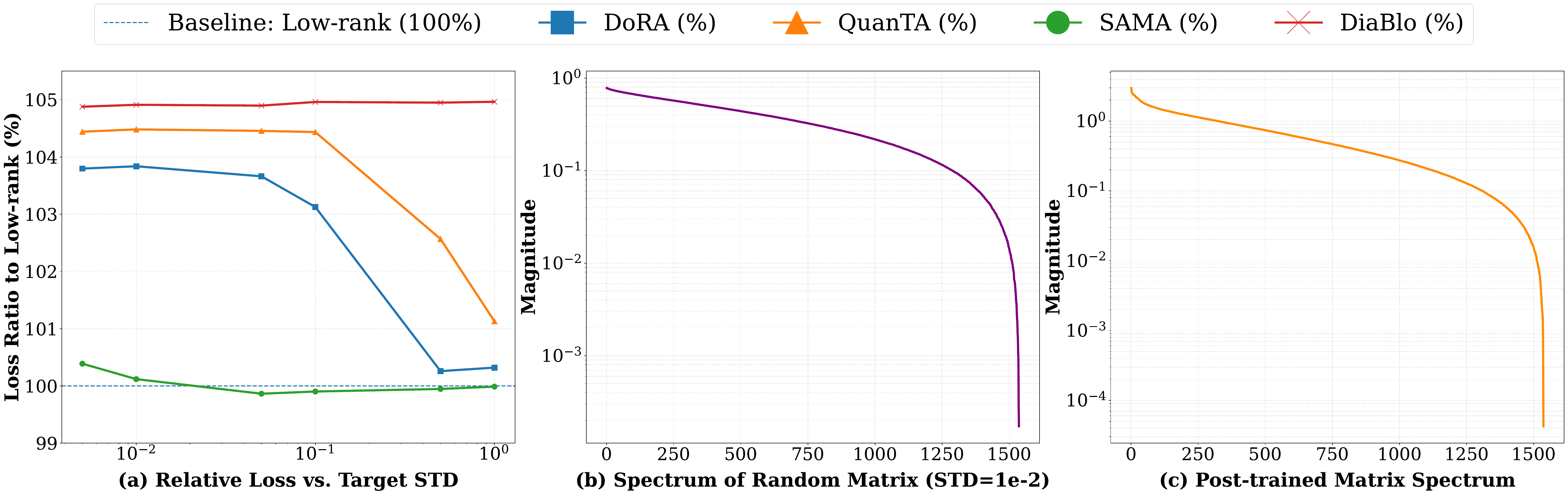}
	\caption{\label{fig:express_sweep_and_spectra} Illustration of random target mode. \repname performs on par or slightly better than Low-rank approximation, outperforms other baselines. (a) Comparison on expressivity of several adapters, given a random target. (b) Singular value spectrum of a matrix with random entries. (c) Singular value spectrum of a fine-tuned model weight. To make alignment with reference matrix, and balance parameter budgets, we set dimension $d=1536$, budgets equivalent to rank $r=32$.}
\end{figure}

\paragraph{Target patterns and additional comparison results} We also apply several settings of left and right orthogonal singular matrix $(U,V)$ in order to control the geometric structure of the target updates.
In the main experiment in \cref{theo:express_approx} and \cref{fig:combined_experiments_uv_compress}~(c), $(U,V)$ are independently sampled from random dense matrices, representing global and the least structured geometry.
In addition, we create two special cases, where $(U,V)$ are sampled from orthogonal bases, representing an extremely simple and highly structured case, shown in \cref{fig:combined_experiments_uv_compress}~(a) and block matrices of sparse orthogonal components, introducing local structure of overall orthogonality, shown in \cref{fig:combined_experiments_uv_compress}~(b). These three settings provide a spectrum from simple local geometry to fully isotropic geometry, allowing us to examine how the crossover point $R_{th}$ varies with the structural complexity of the target matrix.

Moreover, we observe that random matrices share some basic characteristics with LLMs weights, being an important research interest in Deep learning. Hence, we adopt them as an exemplar target in our expressivity tests, then report the output in \cref{fig:express_sweep_and_spectra}. It is obvious that given the comparable number of parameters, Low rank approximation and \repname parameterization considerably attain better compression than the other baselines. Moreover, \repname parameterization in some cases slightly improves upon Low-rank which is supported by the Eckart-Young theorem~\citep{Eckart1936TheAO}.

Next we will move to additional empirical results on downstream tasks.
\subsection{Full results on the DROP dataset}
\label{sec:app_further_drop}
In this section, we provide the full model-by-model comparison underlying the main-paper summary, with several experiments on large models. \Cref{tab:llama_drop_bench} provides detailed outputs, along with \cref{fig:drop_chart} in \cref{sec:exp_drop}. Across all of our experiments, we train models for 2 epochs, except for the QuanTA experiments on Llama2-7B, which use 3 epochs to match the setting in the original paper~\cite{chen2024quanta}.
\tableDropMainResult

Except for an outlier of $\ftmt_{2,2}$, \ftmt performance also steadily improves when the parameter budget increases and exhibits a nearly monotonic trend that suggests efficient parameter utilization. In these settings, to make a completely fair comparison between baselines, we only conduct full fine-tuning experiments on the corresponding modules, namely Query (\texttt{q\_proj}) and Value (\texttt{v\_proj}). As shown in the table, the gain with \ftmt is clearly visible across low-to-high budget regimes of PEFT, where \ftmt narrows or closes the gap to FFT more reliably than the other baselines.

\subsection{Full results on the math reasoning task}
\tableArithMainResultFull

Table~\ref{tab:arith_comparison_full_results} reports the complete arithmetic-reasoning results, extending the main-paper summary with full model-by-model and method-by-method comparisons. These detailed results complement the main discussion by showing the full ranking of baselines under aligned evaluation settings, and by making the relative strengths of lightweight and higher-capacity adapters more transparent.

Overall, the arithmetic-reasoning benchmark shows that \ftmt maintains a favorable balance between accuracy and parameter efficiency across different settings. In particular, the detailed table confirms that the advantage of \ftmt is not limited to a single configuration, but remains visible across multiple model scales and adapter budgets. This pattern is especially meaningful for arithmetic reasoning, where stronger performance typically requires both sufficient expressivity and stable optimization; the results therefore support our claim that \ftmt provides a practical and robust alternative to conventional low-rank adapters in numerically demanding tasks.

\subsection{A comprehensive benchmark of Commonsense reasoning}

\begin{table*}[htbp]
  \centering
  \caption{Comprehensive benchmark of all FT methods on Commonsense reasoning, extending the main-paper summary with full comparisons.
  The best and runner-up results are highlighted in \textbf{bold}, and \underline{underline}, respectively. Higher is better for all tasks. Symbols indicate where the results are taken from -- $\ast$:~\citep{hu2023llmadapters}, $\dagger$: MiLoRA~\citep{wang2024milora}, $\perp$: KaSA~\citep{wang2024kasaknowledgeawaresingularvalue}, $\circ$: DoRA~\citep{liu2024dora}, $\bullet$: SMT~\citep{he2025smt}, $\spadesuit$: S$^2$FT~\citep{yang2024s2ftefficientscalable}, $\square$: HiRA~\citep{huang2025hira}, $\blacktriangledown$: FlexLoRA~\citep{liu2025flexloraentropyguidedflexible}  $\vee$: RepLoRA~\citep{reploara25truong}, $\bigstar$: RefLoRA~\citep{zhang2025reflorarefactoredlowrank}, $\curlywedge$: RandLoRA~\citep{albert2024randlora}, $\circledast$: LaX~\citep{zhang2025laxboost}, $^\triangle$: DiaBlo~\citep{gurses2025diablodiagonalblocks}, respectively. $\blacktriangle$: projected trainable parameters}.
  \label{tab:app:cms_full_results}
  \resizebox{0.98\textwidth}{!}{%
    \begin{tabular}{llcccccccccc}
      \toprule
      \textbf{Model} & \textbf{FT Method} & \textbf{\#Params\%} & \textbf{BoolQ} & \textbf{PIQA} & \textbf{SIQA} & \textbf{HellaS.} & \textbf{WinoG.} & \textbf{ARC-e} & \textbf{ARC-c} & \textbf{OBQA} & \textbf{AVG}($\uparrow$) \\
      \midrule
      \rCrChatGPT \midrule
      \multirow{16}{*}{LLaMA-7B} \rCrSevenFFT \myDashLine \rCrSevenPrefix \rCrSevenSeries \rCrSevenParallel \rCrSevenLoRA \rCrSevenRefLoRAl \rCrSevenRepLoRA \rCrSevenDoRAh \rCrSevenDoRAf \rCrSevenGalore \rCrSevenLaX \rCrSevenLISA \rCrSevenSSFT
      \rCrSevenSMT \rCrSevenSMTBest \myDashLine \rCrSevenSaMAFour \rCrSevenSaMASix  \midrule
      \multirow{16}{*}{LLaMA-13B} \rCrThirteenFFT \myDashLine \rCrThirteenPrefix \rCrThirteenSeries \rCrThirteenParallel \rCrThirteenLoRA \rCrThirteenRepLoRA \rCrThirteenDoRAh \rCrThirteenDoRAf \rCrThirteenLaX \rCrThirteenSSFT \rCrThirteenSMT \rCrThirteenSMTBest \rCrThirteenDiaBlo \myDashLine \rCrThirteenSaMAFour \rCrThirteenSaMASix  \midrule
      \multirow{16}{*}{LLaMA2-7B} \rCrTwoFFT \myDashLine \rCrTwoLoRA \rCrTwoRefLoRAl \rCrTwoPiSSA \rCrTwoMiLoRA  \rCrTwoDoRAh \rCrTwoDoRAf \rCrTwoHiRA \rCrTwoKaSA \rCrTwoSSFT \rCrTwoSMT \rCrTwoSMTBest \rCrTwoDiaBlo \myDashLine \rCrTwoSaMAFour \rCrTwoSaMASix \midrule
      \multirow{16}{*}{LLaMA3-8B} \rCrThreeFFT \myDashLine \rCrThreeLoRA \rCrThreeRefLoRAl \rCrThreePiSSA \rCrThreeMiLoRA \rCrThreeDoRAh \rCrThreeDoRAf \rCrThreeHiRA \rCrThreeFlexLoRA \rCrThreeKaSA \rCrThreeSSFT \rCrThreeRandLoRA \rCrThreeRandLoRABest \rCrThreeSMT \rCrThreeSMTBest \myDashLine \rCrThreeSaMAFour \rCrThreeSaMASix
      \bottomrule
    \end{tabular}
  }
  \label{tab:exp_cms_full}
\end{table*}

\begin{figure}[htbp]
    \centering
    \includegraphics[width=0.98\linewidth]{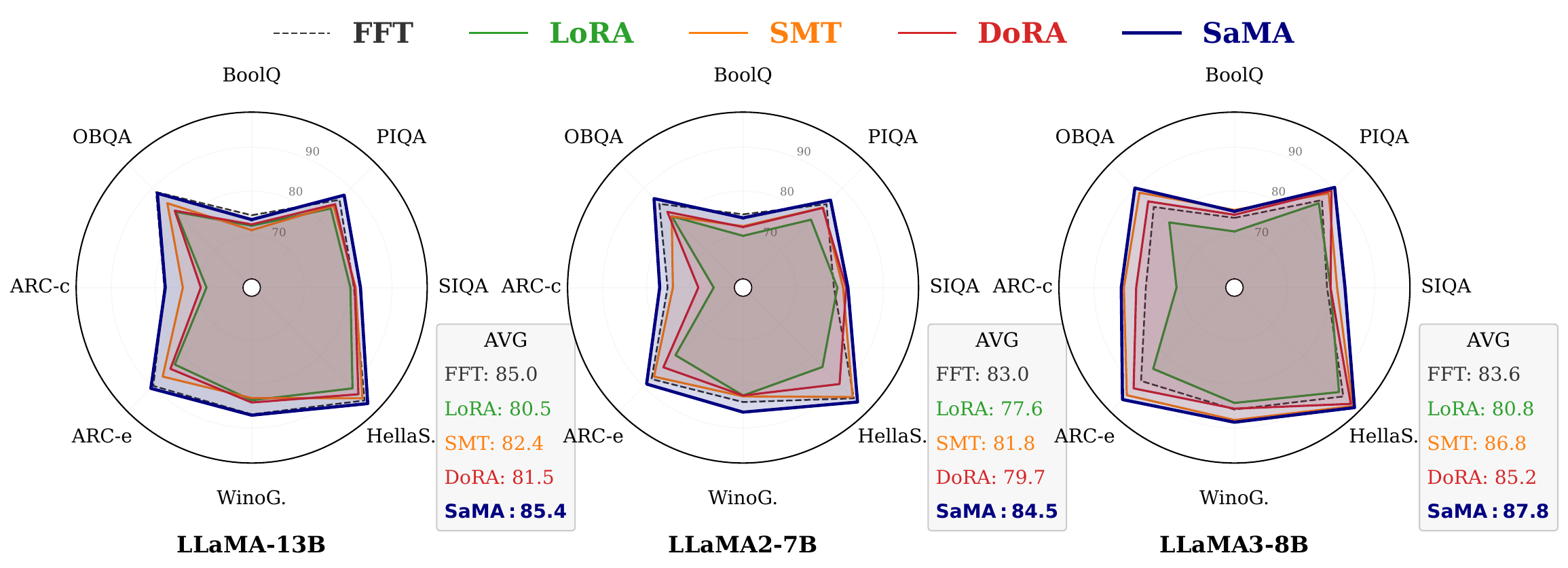}
    \caption{\label{fig:cms_radar_1x3} Task-wise performance comparison across \textbf{\textcolor{black}{FFT(black)}}, \textbf{\textcolor{Green4}{LoRA(green)}}, \textbf{\textcolor{Orange2}{SMT(orange)}}, \textbf{\textcolor{Red3}{DoRA(red)}} and \textbf{\textcolor{Blue4}{\ftmt (blue)}} on the Llama family. \ftmt only uses half as many parameters in these experiments, but maintains more accurate outputs across most of the tasks.}
\end{figure}

We further provide a comprehensive benchmark of the task in \cref{tab:app:cms_full_results}, including results of \ftmt and a diverse set of PEFT methods across several LLaMA models. \Cref{fig:cms_radar_1x3} also demonstrates the main comparison across methods for each of these models, except for LLaMA-7B. Overall, it is apparent that \ftmt outperforms most of the other baselines, having on par performance with FFT and sometimes surpassing it. Even in lightweight adapters, \ftmt still performs well, with slight drops only, on these complex tasks which illustrates its applicability in NLP problems.

Notably, the comprehensive benchmark in \cref{tab:app:cms_full_results} highlights a clear and coherent trend: across model scales and variants, \ftmt consistently delivers cutting-edge or tied-best average accuracy under substantially lower learnable-parameter budgets. \ftmt achieves highly competitive and often the best performance across these subtasks of complicated reasoning without relying on FFT techniques like~\citep{pan2024lisalayerwiseimportance,he2025smt}. The results also indicate that \ftmt scales robustly and reliably from smaller to larger Llama variants, from the first generation to the current one of Llama3, preserving effectiveness rather than showing any signs of trade-off in lightweight adapters. Furthermore, \ftmt also demonstrates the effectiveness and robustness over higher ranks, without any degradation due to optimization. Moreover, the benchmark indicates that \ftmt offers a favorable balance of expressivity, omni-rank, parameter-efficiency, and practical scalability for modern PEFT.

For example, on Llama-13B, $\ftmrm_{16,16}$ reaches an average score of $85.4$ with only $0.336\%$ parameters, compared with LoRA at $80.55$ with $0.67\%$, or FFT at $85.0$. On Llama2-7B, $\ftmrm_{16,16}$ attains $84.5$ on average with $0.42\%$ parameters, compared with SMT at $81.8$ under $0.84\%$ (and $83.4$ at SMT-Best with $4.91\%$), or FFT at $83.0$. On Llama3-8B, $\ftmrm_{16,16}$ reports $87.8$ average with $0.351\%$ parameters, or even $\ftmrm_{4,4}$ achieves $87.4$ with only $0.09\%$ parameters, while DoRA$_{r=32}$ and HiRA with about $0.71\%$, report $85.2$ and $ 86.7$, respectively.
\subsection{Attention fine-tuning on Commonsense reasoning}
\tableCmsAttnResult
Looking closely into details of QuanTA experiments~\citep{chen2024quanta}, it is noticeable that they added adapters into Q-V modules only in each transformer layer in order to keep simple implementation of tensor factorization for square matrices. Given that both methods are adapter-oriented, we argue that it is much fairer to adapt them to the same modules while doing fine-tuning foundation models. Hence, we conduct auxiliary experiments on these Llama model family (including Llama2-13B). That is to establish an equitable assessment between them, then the outputs are reported in \cref{tab:main_results_final}.

\subsection{Efficiency validation}
\label{sec:app_further_eff_valid}
\begin{figure}[htbp]
    \centering
    \includegraphics[width=0.95\linewidth]{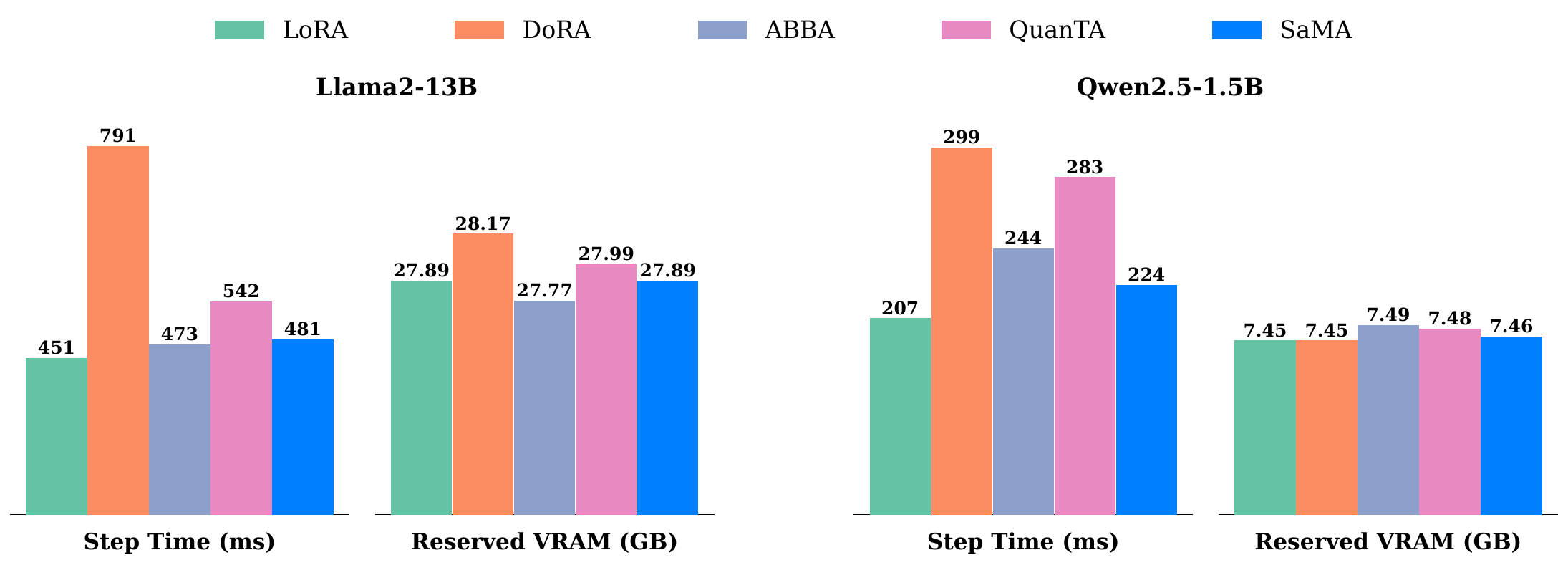}
    \caption{\label{fig:app_timememory_2models} Time complexity and memory footprint on Qwen2.5-1.5B and Llama2-13B. \ftmt generally maintains a memory footprint nearly identical to LoRA, while incurring only a modest step-time overhead, at less than 9 per cent, and remains substantially faster than DoRA and QuanTA.}
\end{figure}

To make comparison of adapter efficiency, we align the configurations of all methods to match the number of trainable parameters. The tests are conducted across several foundation models of different sizes so that they produce an unbiased judgment. The reports are constructed by monitoring time per iteration and reserved memory after 50 warmup steps, averaged 3 times over 25 iterations, while other settings are kept the same across all the experiments.

In particular, we consider a baseline set including LoRA~\citep{hu2022lora}, QuanTA~\citep{chen2024quanta}, \ftmt as 3 distinct representatives of fine-tuning via reparameterization. In addition, we adopt ABBA~\citep{singhal2025abba} -- Hadamard product of low rank matrices, and DoRA~\citep{liu2024dora} -- an improvement of LoRA due to the prevalence of LoRA family.

Besides \cref{fig:exp_time_memory_llama2b13}, additional results are also reported in \cref{fig:app_timememory_2models}. Similar to the pattern in \cref{fig:exp_time_memory_llama2b13}, there exists consistency in the relative resource profiles of these approaches in regard to computational time and memory consumption across different model scales. These results indicate that the structured higher-rank capacity of \ftmt introduces little additional computational and memory overhead in practice.

\subsection{Ablation study on adapter initialization}
\label{sec:app_exp_abinit}

\begin{figure}[htbp]
    \centering
    \includegraphics[width=0.95\linewidth]{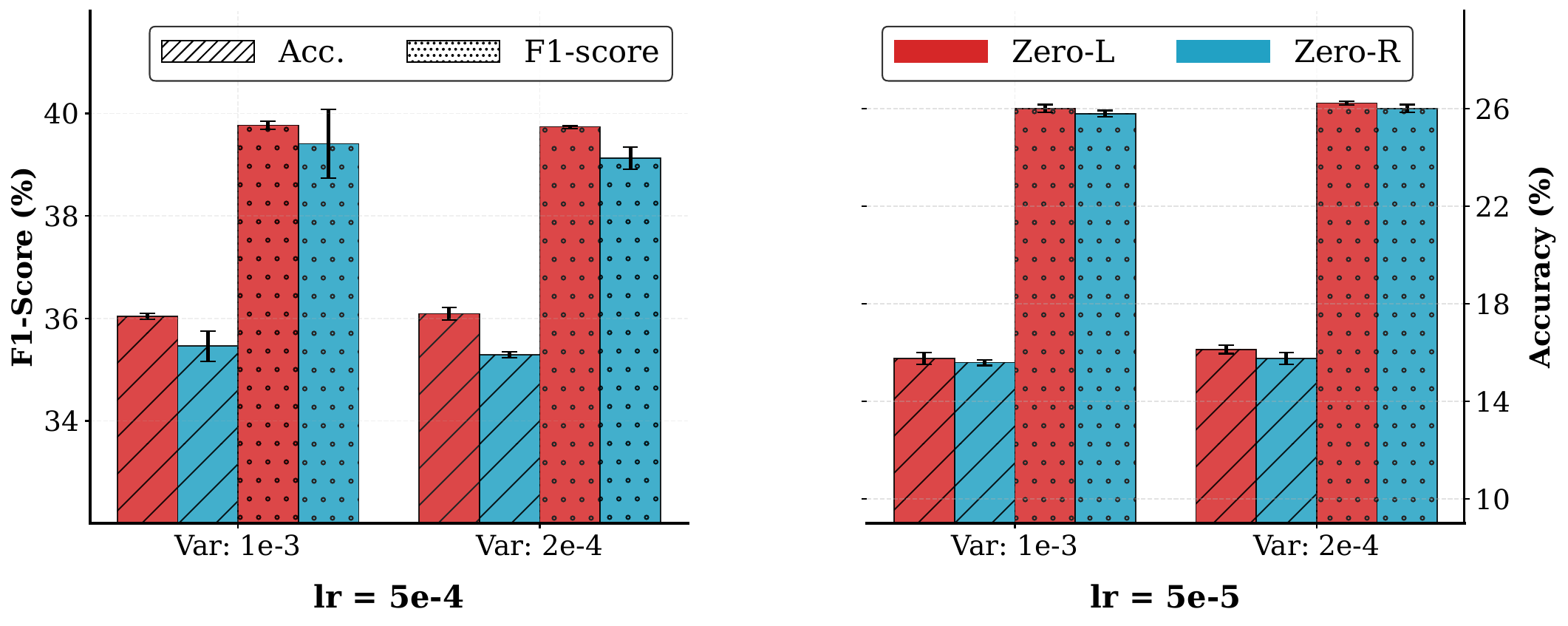}
    \caption{\label{fig:app_exp_abinit} Sensitivity analysis on initialization schemes of \ftmt adapters. Either of the two factors is equal to zero, the remaining factor follows a normal distribution with variance on the horizontal axes. We use both metrics: accuracy (right axis) and F1-score (left axis) for comparison.}
\end{figure}
For these experiments, we fine-tune Qwen2.5-1.5B with a fixed \ftmt configuration of $(4,4)$ for three attention modules: \texttt{q\_proj, k\_proj, v\_proj}. The training set is a 600-sample subset of DROP. The final checkpoints are evaluated after 1 epoch. To improve the robustness of the final conclusion, we consider multiple learning rate and initialization variances. \cref{tab:ablation_init} in \cref{sec:main_ablation_init} and \cref{fig:app_exp_abinit} provide the quantitative outputs and visualization, respectively.

\textbf{Observation: } Across these settings, \ftmt with zero-$\rmL$ initialization consistently yields better accuracy and F1-scores compared to zero-$\rmR$, at about $1-2\%$. Furthermore, zero-$\rmL$ demonstrates superior stability; it achieves a remarkably lower variance across most experiments, with a small standard deviation ($std$) of 0.25.

The results also advocate another critical insight regarding robustness. While performance margins exist between different configurations, our approach exhibits a remarkable intrinsic robustness and stabilizing effects, maintaining reliable efficacy regardless of initialization schemes.

\subsection{\ftmt performance when increasing parameter budgets}

We further evaluate the performance of \ftmt against LoRA and FFT using the Qwen2.5-1.5B model~\citep{qwen2023report} on the DROP dataset. The primary goal is to analyze how performance scales under different parameter budgets $(r)$ -- equivalent to LoRA rank $r$ and $(n_1,m_2)=(r,r)$ in \ftmt. In all settings, we only train Query and Value modules to make a completely fair comparison.

The key finding is that \ftmt performs better when the parameter counts increase, and steadily approaches FFT results. Meanwhile, LoRA, in such case, exhibits a performance gap to FFT, regardless of adding more budget. A similar observation is also reported in \citep{he2025smt}, suggesting more comprehensive experiments to investigate in future work.

\subsection{Fine-tuning Vision transformer}
Along with the main benchmark on complicated NLP downstream tasks on large foundation models, we further demonstrate the versatility and effectiveness of \ftmt in vision domains. Particularly, we extend our evaluation in image classification by post-training the ViT-B/16 architecture~\citep{dosovitskiy2020image}, pretrained on the ImageNet-21K~\citep{deng2009imagenet} on VTAB-1K~\citep{zhai2020vtab1k}. Although the classification task and baseline model are not sophisticated, we still gain overall judgment of \ftmt in lightweight modes for common problems.
\tableAppVtabResults
The VTAB-1K~\citep{zhai2020vtab1k} benchmark is a diverse suite containing 19 datasets. There are 3 main groups: Natural, Specialized, and Structured. While these datasets cover various potential facets such as classification, object categories, distinct semantics; each task includes only 1000 images for training.

\begin{figure}[htbp]
    \centering
    \includegraphics[width=0.90\linewidth]{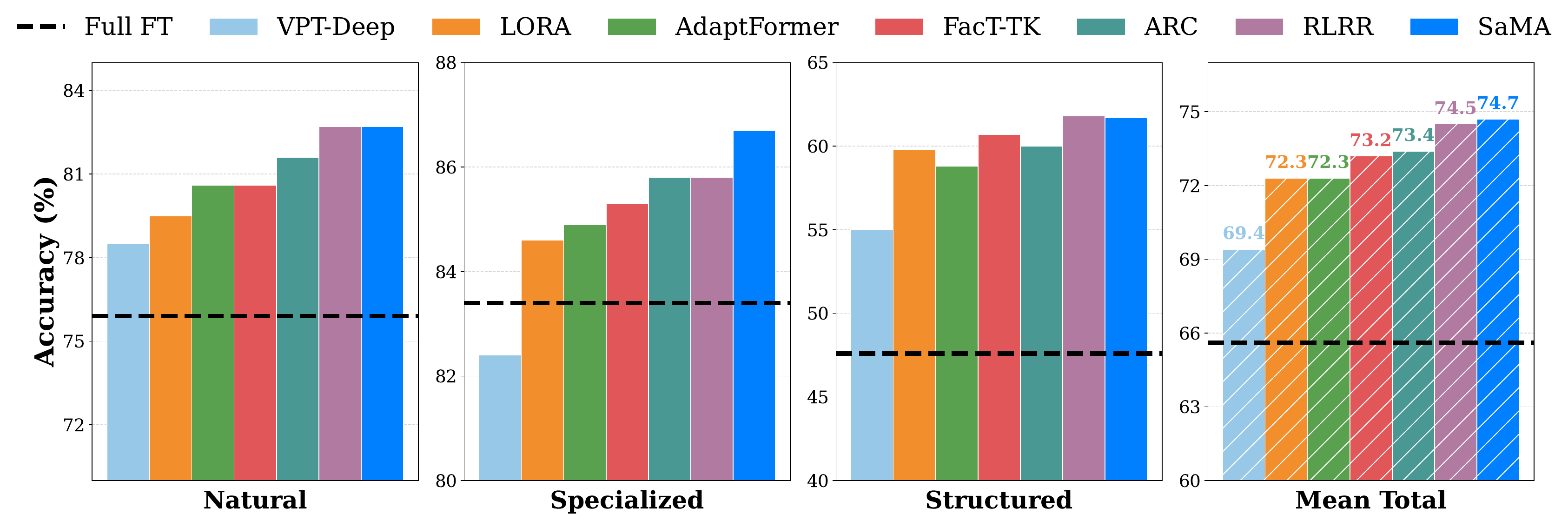}
    \caption{\label{fig:vtab_subplots_output} Performance Comparison on VTAB-1k Benchmark. In the lightweight regimes and image classification tasks, \ftmt uniformly demonstrates effectiveness in terms of group performance and global results.}
\end{figure}

The final outcomes are detailed in \cref{tab:app_vtab_results}, highlighting the superior performance of \ftmt across most settings and in overall metrics. On average, \ftmt with lightweight adapters yield results that are competitive or on par with the baselines, with notable gaps -- exceeding the others on the final scores. We further illustrate the performance on each group and the total sets of noticeable methods in \cref{fig:vtab_subplots_output} where \ftmt performance trends are consistent in each group.

\newpage
\section{Ablation studies }%-- Understanding \repname parameterization updates}
\label{sec:app_ablation}
To afford deeper insight into fine-tuning with \repname matrices, we examine extensive ablation studies on \ftmt and carry out pertinent analyses, further elucidating critical aspects of the method.
\subsection{Module-specific adaptation}
\vspace{-5pt}
\begin{figure}[htbp]
	\centering
	\includegraphics[width=0.80\linewidth]{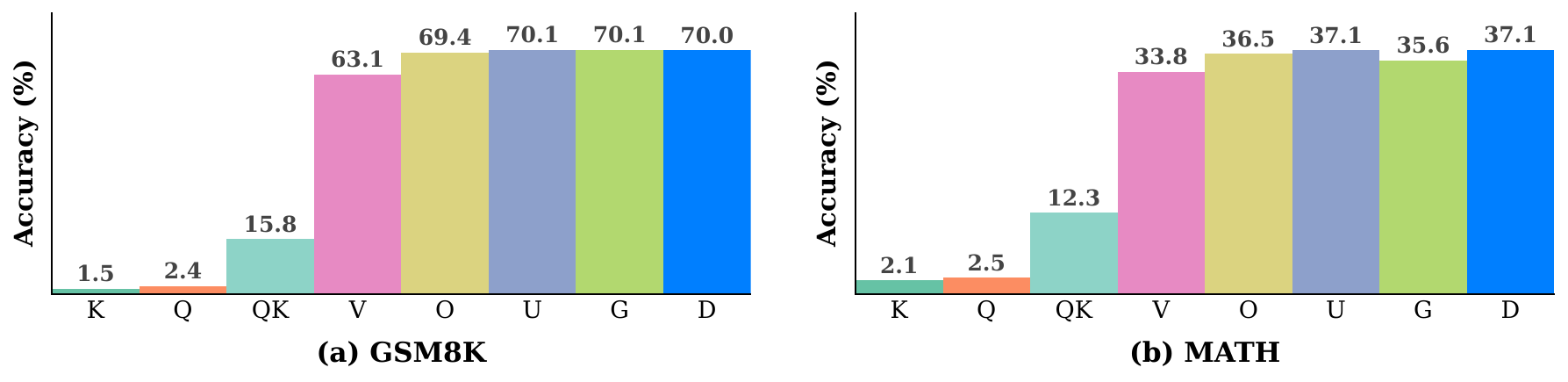}
	\caption{\label{fig:ablation_modules}Comparative contribution of individual components in transformer layers: Key, Query, Value, Output, Up, Gate and Down projections to \ftmt fine-tuning performance.}
\end{figure}

\Cref{fig:ablation_modules} investigates the impact of fine-tuning a single module in transformer layers of Qwen2.5-1.5B with \ftmt adapters, while keeping the number of trainable parameters fixed across settings.\footnote{Except for \textbf{Query-Key} (QK) fine-tuning, which is 10\% more than the others.} A clear pattern emerges across both tasks: tuning only Query or Key contributes very little, whereas adapting the Value and Output projections results in substantially larger gains. One possible explanation is that adapting Query or Key only changes attention weights, but does not directly update the value content being propagated, which may limit their effectiveness. Further rigorous research might be needed to provide a comprehensive analysis.

Meanwhile, the strongest results are consistently obtained in the \texttt{MLP} modules, indicating the most impactful factors. For example, on GSM8K, they all reach around 70\%, compared to 63\% -- 69\% in \texttt{v\_proj} and \texttt{o\_proj}, respectively. The trend is similar on MATH -- a more complex task, where Up and Down-tuning reach 37\%. These reflect the most functional roles of feed-forward layers: neural memory~\citep{geva2021transformer}. They store and adapt task-relevant procedural knowledge in math. However, the gaps between \texttt{up\_proj} and \texttt{down\_proj} are small, suggesting an overall balance between their functions.

\subsection{Asymmetric configuration of \texorpdfstring{\boldmath $(n_1, m_2)$}{(n1, m2)}}
\begin{wraptable}{r}{0.40\textwidth}
  \vspace{-12pt}
  \centering
  \small
  \caption{\ftmt performance with different configurations of hidden rank 64.}
  \label{tab:ablation_rank}
  {\renewcommand{\arraystretch}{1.3}
  \begin{tabular}{ccc}
      \toprule
      \textbf{$(n_1 - m_2)$} & \textbf{\#Params} & \textbf{\textit{F$_1$}-score$_{\pm \text{std}}$} \\
      \midrule
      $(16-4)$ & 3.28M & \(32.61_{\pm 1.03}\)  \\
      $(8-8)$ & 2.62M & $32.58_{\pm0.60}$ \\
      $(4-16)$ & 3.28M &  $31.9_{\pm1.56}$ \\
      \bottomrule
  \end{tabular}}
\end{wraptable}
In \cref{sec_method}, each \ftmt adapter is configured by two primary hyperparameters $(n_1, m_2)$, whose interaction controls both the intermediate rank and the total trainable parameters. An important question is how to choose the two factors from the final rank $r=n_1m_2$. Since it is recommended to use $(n_1, m_2)$ as divisors of input/output dimensions -- to avoid unnecessary padding, it is more straightforward to compare different configurations that result in the same rank, as reported in \cref{tab:ablation_rank}.

Overall, with the same training schemes, the symmetric configuration $(n_1=m_2)$ achieves performance on par with the best asymmetric settings, while providing a more favorable trade-off between parameter efficiency and capability. Moreover, it also shows significant improvement of robustness across random seeds. In the \texttt{MLP} layers here, the asymmetric configurations require about 25\% more parameters than $(8,8)$, yet do not yield consistent gains and exhibit noticeably higher variance across random initializations. By contrast, the symmetric setting offers a better balance between efficacy, parameter cost, and stability, making it the more practical default choice in most experiments.

\section{Understanding the adapter via mechanistic analysis}
\label{sec:app_analysis}
\subsection{Does \ftmt leverage the Omni-rank property?}
\label{sec:app_ana_erank}
To gain deeper insight beyond the main benchmark scores, we further carry out post-hoc analyses on adapters from two complementary perspectives: structure of learned representation and optimization dynamics by fine-tuning Qwen2-1.5B on a subset of DROP.

\begin{figure}[htbp]
    \centering
    \includegraphics[width=0.80\linewidth]{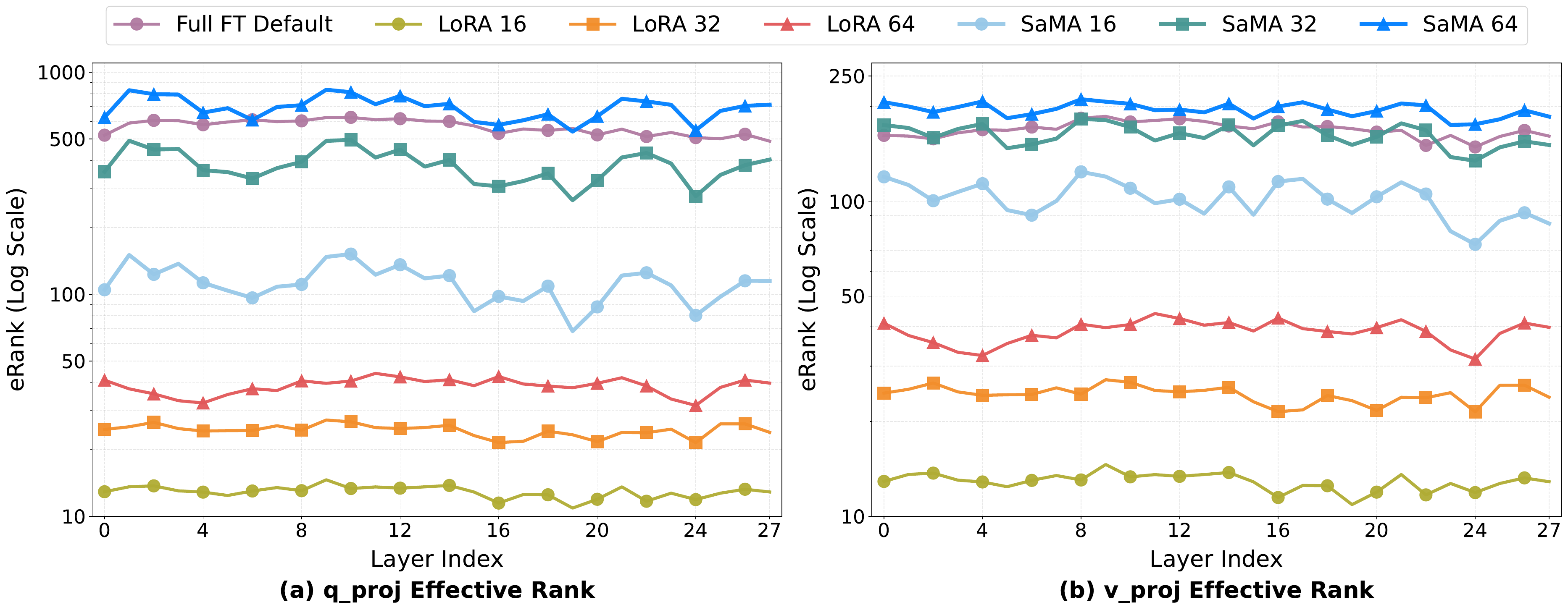}
    \caption{\label{fig:app_ana_erank} Effective rank of \texttt{q\_proj} and \texttt{v\_proj} modules on Qwen2-1.5B, trained on DROP. \ftmt utilizes higher nominal capacity and converts it into a set of rich update features.}
\end{figure}

While \cref{theo_sama_rank} provides a generic formula for the adapter nominal ranks, a natural question is whether the higher rank is translated into the learned update, or whether the optimization collapses the solution into a much lower-dimensional subspace when the parameter budget increases. In such cases, the fine-tuning adapter is really in low dimensions. To examine this, we analyze the effective rank \citep{roy2007effective} (See \cref{sec:app_detail_express}) of the adapter weights and therefore provides a sharp view of how hidden features span in practice.

\begin{figure}[htbp]
    \centering
    \includegraphics[width=0.98\linewidth]{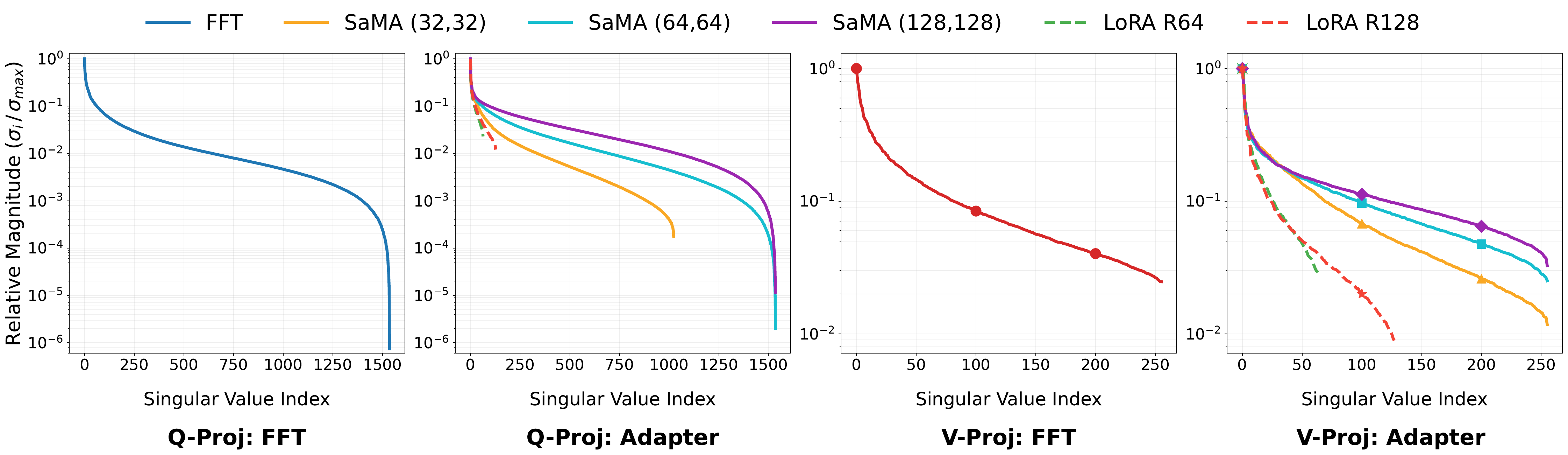}
    \vspace{3pt}
    \caption{\label{fig:ana_spectrum_fft_lora} Singular value spectrum comparison between \ftmt and FFT on Q,V modules. Complementing \cref{fig:app_ana_erank}, \ftmt shares richer long-tailed spectrum patterns in common with FFT, being able to achieve a diverse singular value profile while LoRA shows a sharper energy concentration}
\end{figure}

As shown in \cref{fig:app_ana_erank}, \ftmt tends to preserve intense long-tailed spectrum records in both \texttt{q\_proj} and \texttt{v\_proj} modules, and might surpass the effective rank of FFT adapter\footnote{We subtract the pre-trained matrices from the fine-tuned matrices, then analyze the adapter $\Delta \rmW$} if there are sufficient parameters. It indicates that:
\begin{enumerate*}[label=(\roman*)]
    \item the learned updates occupy a broader subspace rather than concentrate most into a few dominant directions.
    \item the long-tailed singular values really matter in matrix expressiveness, hence increasing adapter rank is a means of making performance improvement.
\end{enumerate*} 
This observation is consistent with the point that \ftmt better exploits its allocated capacity and is less prone to dimensional collapse due to the ``strictly low intrinsic rank assumption''.
This interpretation is further supported by the singular value spectra in \cref{fig:ana_spectrum_fft_lora}. Apart from low-rank updates, \ftmt exhibits a less concentrated but long-tailed spectrum by virtue of higher rank property, while remaining qualitatively close to FFT. These paradigm suggests that the nominal capacity of \ftmt is not merely theoretical, but translated into a rich set of learned features in practice. In addition, compared to \ftmt and FFT, LoRA exhibits a faster spectral decay, even at relatively higher ranks such as $r=64 \text{ or } 128$.

% Compared with \ftmt, LoRA exhibits a faster spectral decay, especially in \texttt{v\_proj}, even at relatively high nominal ranks such as $r=64$ and $r=128$.
% 
\subsection{Gradient behavior during optimization}
\label{sec:app_ana_grad_norm}
\begin{figure}[htbp]
    \centering
    \includegraphics[width=0.95\linewidth]{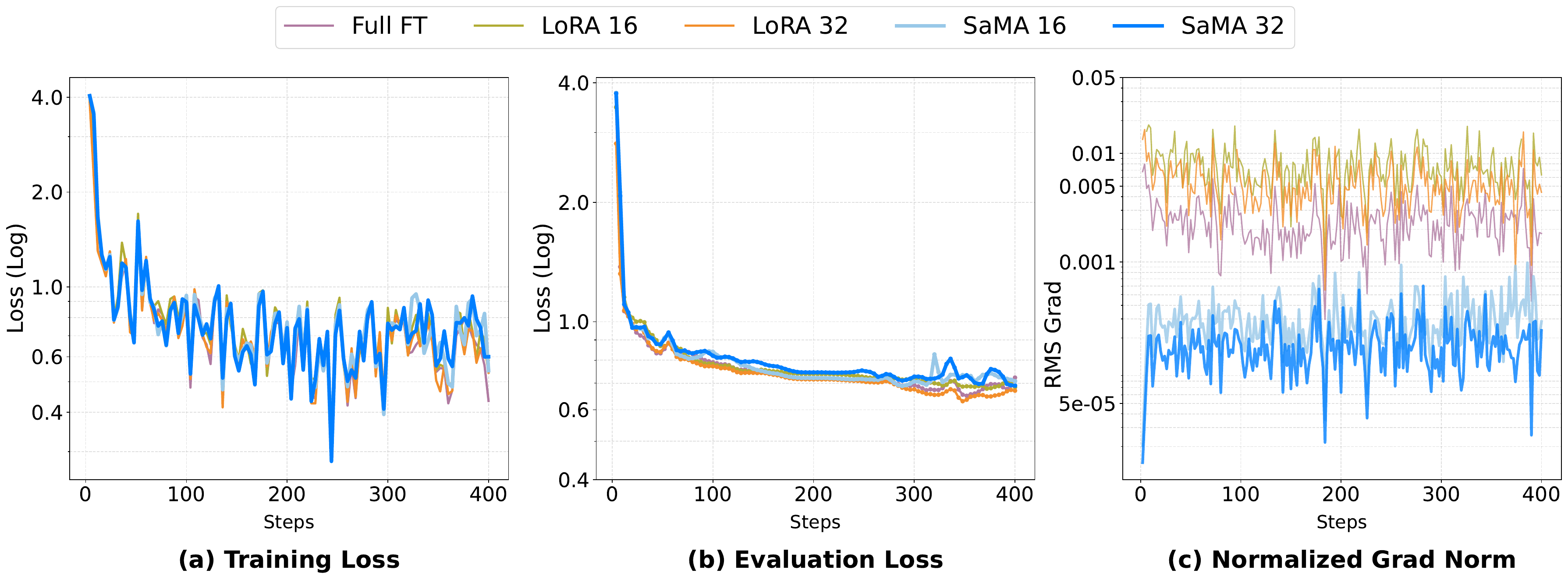}
    \caption{\label{fig:app_ana_grad} Fine-tuning metric comparison: (a) Training loss. (b) Evaluation loss (c) Normalized Gradient norm on DROP. Although different adapters achieve broadly similar loss trajectories, \ftmt appears more stable in gradient behavior, consistent with better-conditioned optimization dynamics.}
\end{figure}

As shown in \cref{fig:app_ana_grad}, the loss curves of different adapters remain mainly comparable, whereas their gradient behavior is more clearly separated. In particular, \ftmt exhibits more stable gradient norms, indicate flat minima~\citep{hochreiter1997flat}.
The key distinction therefore lies not only in the final loss, but in the landscape as well as the optimization path taken to reach them that have critical effects on generalizability.

\begin{figure}[htbp]
    \centering
    \includegraphics[width=0.98\linewidth]{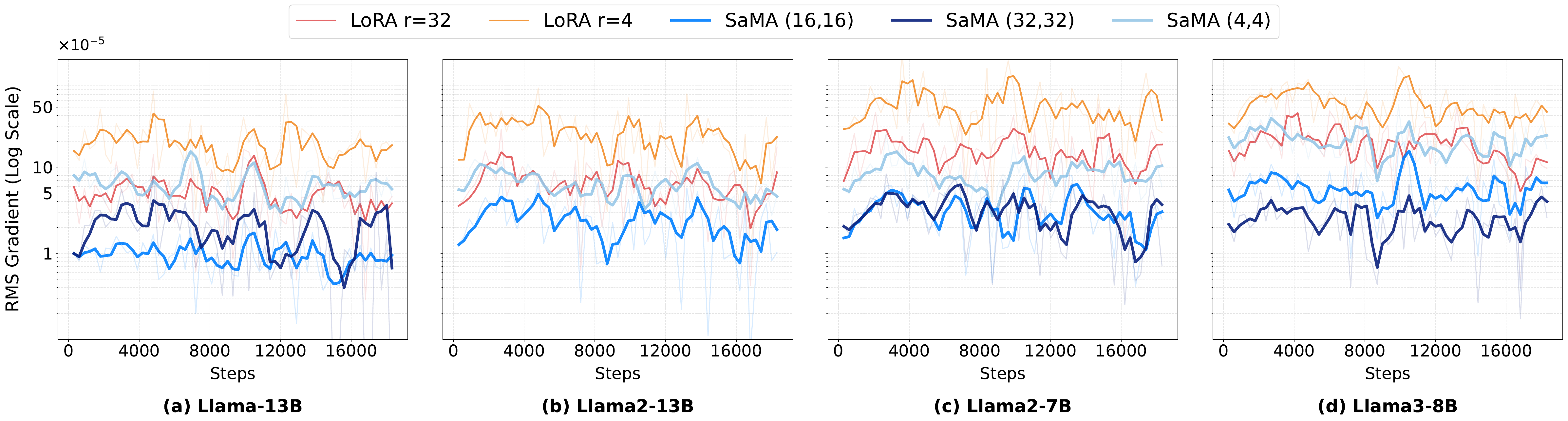}
    \caption{\label{fig:ana_cms_grad_norm} Empirical evaluation for moving average of normalized, by parameter count, gradient norms of full commonsense training (raw gradient data in transparent lines). \ftmt exhibits lower normalized gradient norms. The more parameter counts, the more norms tend to decrease, suggesting flatter minima and stabilizing effects.}
\end{figure}

We further investigate this phenomenon on the full commonsense reasoning experiments across Llama family, illustrating the output in \cref{fig:ana_cms_grad_norm}. The lower normalized gradient norms of \ftmt are broadly in consistent with better-conditioned optimization dynamics under a less degenerate parameterization~\cite{dinh2017sharp}. This observation also aligns with a long line of work relating gradients and flatter solutions to improve generalization gaps~\citep{keskar2017largebatch} via flat minima~\citep{hochreiter1997flat}, as well as with optimization methods that explicitly favor flatter neighborhoods~\citep{foret2020sharpnessaware}. However, there are many facets that need to be thoroughly examined in the future work, since the complicated relationship between flatness and generalization also depends on the parameterization and the notion of sharpness~\cite{dinh2017sharp} used while different adapter designs induce distinct invariant manifolds and degrees of gauge symmetry (See \cref{sec:express_repara,sec:app_repara}).

\newpage
\section{Dataset and experimental details}
\label{sec:app_data_exp_details}
\subsection{Details of datasets on experiments}
\label{app_dataset_infor}
\tableAppDatasetInfo
\textbf{DROP}~\citep{dua2019drop} is a reading comprehension benchmark designed to test discrete reasoning over natural language passages. Unlike traditional Q\&A datasets, DROP requires models to perform operations such as counting, addition, and comparison to answer correctly. In ~\citep{chen2024quanta}, the dataset is used to evaluate the performance of high-rank update with QuanTA.

Following the QuanTA experimental setup~\citep{chen2024quanta}, we take 2800 samples from the training set, and 1200 samples from the validation set as our test set since there is no official public test set on Hugging Face. Then, the 2800-sample raw training set is partitioned into a training set with 2000 samples, and a validation set with 800 samples. Moreover, $F_1$ score is also used as a metric for this task, which measures the closeness of the generative output compared to the ground truth answer.

\paragraph{Commonsense-147k} is a comprehensive training collection that is refined from commonsense170k~\citep{hu2023llmadapters} by removing all duplicates. It contains a mixture of data samples from 8 different benchmark datasets in the domain: 
\begin{itemize}
    \item \textbf{BoolQ}~\citep{clark2019boolq} dataset: consists of naturally occurring yes/no questions, generated in unprompted and unconstrained settings. The context passage in the original triple is eliminated.
    \item Physical Interaction: Question Answering or \textbf{PIQA}~\citep{Bisk2019PIQARA} dataset: comprises samples structured as multiple-choice questions, each presenting a question with two possible solutions that require physical commonsense to answer.
    \item Social-IQA or \textbf{SIQA}~\citep{sap2019socialIQa} dataset: focuses on social commonsense reasoning regarding the pragmatic implications of characters' intentions, reactions or communal reactions, which can measure LLMs' abilities to handle social reasoning by identifying the most plausible answer.
    \item \textbf{Hellaswag}~\citep{zellers2019hellaswag} dataset: a multiple choice benchmark for grounded commonsense reasoning. It includes natural language inference questions, offering a context and the most plausible continuation of a given situation from several candidates, many of which are adversarially generated to be misleading but syntactically plausible.
    \item \textbf{WinoGrande}~\citep{Keisuke2021winogrande} dataset: a large-scale pronoun resolution benchmark inspired by the Winograd Schema Challenge~\citep{winograd2012challenge}, but adjusted to improve the scale and robustness against the dataset-specific bias. it requires models to resolve ambiguous pronouns using commonsense reasoning rather than simple syntactic cues.
    \item ARC-Easy and ARC-Challenge (or \textbf{ARC-e} and \textbf{ARC-c}): two partitions from AI2 Reasoning Challenge~\citep{allenai2019arc} dataset, which consists of grade-school level, science example questions. ARC-Easy composes of quizzes solvable with straightforward retrieval or basic reasoning, while ARC-Challenge includes more difficult questions designed to enhance deeper reasoning and knowledge integration.
    \item Open book question answering - \textbf{OBQA}~\citep{mihaylov2018obqa} dataset contains multiple-choice elementary-level science questions requiring multi-step reasoning, which makes use of additional common and provided core science facts to get the final answer.
\end{itemize}
For arithmetic tasks, we concentrate on more difficult datasets with suitable test set size such as GSM8K, along with the more challenging problems of MATH. It could be considered as a standard mathematical reasoning benchmark with complicated inferences compared to MultiArith, AddSub, AquA, SingEq, SVAMP in~\cite{hu2023llmadapters}. In detail, we use these following datasets:
\begin{itemize}
    \item \textbf{MetaMathQA~\cite{yu2023metamath}:} the training set. All MetaMathQA questions are bootstrapped from the training sets of GSM8K and Hendrycks MATH by rewriting the samples from multiple viewpoints, but none of the augmented data is from the testing set. The dataset involves forward reasoning and backward reasoning samples.
    \item \textbf{Hendrycks MATH~\cite{hendrycks2021measuringmathematicalproblem}:} a benchmark of competition-style math problems collected from exams such as AMC and AIME. Each problem includes step-by-step solutions and final answers, covering diverse topics like algebra, geometry, number theory, and probability. Designed to test multi-step reasoning rather than rote formula application, MATH highlights the challenges large models face in genuine mathematical problem solving with explanation.
    \item \textbf{GSM8K~\cite{cobbe2021gsm8k}:} consists of high-quality grade school math word problems, each paired with a detailed explanation. It was designed as a benchmark for evaluating arithmetic reasoning and problem-solving in language models. Compared to MATH, GSM8K focuses on everyday math problems that require clear logical reasoning and generalization, rather than advanced techniques.
\end{itemize} 
\subsection{Experimental descriptions}
\label{app_exps}

\begin{wraptable}{r}{0.43\textwidth}
    \centering
    \vspace{-12pt} %
    \captionsetup{position=top,skip=5pt}
    \renewcommand{\arraystretch}{1.1}
    \tableMainCommonInfo
    \vspace{1pt} %
\end{wraptable}

All fine-tuning and evaluation processes are conducted on a single NVIDIA H200/H100 GPU with 140GB/80GB of memory. All the results reported in this section are averaged over multiple experiments and varying random seeds; the terms ``parameters'' and ``\#params'' here always refer to the number of trainable parameters. In these cases, all experiments are conducted using \texttt{bfloat16} precision in order to reduce memory usage, AdamW~\citep{ilya2019adamw} as the optimizer, and greedy sampling ($\text{vLLM: } \textit{temperature}=0$) unless otherwise specified in \cref{tab:app_hyper_train_common_info}. In each dataset, batch sizes reported here are the effective training batch sizes, including gradient accumulation steps if any.

\subsubsection{Baselines}
We evaluate \ftmt performance and compare it to the most common and up-to-date baselines. In this section, we provide a brief description of these baselines appearing in our experiments. In fact, we mostly consider fundamental fine-tuning techniques, along with many LoRA variants due to the advances in recent research works.

\textbf{Full fine-tuning FFT:} initializes the model with pre-trained weights and biases, then updates all parameters during training on a downstream task. For LLMs, this technique requires substantial memory and computational cost, but typically serves as a comparative performance bound of PEFT.

\textbf{Series~\citep{houlsby2019adapter}:} The original bottleneck adapter strategy that sequentially inserts trainable modules -- composed of a down-projection, a non-linearity, and an up-projection -- after the self-attention and feedforward neural network (FFN) layers,
while keeping the pre-trained parameters frozen.

\textbf{Parallel~\citep{he2021parallel}:} A structural variant that places the adapter modules alongside the frozen transformer layers (Attention or FFN) rather than sequentially. This approach maintains the original network depth by processing the input in parallel to the frozen weights and summing the results.

\textbf{LoRA~\citep{hu2022lora}:} freezes the pre-trained weights $\rmW_0 \in \mathbb{R}^{d \times k}$ and injects trainable rank decomposition matrices $B \in \mathbb{R}^{d \times r}$ and $A \in \mathbb{R}^{r \times k}$, applying the update as $\Delta \rmW = \rmB \rmA$. LoRA constrains the low intrinsic rank update $r \ll \min(d, k)$ to significantly reduce the number of trainable parameters.

However, the \textbf{rsLoRA}~\citep{dzievski2023rslora} variant justifies this scaling to $\nicefrac{\alpha}{\sqrt{r}}$. This rank-stabilized scaling preserves the update norm as the rank increases, ensuring that the optimization dynamics and optimal hyperparameters remain consistent across different rank choices $r$.

\textbf{DoRA~\citep{liu2024dora}:} improves upon LoRA by structurally decoupling the weight update into a learnable magnitude vector $\rvm$ and a directional component updated via low-rank adaptation, $\rmW + \Delta \rmW$, which comes with a trade-off with forward and backpropagation graphs during training time.

\textbf{PiSSA~\citep{meng2024pissa}:} initializes the low-rank adapter matrices $\rmA$ and $\rmB$ using the top-$r$ principal components from the Singular Value Decomposition (SVD) of the original weight matrix $\rmW_0$, rather than random Gaussian initialization.

\textbf{MiLoRA~\citep{wang2024milora}:} leverages SVD of pretrained weight matrices.
Unlike LoRA and PiSSA, MiLoRA selectively updates the \emph{minor components} -- those with the smallest singular values -- while preserving the major components of foundation weight matrices that encode core pretrained knowledge.

\textbf{KaSA~\citep{wang2024kasaknowledgeawaresingularvalue}:} Unlike PiSSA and MiLORA, which use SVD merely for initialization, KaSA first truncates minor singular values to remove noisy knowledge. It then reparameterizes task-specific updates using learnable singular values to dynamically activate the most relevant components.

\textbf{LoRA-Pro~\citep{wang2024loraproarelowrank}:} An optimization-centric variant that mathematically aligns the gradient updates. LoRA-Pro introduces a closed-form gradient correction ensuring that the weight update $\Delta W$ closely approximates the projection of the FFT gradient onto the low-rank subspace defined by $r$.

\textbf{HiRA~\citep{huang2025hira}:} A high-rank adaptation method that constructs the weight update $\Delta W$ using the Hadamard product (element-wise multiplication -- denoted as the operator $\odot$) of constituent matrices, rather than the standard matrix multiplication used in LoRA: $\Delta \rmW = (\rmB_1 \rmA_1) \odot \rmW_0$.

\textbf{ABBA~\citep{singhal2025abba}:} An alternative to HiRA~\citep{huang2025hira} that reparameterizes the weight update $\Delta W$ as the Hadamard product of two independent adapters: $\Delta \rmW = (\rmB_1 \rmA_1) \odot (\rmB_2 \rmA_2)$. By decoupling the update from $W_0$, ABBA allows the effective rank of the adaptation to reach up to $r_1 r_2$.

\textbf{SMT~\citep{he2025smt}:} constrains weight updates to a specific sparse or block-diagonal structure, trading off theoretical expressivity for computational and memory efficiency. The primary criterion for block preference is the gradient score during the first phase of warm-up.

\textbf{QuanTA~\citep{chen2024quanta}:} is a high-rank adaptation method inspired by quantum-style tensor factorization. It applies small 4D tensor interactions across tensorized feature dimensions through a sequence of transformations with $\binom{N}{2} = \frac{N(N-1)}{2} $ factors to enable richer mixing.
\paragraph{S$^2$FT~\citep{yang2024s2ftefficientscalable}:} selects sparsely and computes densely by updating a small subset of attention heads and hidden channels along with utilizing co-permutation to group these parameters into dense submatrices.

\textbf{LISA~\citep{pan2024lisalayerwiseimportance}:} does partial FFT by randomly freezing transformer layers with a given probability, except for embedding layers which always remain trainable to achieve  memory efficiency, based on LoRA’s skewed weight norm distribution.

\textbf{Galore~\citep{zhao2024galorememoryefficientllm}:} resolves the memory bottleneck in LLM training by projecting the weight gradients into a low-rank subspace, rather than approximating the weight matrices themselves. By dynamically switching these projection subspaces over time, GaLore enables FFT but is more memory-efficient.

\textbf{DiaBlo~\citep{gurses2025diablodiagonalblocks}:} directly updates only the structured diagonal blocks of the pre-trained weight matrices while keeping the off-diagonal blocks frozen. By avoiding matrix multiplication, DiaBlo simplifies the optimization landscape, targeting stable and robust convergence.

\textbf{FlexLoRA~\citep{liu2025flexloraentropyguidedflexible}:} a dynamic low-rank adaptation framework that simultaneously prunes redundant ranks in less important layers and expands ranks in critical layers under a global budget.

\textbf{RepLoRA~\citep{reploara25truong}:} reparameterizes low-rank matrices as the outputs of lightweight MLPs, motivated by theoretical connections to Mixture of Experts (MoE) models.

\textbf{RefLoRA~\citep{zhang2025reflorarefactoredlowrank}:} addresses the slow convergence and imbalanced weight updates of standard LoRA. The method dynamically applies the optimal low-rank factorization at each training step by minimizing an upper bound on the loss function.

\textbf{RandLoRA~\citep{albert2024randlora}:} enables full-rank weight updates by modeling them as a linear combination of fixed, low-rank random matrices, restricting the optimization solely to their diagonal coefficients.

\textbf{LaX~\citep{zhang2025laxboost}:} establishes residual connections that directly route the intermediate latent features from a preceding low-rank layer into the up-projection process of the current layer.

\subsubsection{The DROP dataset}
\label{app:sec_exp_drop}
\tableDropMainCommonInfo
\tableDropMainConfig
In \cref{tab:app_hyper_drop_train_common_info}, we show the common setting for DROP experiments along with hyperparameters in \cref{tab:app_hyper_drop_config}. The Llama2 family is used as the foundation models.
\subsubsection{Commonsense reasoning}
\label{app:data:cms}
To verify the effectiveness of our approach on commonsense reasoning downstream applications, we adapt several pretrained models of the Llama family, including Llama-7B, Llama-13B~\citep{touvron2023llama}, Llama2-7B and 13B~\citep{touvron2023llama2openfoundation}, Llama3-8B~\citep{grattafiori2024llama3herdmodels}.
This paper~\citep{hu2023llmadapters} introduces an integration of 170k samples from 8 different tasks, but we compare with their metadata and realize that there are many duplicates. Therefore,
we pre-process the joint COMMONSENSE170K dataset by removing duplicate samples, yielding COMMONSENSE-147K. Then, following common protocols in this domain~\citep{hu2023llmadapters}, we first fine-tune the models on the refined training data and evaluate the same adapted models on the full test sets of downstream tasks. Similar to the methodology used for the DROP dataset, we extract a subset of 1000 samples from COMMONSENSE-147K, then optimize the hyperparameters on this validation set. In the end, we select the best checkpoint based on validation set performance to evaluate the performance on the benchmark. The general hyperparameters and adapter configurations are listed in \cref{hyper:cms_full_train_info,hyper:cms_info_detailed_configs}, respectively.
\tableCmsFullCommonInfo
\tableCmsFullAdapterConfigInfo

\textbf{Attention fine-tuning -- a comparison with QuanTA} \quad The hyperparameters for these experiments are provided in \cref{hyper:cms_attn_train_info}:
\tableCmsAttnInfo

In \cref{tab:main_results_final}, although the total number of trainable parameters in \ftmt is slightly lower than that of QuanTA, \ftmt still outperforms in most of the global and local metrics. It further consolidates \ftmt capacity for fine-tuning on attention modules.
\subsubsection{Arithmetic reasoning}
\tableArithMainCommonInfo
\tableArithMainConfig
For arithmetic reasoning tasks, we fine-tune pretrained models of Mistral-7B~\cite{jiang2023mistral7b} and Gemma-2 9B~\cite{gemmateam2024gemma2b9} on a 20K-sample subset of MetaMathQA~\cite{yu2023metamath}, and evaluate their performance on test sets of MATH~\cite{hendrycks2021measuringmathematicalproblem} and GSM8K~\cite{cobbe2021gsm8k}, which require multi-step reasoning -- following~\cite{singhal2025abba}.
The experimental hyperparameters are provided in \cref{tab:app_hyper_arith_train_common_info,tab:app_hyper_arith_config}. In these experiments, we fine-tune all linear modules in each transformer layer, including attention and MLP projections.
\subsubsection{Vision fine-tuning}
\Cref{tab:vit_hyper} shows hyperparameters used in the experiments with VTAB-1k~\citep{zhai2020vtab1k} on Vision Transformers.
\begin{table}[htbp]
    \centering
    \caption{Implementation details of configurations, optimizer, and hyper-parameters. We select the best hyper-parameters for each task via grid search.}
    \label{tab:vit_hyper}
    \begin{tabular}{@{}ll@{}}
        \toprule
        \textbf{Configuration} & \textbf{Value} \\ \midrule
        Optimizer & AdamW \\
        Learning Rate & $\{0.1, 0.05, 0.02, 0.01, 0.005, 0.002, 0.001\}$ \\
        Weight Decay & $\{0.05, 0.01, 0.005, 0.001\}$ \\
        Dropout Rate & $\{0.2, 0.4\}$ \\
        Batch Size & $\{32\}$ \\
        Learning Rate Schedule & Cosine Decay \\
        Training Epochs & $(24, 29)$ \\
        Module & All transformer projections \\
        Warmup Epochs & 1 \\ \bottomrule
    \end{tabular}
\end{table}
\subsection{Template of Instruction fine-tuning}
Throughout this work, we use task-specific but structurally consistent prompt templates for both fine-tuning and inference. This subsection summarizes the templates used for DROP, commonsense reasoning, and math reasoning.
\subsubsection{Prompt template for DROP}
\begin{figure}[htbp]
    \centering
    \begin{promptbox}
        Below is a passage followed by a corresponding question that describes a task \\
        Write a response that appropriately completes the request with your answer. \\
        \\
        \#\#\# Instruction: \\
        Passage: \{passage\} \\
        Question: \{question\} \\
        \\
        \#\#\# Response:
    \end{promptbox}
    \caption{DROP dataset - Prompt template.}
    \label{fig:app:drop_prompt}
\end{figure}
Similar to \cref{app:sec:cms_prompt}, we use this prompt template during fine-tuning and inference of DROP~\citep{dua2019drop} dataset. The dataset consists of "Passage" and "Question" fields. Therefore, in the templates, we detail these headings and contents into the main body part, as in \cref{fig:app:drop_prompt}
\subsubsection{Commonsense reasoning data and prompt template}
\label{app:sec:cms_prompt}
\begin{table}[htbp]
    \centering
    \small
    \caption{The data template of each dataset used to create commonsense reasoning data samples.}
    \label{tab:app:data_template}
    % Cấu trúc: Cột 1 rộng 2.5cm, Cột 2 rộng 12cm (tổng ~14.5cm cho khổ giấy A4)
     \begin{tabular}{lp{10.5cm}}

        \toprule
        Dataset & Fine-tuning Data Template \\
        \midrule
        BoolQ & Please answer the following question with true or false, question: [QUESTION] \\
              & Answer format: true/false \\
              & the correct answer is [ANSWER] \\
        \midrule
        PIQA & Please choose the correct solution to the question: [QUESTION] \\
             & Solution1: [SOLUTION\_1] \\
             & Solution2: [SOLUTION\_2] \\
             & Answer format: solution1/solution2 \\
             & the correct answer is [ANSWER] \\
        \midrule
        SIQA & Please choose the correct answer to the question: [QUESTION] \\
             & Answer1: [ANSWER\_1] \\
             & Answer2: [ANSWER\_2] \\
             & Answer3: [ANSWER\_3] \\
             & Answer format: answer1/answer2/answer3 \\
             & the correct answer is [ANSWER] \\
        \midrule
        HellaSwag & Please choose the correct ending to complete the given sentence: \\ & \qquad \qquad[ACTIVITY\_LABEL]: [CONTEXT] \\
                  & Ending1: [ENDING\_1] \\
                  & Ending2: [ENDING\_2] \\
                  & Ending3: [ENDING\_3] \\
                  & Ending4: [ENDING\_4] \\
                  & Answer format: ending1/ending2/ending3/ending4 \\
                  & the correct answer is [ANSWER] \\
        \midrule
        WinoGrande & Please choose the correct answer to fill in the blank \\
                   & \qquad \qquad to complete the given sentence: [SENTENCE] \\
                   & Option1: [OPTION\_1] \\
                   & Option2: [OPTION\_2] \\
                   & Answer format: option1/option2 \\
                   & the correct answer is [ANSWER] \\
        \midrule
        ARC-e \& ARC-c & Please choose the correct answer to the question: [QUESTION] \\
                     & Answer1: [ANSWER\_1] \\
                     & Answer2: [ANSWER\_2] \\
                     & Answer3: [ANSWER\_3] \\
                     & Answer4: [ANSWER\_4] \\
                     & Answer format: answer1/answer2/answer3/answer4 \\
                     & the correct answer is [ANSWER] \\
        \midrule
        OBQA & Please choose the correct answer to the question: [QUESTION] \\
             & Answer1: [ANSWER\_1] \\
             & Answer2: [ANSWER\_2] \\
             & Answer3: [ANSWER\_3] \\
             & Answer4: [ANSWER\_4] \\
             & Answer format: answer1/answer2/answer3/answer4 \\
             & the correct answer is [ANSWER] \\
        \bottomrule
    \end{tabular}
\end{table}%
\begin{figure}[htbp]
    \centering
    \begin{promptbox}
        Below is an instruction that describes a task, paired with an input that provides further context. \\
        Write a response that appropriately completes the request. \\
        \\
        \#\#\# Instruction: \\
        \{instruction\} \\
        \\
        \#\#\# Input: \\
        \{input\} \\
        \\
        \#\#\# Response:
    \end{promptbox}
    \caption{Commonsense reasoning - Prompt template with a non-empty input field.}
    \label{fig:app:cms_prompt_non_empty}
\end{figure}

\begin{figure}[htbp]
    \centering
    \begin{promptbox}
        Below is an instruction that describes a task, paired with an input that provides further context. \\ 
        Write a response that appropriately completes the request. \\
        \\
        \#\#\# Instruction: \\
        \{instruction\} \\
        \\
        \#\#\# Response:
    \end{promptbox}
    \caption{Commonsense reasoning - Prompt template for examples with an empty input field.}
    \label{fig:app:cms_prompt_empty}
\end{figure}%
As described in \cref{app:data:cms}, the commonsense reasoning dataset collection includes 8 different sets. Each one in the domain entails distinct tasks, following~\citep{hu2023llmadapters,stf2023alpaca}, we create a structured framework that begins the same dialog template for disparate problems across the collection, followed by separately articulating the objective of the task, then the presentation of the relevant content and the corresponding solution.

\cref{fig:app:cms_prompt_non_empty,fig:app:cms_prompt_empty} shows the common instructional part of the template, while \cref{tab:app:data_template} demonstrates the distinct templates following the common part used to gather the commonsense reasoning data samples for fine-tuning.

\subsubsection{Math reasoning prompt template}
\begin{figure}[htbp]
    \centering
    \begin{promptbox}
        Below is an instruction that describes a task. \\
        Write a response that appropriately completes the request. \\ \\
        \\
        \#\#\# Instruction: \\
        \{instruction\} \\
        \\
        \# Let's think step by step. \\
        \#\#\# Response:
    \end{promptbox}
    \caption{Math reasoning - Prompt template.}
    \label{fig:app:math20k_prompt}
\end{figure}
\cref{fig:app:math20k_prompt} shows the actual prompt template when fine-tuning the model and generating outputs on math datasets~\citep{yu2023metamath,hendrycks2021measuringmathematicalproblem,cobbe2021gsm8k}

\section{Broader Impact}
\label{sec:app_broad_impact}
This paper presents work that aims to advance the field of machine learning. The proposed method, \ftmt, along with \repname parameterization, are designed to efficiently train LLMs and enhance fine-tuning performance. This will facilitate further research regarding PEFT in Large foundation models in the future. We believe that this work does not cause any significant societal consequences.

\clearpage
\section*{NeurIPS Paper Checklist}

\begin{enumerate}

\item {\bf Claims}
    \item[] Question: Do the main claims made in the abstract and introduction accurately reflect the paper's contributions and scope?
    \item[] Answer: \answerYes{} % Replace by \answerYes{}, \answerNo{}, or \answerNA{}.
    \item[] Justification: We theoretically and empirically verify the claims and contributions made in the abstract and introduction. See the abstract and introduction part.

\item {\bf Limitations}
    \item[] Question: Does the paper discuss the limitations of the work performed by the authors?
    \item[] Answer: \answerYes{} % Replace by \answerYes{}, \answerNo{}, or \answerNA{}.
    \item[] Justification: We discussed the limitations in \cref{sec:conclusion}.

\item {\bf Theory assumptions and proofs}
    \item[] Question: For each theoretical result, does the paper provide the full set of assumptions and a complete (and correct) proof?
    \item[] Answer: \answerYes{} % Replace by \answerYes{}, \answerNo{}, or \answerNA{}.
    \item[] Justification: See \cref{sec:app_theoretical}.

    \item {\bf Experimental result reproducibility}
    \item[] Question: Does the paper fully disclose all the information needed to reproduce the main experimental results of the paper to the extent that it affects the main claims and/or conclusions of the paper (regardless of whether the code and data are provided or not)?
    \item[] Answer: \answerYes{} % Replace by \answerYes{}, \answerNo{}, or \answerNA{}.
    \item[] Justification: Hyperparameters are in \cref{sec:app_data_exp_details}. The code is provided in the supplementary material.

\item {\bf Open access to data and code}
    \item[] Question: Does the paper provide open access to the data and code, with sufficient instructions to faithfully reproduce the main experimental results, as described in supplemental material?
    \item[] Answer: \answerYes{} % Replace by \answerYes{}, \answerNo{}, or \answerNA{}.
    \item[] Justification: All the code and data are available in the supplementary material.

\item {\bf Experimental setting/details}
    \item[] Question: Does the paper specify all the training and test details (e.g., data splits, hyperparameters, how they were chosen, type of optimizer) necessary to understand the results?
    \item[] Answer: \answerYes{} % Replace by \answerYes{}, \answerNo{}, or \answerNA{}.
    \item[] Justification: We provided all of our experimental settings in \cref{sec:app_data_exp_details}.

\item {\bf Experiment statistical significance}
    \item[] Question: Does the paper report error bars suitably and correctly defined or other appropriate information about the statistical significance of the experiments?
    \item[] Answer: \answerYes{} % Replace by \answerYes{}, \answerNo{}, or \answerNA{}.
    \item[] Justification: We conducted extensive hyperparameter searches for all baselines and our proposed method. Experimental results are tested multiple times to ensure stability and reliability. See \cref{sec:app_data_exp_details}.

\item {\bf Experiments compute resources}
    \item[] Question: For each experiment, does the paper provide sufficient information on the computer resources (type of compute workers, memory, time of execution) needed to reproduce the experiments?
    \item[] Answer: \answerYes{} % Replace by \answerYes{}, \answerNo{}, or \answerNA{}.
    \item[] Justification: See \cref{sec:app_data_exp_details}. Experiments are conducted on a single NVIDIA H100 or H200 GPU.
    
\item {\bf Code of ethics}
    \item[] Question: Does the research conducted in the paper conform, in every respect, with the NeurIPS Code of Ethics \url{https://neurips.cc/public/EthicsGuidelines}?
    \item[] Answer: \answerYes{} % Replace by \answerYes{}, \answerNo{}, or \answerNA{}.
    \item[] Justification: We have reviewed the code of ethics and confirmed that our research complies with its guidelines.

\item {\bf Broader impacts}
    \item[] Question: Does the paper discuss both potential positive societal impacts and negative societal impacts of the work performed?
    \item[] Answer: \answerYes{} % Replace by \answerYes{}, \answerNo{}, or \answerNA{}.
    \item[] Justification: See \cref{sec:app_broad_impact}
    
\item {\bf Safeguards}
    \item[] Question: Does the paper describe safeguards that have been put in place for responsible release of data or models that have a high risk for misuse (e.g., pre-trained language models, image generators, or scraped datasets)?
    \item[] Answer: \answerNA{} % Replace by \answerYes{}, \answerNo{}, or \answerNA{}.
    \item[] Justification: This paper presents work whose goal is to advance the field of LLM optimization. All of the datasets and models are public and we have checked them all. Our paper poses none of the risks mentioned in the question.

\item {\bf Licenses for existing assets}
    \item[] Question: Are the creators or original owners of assets (e.g., code, data, models), used in the paper, properly credited and are the license and terms of use explicitly mentioned and properly respected?
    \item[] Answer: \answerYes{} % Replace by \answerYes{}, \answerNo{}, or \answerNA{}.
    \item[] Justification: We have cited the original papers that produced the assets properly.

\item {\bf New assets}
    \item[] Question: Are new assets introduced in the paper well documented and is the documentation provided alongside the assets?
    \item[] Answer: \answerYes{} % Replace by \answerYes{}, \answerNo{}, or \answerNA{}.
    \item[] Justification: All the code is available in the supplementary material, along with an explicit README.md.

\item {\bf Crowdsourcing and research with human subjects}
    \item[] Question: For crowdsourcing experiments and research with human subjects, does the paper include the full text of instructions given to participants and screenshots, if applicable, as well as details about compensation (if any)? 
    \item[] Answer: \answerNA{} % Replace by \answerYes{}, \answerNo{}, or \answerNA{}.
    \item[] Justification: This paper does not involve crowdsourcing nor research with human subjects.

\item {\bf Institutional review board (IRB) approvals or equivalent for research with human subjects}
    \item[] Question: Does the paper describe potential risks incurred by study participants, whether such risks were disclosed to the subjects, and whether Institutional Review Board (IRB) approvals (or an equivalent approval/review based on the requirements of your country or institution) were obtained?
    \item[] Answer: \answerNA{} % Replace by \answerYes{}, \answerNo{}, or \answerNA{}.
    \item[] Justification: This paper does not involve crowdsourcing nor research with human subjects.

\item {\bf Declaration of LLM usage}
    \item[] Question: Does the paper describe the usage of LLMs if it is an important, original, or non-standard component of the core methods in this research? Note that if the LLM is used only for writing, editing, or formatting purposes and does \emph{not} impact the core methodology, scientific rigor, or originality of the research, declaration is not required.
    \item[] Answer: \answerNA{} % Replace by \answerYes{}, \answerNo{}, or \answerNA{}.
    \item[] Justification: LLMs are not used in our core method development.

\end{enumerate}

\end{document}